\documentclass[10pt,twocolumn,letterpaper]{article}

\usepackage{cvpr}              

\def\cvprPaperID{7054} 
\def\confName{CVPR}
\def\confYear{2022}

\usepackage{amsmath,amsfonts,bm}

\def\eqref#1{equation~\ref{#1}}

\def\1{\bm{1}}

\def\rvu{{\mathbf{i}}}

\def\rvs{{\mathbf{s}}}

\def\rvu{{\mathbf{u}}}

\def\rvx{{\mathbf{x}}}
\def\rvy{{\mathbf{y}}}
\def\rvz{{\mathbf{z}}}

\def\rmI{{\mathbf{I}}}

\DeclareMathAlphabet{\mathsfit}{\encodingdefault}{\sfdefault}{m}{sl}
\SetMathAlphabet{\mathsfit}{bold}{\encodingdefault}{\sfdefault}{bx}{n}

\def\gN{{\mathcal{N}}}

\usepackage[accsupp]{axessibility} 

\usepackage[colorlinks=true, pagebackref=true]{hyperref}
\usepackage{url}
\usepackage{xcolor}
\usepackage{lipsum}
\usepackage{graphicx}
\usepackage{subcaption}
\usepackage{xspace}
\usepackage{booktabs}
\usepackage{multirow}
\usepackage{wrapfig}
\usepackage{amsthm}
\usepackage{amsmath}
\usepackage[ruled]{algorithm2e}
\usepackage{commath}
\usepackage{breqn}
\usepackage{cuted}
\usepackage{capt-of}
\usepackage{pifont}
\usepackage{enumitem}
\usepackage[numbers,sort]{natbib} 

\newcommand{\bluetext}[1]{{#1}}
\newcommand{\vikash}[1]{{\color{blue}}}

\title{Generating High Fidelity Data from Low-density Regions using Diffusion Models
}

\author{Vikash Sehwag$^{\dagger}$ \enspace\enspace  Caner Hazirbas$^{\ddagger}$ \enspace  Albert Gordo$^{\ddagger}$ \enspace Firat Ozgenel$^{\ddagger}$ \enspace Cristian Canton Ferrer$^{\dagger}$ \vspace{0.1cm}\\
{$^{\dagger}$ Princeton University, $^{\ddagger}$ Meta AI \vspace{0.1cm}}\\
{\texttt{\footnotesize vvikash@princeton.edu, \{hazirbas, agordo, firatozgenel, ccanton\}@fb.com}}\\
}

\begin{document}

\twocolumn[{%
    \renewcommand\twocolumn[1][]{#1}%
    \maketitle
    \begin{center}
        \centering
        \captionsetup{type=figure}
        \includegraphics[width=0.95\textwidth]{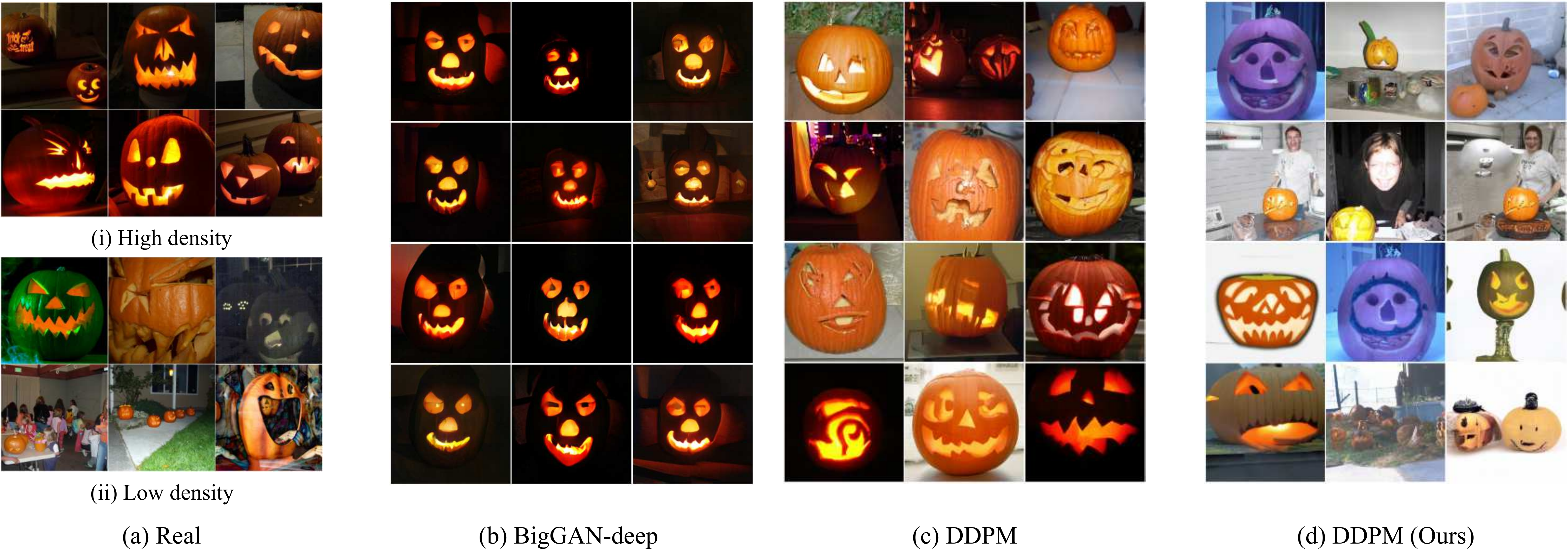}
        \vspace{-10pt}
        \captionof{figure}{\textbf{Real vs synthetic data.} We compare synthetic images from different generative models with real images from the low-density (1.a.i) and high-density (1.a.ii) neighborhoods of the data manifold, respectively. In 1.b we show uniformly sampled images from BigGAN~\cite{brock2018bigGandeep} and in 1.c we display images generated using the conventional uniform sampling process from the diffusion model (DDPM~\cite{ho2020denoisingdiffusion, dhariwal2021diffBeatGANs}). While diffusion model achieves much higher diversity than GANs, uniform sampling from them rarely generates samples from low-density neighborhoods. (1.d) Our framework guides the sampling process in diffusion models to low-density regions and generates novel high fidelity instances from these regions.\label{fig:teaser}\protect\footnotemark}
    \end{center}%
}]

\begin{abstract}
Our work focuses on addressing sample deficiency from low-density regions of data manifold in common image datasets. We leverage diffusion process based generative models to synthesize novel images from low-density regions. We observe that uniform sampling from diffusion models predominantly samples from high-density regions of the data manifold. Therefore, we modify the sampling process to guide it towards low-density regions while simultaneously maintaining the fidelity of synthetic data. We rigorously demonstrate that our process successfully generates novel high fidelity samples from low-density regions. We further examine generated samples and show that the model does not memorize low-density data and indeed learns to generate novel samples from low-density regions.\footnotetext{ImageNet~\cite{deng2009imagenet,ILSVRC15} has no explicit category for humans, though one might be present in some images. Thus generative models might generate synthetic images that include a human. We further conduct a rigorous analysis to validate whether the network has memorized any such information from training samples.}
\end{abstract}

\vspace{-15pt}
\section{Introduction}
Most common image datasets have a long-tailed distribution of sample density\footnote{We refer to the long-tail~\wrt sample density for each class. It is different from the long-tailed distribution over classes~\cite{liu2019ImageNetLongTail},~\ie, when some classes are heavily underrepresented than others.}, where the majority of samples lie in high-density neighborhoods of the data manifold. Samples from low-density regions often comprise novel attributes (Figure~\ref{fig:teaser}a) and have higher entropy than high-density samples~\cite{agarwal2020VOG}. However, due to their lower likelihood, curating even a small amount of such samples requires a dedicated effort~\cite{hendrycks2021naturalAdv}. 


Our goal is to leverage generative models to generate synthetic images from low-density neighborhoods. A natural requirement for this task is that the model should generalize to low-density regions. While generative adversarial networks (GANs) excel at generating high-fidelity samples, they have poor coverage, thus struggle to generate high-fidelity samples from low-density regions~\cite{brock2018bigGandeep} (Figure~\ref{fig:teaser}b). In contrast, autoregressive models have a high coverage but fail to generate high fidelity images~\cite{child2019sparseTransformer}.  We use diffusion-based models due to their ability to achieve high fidelity and high coverage of the distribution, simultaneously~\cite{ho2020denoisingdiffusion, nichol2021improvedDdpm}.

In training diffusion models, the goal is to approximate data distribution, which is often long-tailed. Diffusion models excel at this task, as we observe that the density distribution of uniformly sampled instances from the diffusion model is very similar to real data. 

Thus uniform sampling from these models leads to an imitation of real data density distribution, i.e., a long-tailed density distribution, where it generates samples from high-density regions with a much higher probability than from low-density regions (figure~\ref{fig:teaser}c). 
To alleviate this issue, we first modify the sampling process to include an additional guidance signal to steer it towards low-density neighborhoods. However, at higher magnitudes of this signal, the generative process is steered off the manifold, thus generating low fidelity samples. We circumvent this challenge by including a second guidance signal which incentives diffusion models to generate samples that are close to the real data manifold.




Since a very limited number of training samples are available from low-density regions, it is natural to ask whether diffusion models are generalizing in the low-density regions or simply memorizing the training data. After all, recent works have uncovered such memorization in language-based generative models~\cite{carlini2021LmMemorization, carlini2019secretSharer}. We conduct an extensive analysis to justify that diffusion models do not show signs of memorizing training samples from low-density neighborhoods and indeed learn to interpolate in these regions. We make the following key contributions.

\begin{itemize}[noitemsep,topsep=5pt,parsep=5pt,partopsep=0pt]
    \item We propose an improved sampling process for diffusion models that can generate samples from low-density neighborhoods of the training data manifold. 
    \item We validate the success of our approach using three different metrics for neighborhood density and provide extensive comparisons with the baseline sampling process in diffusion models. 
    \item We show that our sampling process successfully generates novel samples, which aren't simply memorized training samples, from low-density regions. This observation from our sampling process also uncover that despite a limited number of training images available from low-density regions, diffusion models successfully generalize in low-density regions. 
\end{itemize}

\section{Related work}
Diffusion-based probabilistic models~\cite{ho2020denoisingdiffusion, nichol2021improvedDdpm, dhariwal2021diffBeatGANs} and its closely related variants~\cite{song2019ScoreEstimateGrad, song2020ImprovedTechScore} are likelihood-based models that learns data distribution by learning the reverse process of the forward diffusion process. Following latest advances~\cite{dhariwal2021diffBeatGANs}, diffusion models achieve state-of-the-art performance, outperforming other classes of generative models, such as Generative adversarial networks (GANs), VQ-VAE~\cite{razavi2019VQVAE2}, and Autoregressive models~\cite{child2019sparseTransformer} on various metrics in image fidelity and diversity~\cite{nichol2021improvedDdpm, dhariwal2021diffBeatGANs}. Some of the key factors behind their success are the innovation on the architecture of the diffusion models~\cite{ho2020denoisingdiffusion, dhariwal2021diffBeatGANs}, simplified formulation for the training objective~\cite{ho2020denoisingdiffusion}, and use of cascaded diffusion processes~\cite{nichol2021improvedDdpm, dhariwal2021diffBeatGANs, ho2021cascadedDiffusion}.

Sampling from diffusion models is quite slow since it requires an iterative denoising operation. Reducing this overhead by developing fast sampling techniques is a topic of tremendous research interest~\cite{song2020ddim, jolicoeur2021gottaGoFast, kong2021fastSampleDDPM, watson2021LearnEfficientSampleDDPM}. Orthogonal to this direction, our interest is in sampling data from low-density neighborhoods. We further show that our sampling approach can be easily integrated with fast sampling techniques.

To measure neighborhood density around a sample, we use the Gaussian model of training data in the embedding space of a pre-trained classifier. Modeling images in embedding space is a common approach, particularly due to their alignment with human perception~\cite{zhang2018Lpips}, in numerous vision applications, such as outlier detection~\cite{sehwag2020ssd} and instance selection~\cite{devries2020instanceSelection}.

Across generative models, given a distribution learned by the model, there have been previous attempts in sampling from a targeted data distribution. Discriminator rejection sampling (DRS) and its successors \cite{azadi2018DRSGan, ding2020subsampleGANImfinetune} consider rejection sampling using the discriminator in a generative adversarial network (GAN). Similarly, Razavi~\etal~\cite{razavi2019VQVAE2} exploits a pre-trained classifier to reject samples that are classified with low confidence. Most often the goal is to filter out low fidelity samples, thus improving the quality of synthetic data. In contrast, our goal is to generate high fidelity samples from low-density regions of the data manifold. These samples are rarely generated by the model under uniform sampling, thus sampling them using a naive classifier-based rejection sampling approach leads to high-cost overheads. We instead opt to modify the generative process of diffusion models to guide it towards low-density neighborhoods of the data manifold.

The most closely related work to ours is from Li~\etal\cite{li2020ClassEmbedSmoothGAN}, which smoothes class embeddings of a BigGAN model to generate diverse images. In contrast, we focus on diffusion-based generative models. We also demonstrate the limitation of their approach with diffusion models in Appendix~\ref{app: SmoothEmbed}.

\section{Low-density sampling from diffusion models}
In this section, we first provide an overview of the sampling process in diffusion-based generative models. Next, we describe our modification in the sampling process for low-density sampling. 

\subsection{Overview of diffusion models}
\vikash{First define the task of learning model distribution from the data distribution.}
Denoising diffusion probabilistic models (DDPM)~\cite{ho2020denoisingdiffusion} model the data distribution by learning the reverse process (generative process) for a forward diffusion process. The forward process is often a Markov chain with Gaussian transitions,~\ie, $q(\rvx_t|\rvx_{t-1}) := \gN(\rvx_t; \sqrt{1-\beta_t}\rvx_{t-1}, \beta_t\mathbf{I})$. Given a large number of timesteps ($T$), this diffusion process sufficiently destroys the information in input samples ($\rvx_0$) such that $p(\rvx_T) := \gN(\rvx_t; \mathbf{0}, \rmI)$.

Reverse or generative process is also assumed to be a Markov process with Gaussian transitions that learns the inverse mapping,~\ie, $p(\rvx_{t-1} | \rvx_t)$, at each time step. This process is usually modeled with a deep neural network, parameterized by $\theta$, that learns the Gaussian transition such that $p_\theta(\rvx_{t-1} | \rvx_t) := \gN(\rvx_{t-1}; \bm{\mu}_\theta(\rvx_t, t), \mathbf{\Sigma}_\theta(\rvx_t, t))$. \vikash{Bold mu in equations.} 
\begin{align}\label{eq: base_nll}
    p_\theta(\rvx_0) = p(\rvx_T) \prod_{t=1}^{T} p_\theta (\rvx_{t-1}|\rvx_t)
\end{align}
The model is trained by maximizing the variational lower bounds on the negative log likelihood over the training data.

In order to sample synthetic data from diffusion models, we first sample a latent vector $\rvx_T \sim \gN(\mathbf{0}, \mathbf{I})$ and iteratively denoise it using the following procedure in reverse process.
\vspace{-15pt}
\begin{align}
    \rvx_{t-1} = \bm{\mu}_{\theta}(\rvx_t, t) + \mathbf{\Sigma}_{\theta}^{1/2}(\rvx_t, t)\rvz,~~~~\rvz \sim \gN(\mathbf{0}, \mathbf{I})
\end{align}

We refer to this approach as \textit{baseline} sampling process.

\subsection{Generating synthetic images from low-density regions on the data manifold}
In this section, we present our approach to generating samples from low-density regions of data manifold using diffusion-based models.

\subsubsection{Identifying low-density regions on data manifold}
Given a data distribution $q(\rvx)$, low-density regions or neighborhoods are part of the data manifold that have significantly lower sample density than the others. To develop techniques to sample from these regions, the first step is to characterize them.

\textit{Limitation of likelihood estimates from the diffusion model.} A natural choice to characterize manifold density is to use the likelihood estimate from the diffusion model itself (Equation~\ref{eq: base_nll}). After all, we expect the likelihood of getting a sample from high-density regions being higher than the low-density regions. However, due to its intractability for diffusion-based models, the likelihood estimates from the model are only an approximation of exact likelihood~\cite{sohl2015NonEqDDPM,ho2020denoisingdiffusion}. We find that these likelihood estimates are not a reliable predictor of manifold density as they fail to align with multiple commonly used metrics or with human judgment (Appendix~\ref{app: diffLikelihood}).
This trend aligns with a similar limitation of likelihood estimates in autoregressive models~\cite{nalisnick2018GenDontKnow}. 

We shift our focus to discriminative models since they are well-known to learn meaningful embeddings that align with human perception for images~\cite{zhang2018Lpips}. We measure the manifold density by estimating the likelihood of data in the embedding space. Let $(g \circ f)(.)$ be a discriminative model, where $f$ extracts embeddings for the input image and $g$ is the head classifier, most often a linear model. We model embeddings of each class using a Gaussian model and estimate the log-likelihood of a given image ($\rvx_i$) with class label $\rvy_i$ from this model. We refer to the negative log-likelihood as \textit{Hardness score ($H$)}. 
\vspace{-5pt}
\begin{align}
    \begin{split}
        H(\rvx, y) = \frac{1}{2} \Bigl[\bigl(f(\rvx) &- \mu_y\bigr)^T\Sigma_y^{-1} \bigl(f(\rvx) - \mu_y\bigr) \\ &+ \ln\bigl(\det(\Sigma_y)\bigr) + k\, \ln (2\pi) \Bigr]
    \end{split}
\end{align}

$\mu_y$ and $\Sigma_y$ refer to sample mean and sample covariance for embeddings of class $y$ and $k$ is the dimension of embedding space. We provide further analysis in Appendix~\ref{app: hardnessJustify} to justify that the decrease in manifold density leads to an increase in hardness score.

To sample from low-density regions, our approach is to guide the diffusion model to generate samples with high hardness scores,~\ie, equivalent to achieving low likelihood in the correct class. We maximize the following contrastive guiding loss for this task.
\vspace{-5pt}
\begin{equation}
    L_{g_1} (\rvx_i, y_i)  = \textnormal{log}\left[\frac{exp(H(\rvx_i, y_i)/\tau)}{\sum_{j=1}^{C}exp(H(\rvx_j, j)/\tau)}\right]
\end{equation}
where $\tau$ is the temperature and $C$ is the total number of classes. \vikash{Fix the notation issues in this section.}

This formalization of guiding loss function is fairly similar to cross-entropy loss on output softmax probabilities,~\ie, $(g \circ f)(.)$. Thus we also consider an equivalent loss function where instead of hardness score, we minimize the output softmax probability in the correct class.

\smallskip
\noindent\textbf{Incorporating guiding loss in sampling process.} The next step is to guide the sampling process to low-density regions by minimizing the log-likelihood of generated samples at each time step. We modify the sampling process as follows.
\vspace{-5pt}
\begin{align}\label{eq: sample_alpha}
\begin{split}
    \rvx_{t-1} = \bm{\mu}_{\theta}(\rvx_t, t) &+\, \mathbf{\Sigma}_{\theta}^{1/2}(\rvx_t, t)\, \rvz\\
    &+ \alpha\, \mathbf{\Sigma}_{\theta}(\rvx_t, t)\, \mathbf{\nabla}^*L_{g_1}(\rvx_{t}, y)
\end{split}
\end{align}
where $\rvz \sim \gN(\mathbf{0}, \mathbf{I})$, $\nabla^*$ refers to normalized gradients, and $\alpha$ is a scaling hyperparameter. We normalize gradients to disentangle the choice of scaling hyperparameter, $\alpha$, from the diffusion process time steps, $t$ (Appendix~\ref{app: normalGrad}). This formulation of sampling process is similar to Dhariwal~\etal~\cite{dhariwal2021diffBeatGANs}, with the modification that our loss function is designed to guide towards low-density regions and that we use normalized gradients. 

\subsubsection{Maintaining fidelity when minimizing likelihood}
\begin{figure}[!htb]
    \centering
    \includegraphics[width=0.9\linewidth]{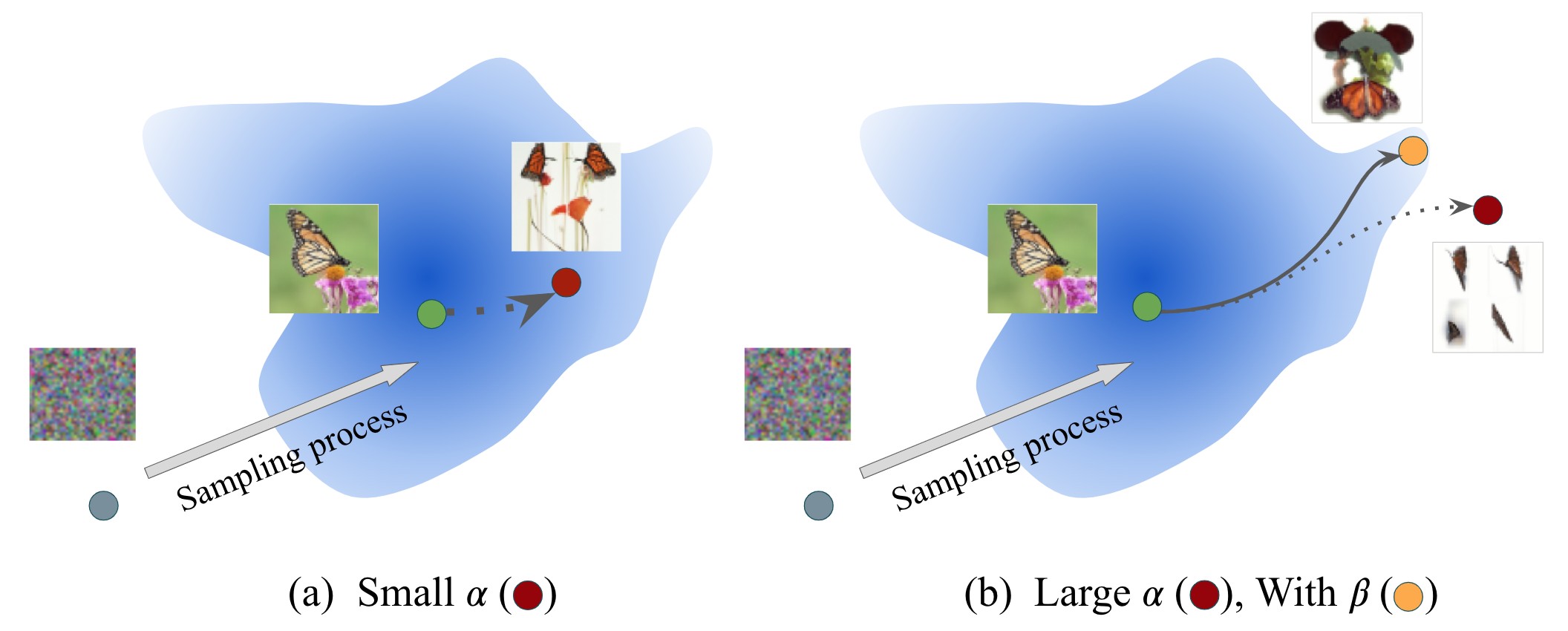}
    \caption{An illustration demonstrating that small $\alpha$ values successfully guide the sampling process to low-density regions (lighter colors) on the data manifold. However, at large values of $\alpha$, using additional guidance (by using a non-zero $\beta$) from the binary discriminator (Eq.~\ref{eq: final_sampling}) helps in staying close to data manifold. We provide a demonstration of it in figure~\ref{fig: a_b_grid}.}
    \vspace{-10pt}
    \label{fig: overview_steer}
\end{figure}

We find that the sampling process in Equation~\ref{eq: sample_alpha} is highly successful at smaller values of $\alpha$. However, with higher values of $\alpha$, the guidance term dominates the Gaussian transition term from the diffusion model and steers the sampling process off data-manifold, thus generating very low fidelity images (as illustrated in Figure~\ref{fig: overview_steer}). Its effect is exacerbated by model distribution often not being a good approximation of data distribution in low-density regions, in particular, due to the reason that a very limited number of training samples are available from low-density regions.
\vspace{-10pt}

We include another term in the sampling process to compel it to stay close to the data manifold. In particular, we train a binary discriminator, with hardness score $H'$, that distinguishes between synthetic and real samples. While sampling, we enforce synthetic images to stay close to the real data manifold by maximizing the following loss value. 
\begin{equation}
    \vspace{-5pt}
    L_{g_2} (\rvx_i)  = -\textnormal{log}\left[\frac{exp(H'(\rvx_i, 1)/\tau)}{\sum_{j=0}^{1}exp(H'(\rvx_j, j)/\tau)}\right]
\end{equation}
Here class zero and one represents synthetic and real images, respectively. In low-density regions, where model distribution is likely a poor approximation of real data distribution, this objective forces the diffusion model to generate samples that are closest to the real data manifold. Our final sampling process is following.
\vspace{-5pt}
\begin{align} \label{eq: final_sampling}
\vspace{-5pt}
\begin{split}
    \rvx_{t-1} = \bm{\mu}_{\theta}(\rvx_t, t) &+ \mathbf{\Sigma}_{\theta}^{1/2}(\rvx_t, t)\,\rvz\\
    &+ \alpha\, \mathbf{\Sigma}_{\theta}(\rvx_t, t)\, \mathbf{\nabla}^* L_{g_1}(\rvx_{t}, y)\\
    &+ \beta\, \mathbf{\Sigma}_{\theta}(\rvx_t, t)\, \mathbf{\nabla}^* L_{g_2}(\rvx_{t})
\end{split}
\end{align}

\begin{figure}[!b]
    \centering
    \includegraphics[width=\linewidth]{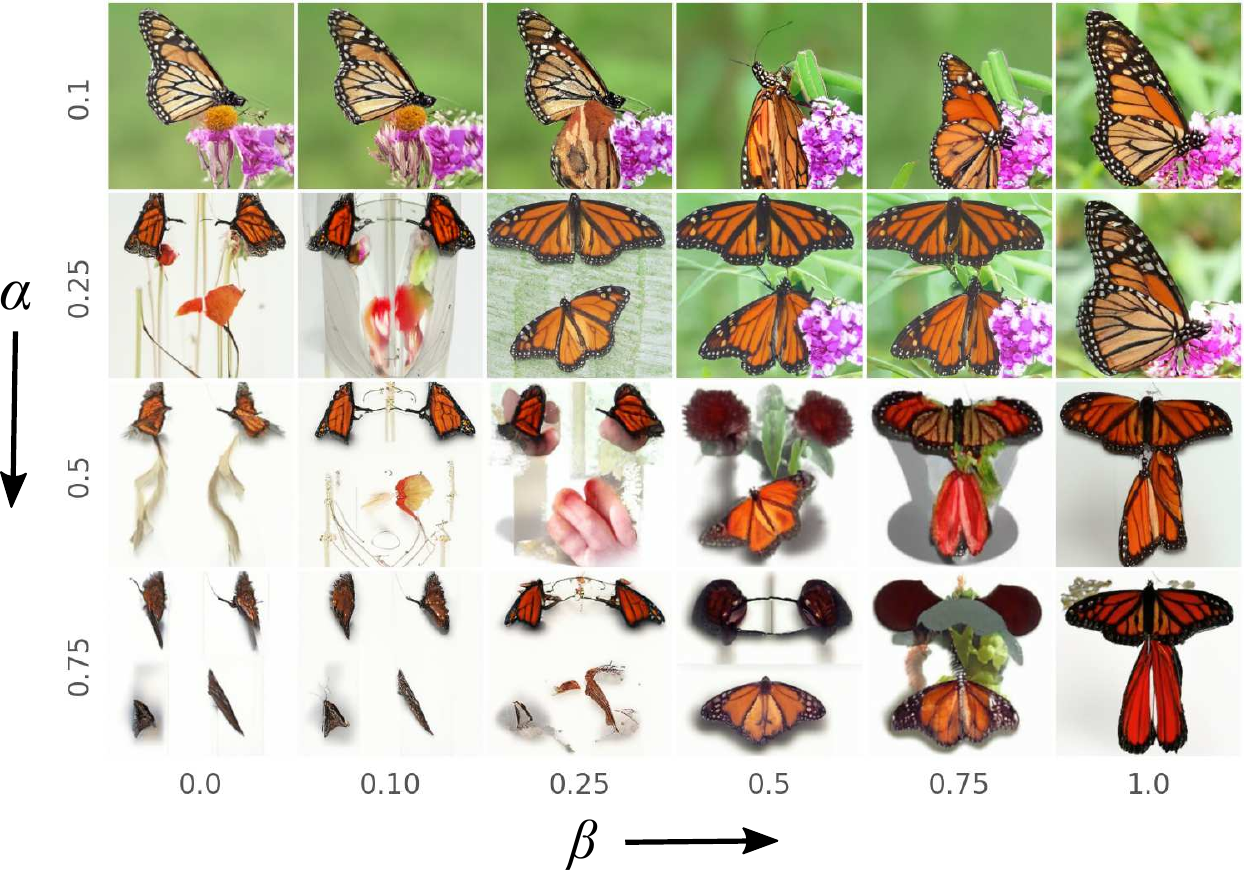}
    \caption{\textbf{Controlling hardness and fidelity.} Effect of increasing $\alpha$ (y-axis) and $\beta$ (x-axis) on synthetic images. Increasing $\alpha$ forces the model to sample from low-density regions while $\beta$ forces the sampling process to stay close to real data manifold. Salient impact of $\beta$ includes improving foreground semantics to correctly represent the class and preserving background information.}
    \label{fig: a_b_grid}
\end{figure}

\vikash{Highlight that we refer to both loss terms as guidance loss function.}

where $\rvz \sim \gN(\mathbf{0}, \mathbf{I})$, $\nabla^*$ refers to normalized gradients, and $\alpha, \beta$ are scaling hyperparameters. To further demonstrate the combined effect of $\alpha$ and $\beta$, we provide synthetic images with a grid search over both hyperparameters in Figure~\ref{fig: a_b_grid}. We also detail our final approach in Algorithm~\ref{alg: final}.

\SetKwComment{Comment}{/* }{ */}
\begin{algorithm}
\caption{Sampling from low-density
regions.}\label{alg: final}
$\mathbf{Input:}~\textnormal{Class label}~(y), \alpha, \beta$\\
$\mathbf{Function:}~Normalize~(\rvu): \textnormal{return} ~\rvu/\norm{\rvu}$ \\
$\rvx_T \sim \gN(\mathbf{0}, \mathbf{I})$\\
\For{$i\gets T$ \KwTo \textnormal{1}}{
    \eIf{$t >$ \textnormal{1}}{
        $\rvz \sim \gN(\mathbf{0}, \mathbf{I})$, $\rvs \gets \mathbf{I}$
        
      }{
        $\rvz \gets \mathbf{0}$, $\rvs \gets \mathbf{0}$
      }
    $\rvu_1 = \alpha\, \mathbf{\Sigma}_{\theta}(\rvx_t, t)\, Normalize\left(\mathbf{\nabla} L_{g_1}\left(\rvx_{t}, y\right)\right)$\\
    $\rvu_2 = \beta\, \mathbf{\Sigma}_{\theta}(\rvx_t, t)\, Normalize\left(\mathbf{\nabla} L_{g_2}\left(\rvx_{t}\right)\right)$\\
    $\rvx_{t-1} = \bm{\mu}_{\theta}(\rvx_t, t) + \mathbf{\Sigma}_{\theta}^{1/2}(\rvx_t, t)\, \rvz  + \rvs (\rvu_1 + \rvu_2)$
}
\Return $\rvx_0$
\end{algorithm}

\section{Experimental results}

\noindent\textbf{Experimental setup.}
We use a U-Net-based architecture with adaptive group normalization for the diffusion model~\cite{dhariwal2021diffBeatGANs}. We consider the encoder from U-Net for the classifier architecture. Both classifier and diffusion model are conditioned on the diffusion process timestep. We consider $T=1000$ for the diffusion process. When sampling, we use $250$ timesteps, as it speeds up the sampling process while incurring negligible cost in the image quality.

\begin{figure}[!b]
    \vspace{-10pt}
    \centering
    \begin{subfigure}[b]{0.45\linewidth}
        \includegraphics[width=\linewidth]{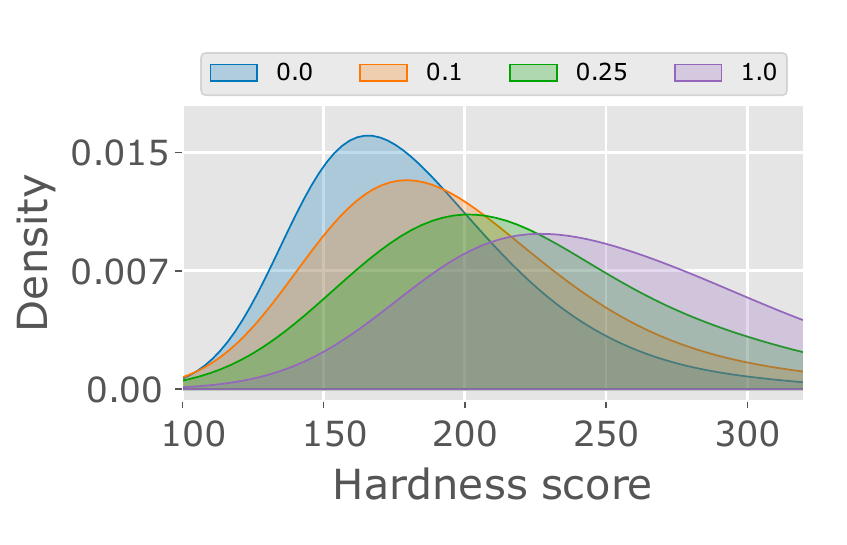}
        \caption{$\alpha$ increases the hardness score.}
        \label{fig: ablate_alpha}
    \end{subfigure}
    \hspace{10pt}
    \begin{subfigure}[b]{0.45\linewidth}
        \includegraphics[width=0.9\linewidth]{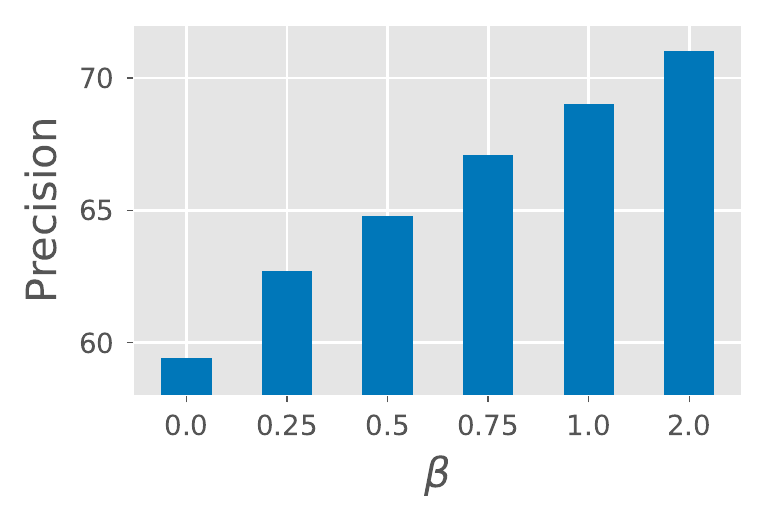}
        \caption{$\beta$ improves the fidelity}
        \label{fig: ablate_beta}
    \end{subfigure}
    \vspace{-5pt}
    \caption{\textbf{Validating effect of hyperparameters.} Quantitative results validating the desired effect of hyperparameters $\alpha$ and $\beta$. }
    \vspace{-15pt}
\end{figure}

\begin{figure*}[!btp]
     \centering
     \begin{subfigure}[t]{0.3\textwidth}
        \raisebox{-\height}{\includegraphics[width=\linewidth]{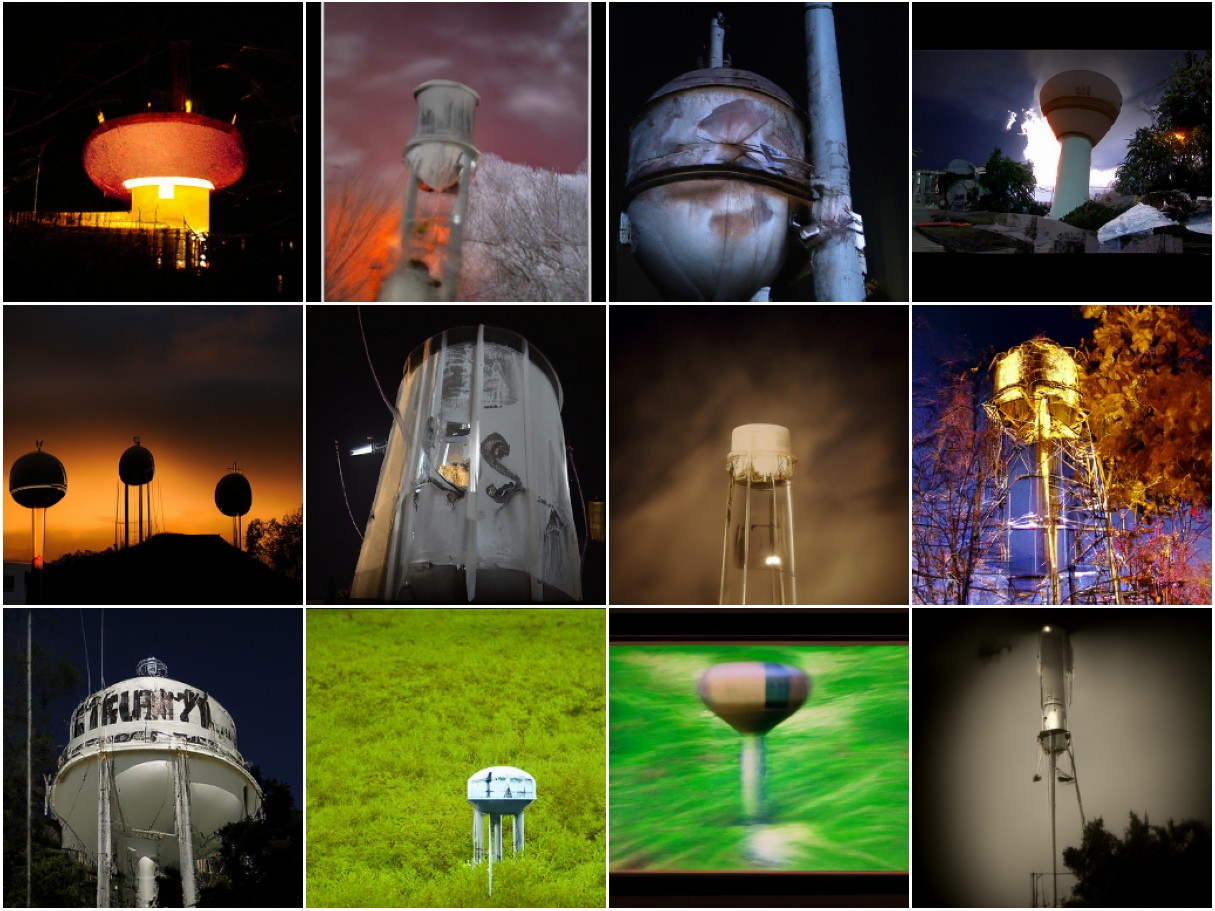}}%
        \vspace{10pt}
        \raisebox{-\height}{\includegraphics[width=\linewidth]{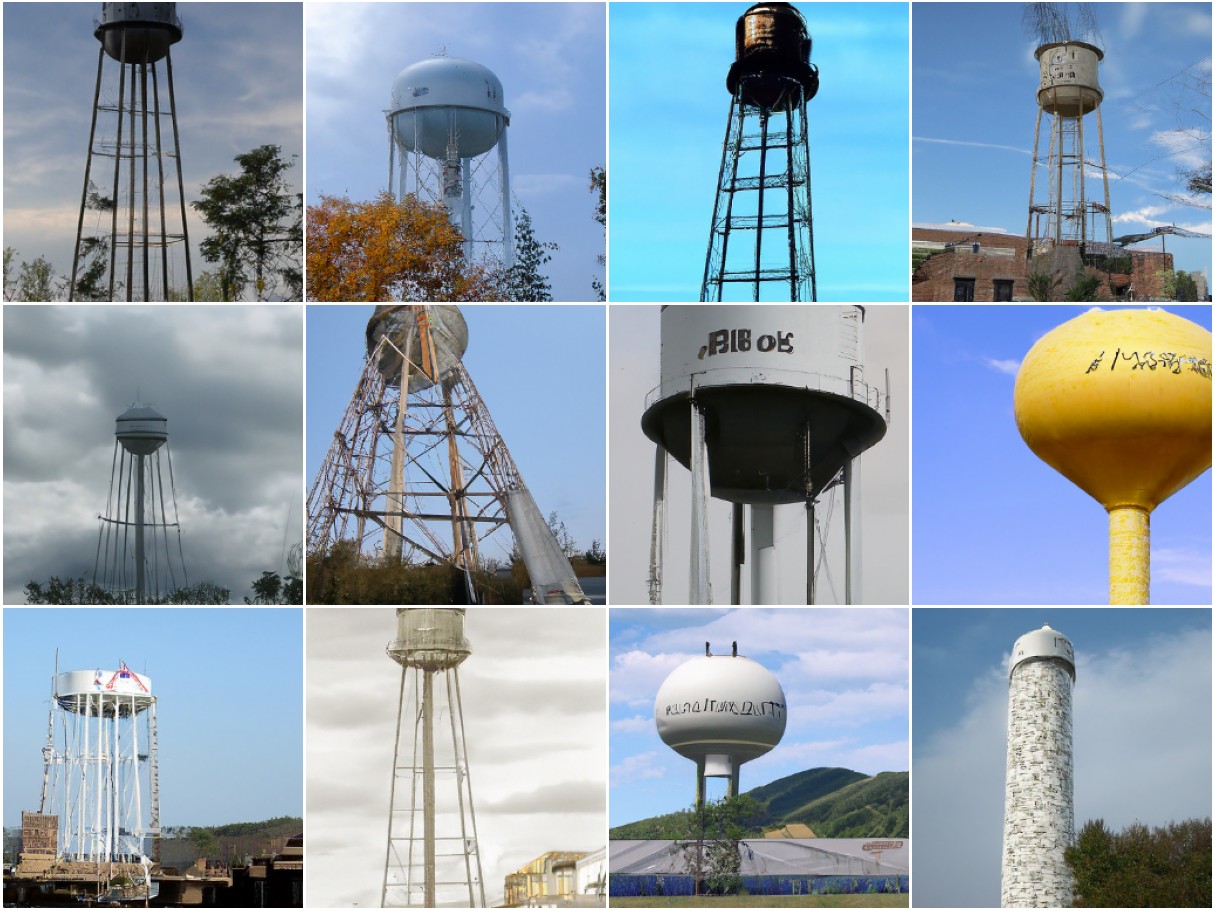}}
        \caption{Water tower}
    \end{subfigure}
    \hfill
    \begin{subfigure}[t]{0.3\textwidth}
        \raisebox{-\height}{\includegraphics[width=\linewidth]{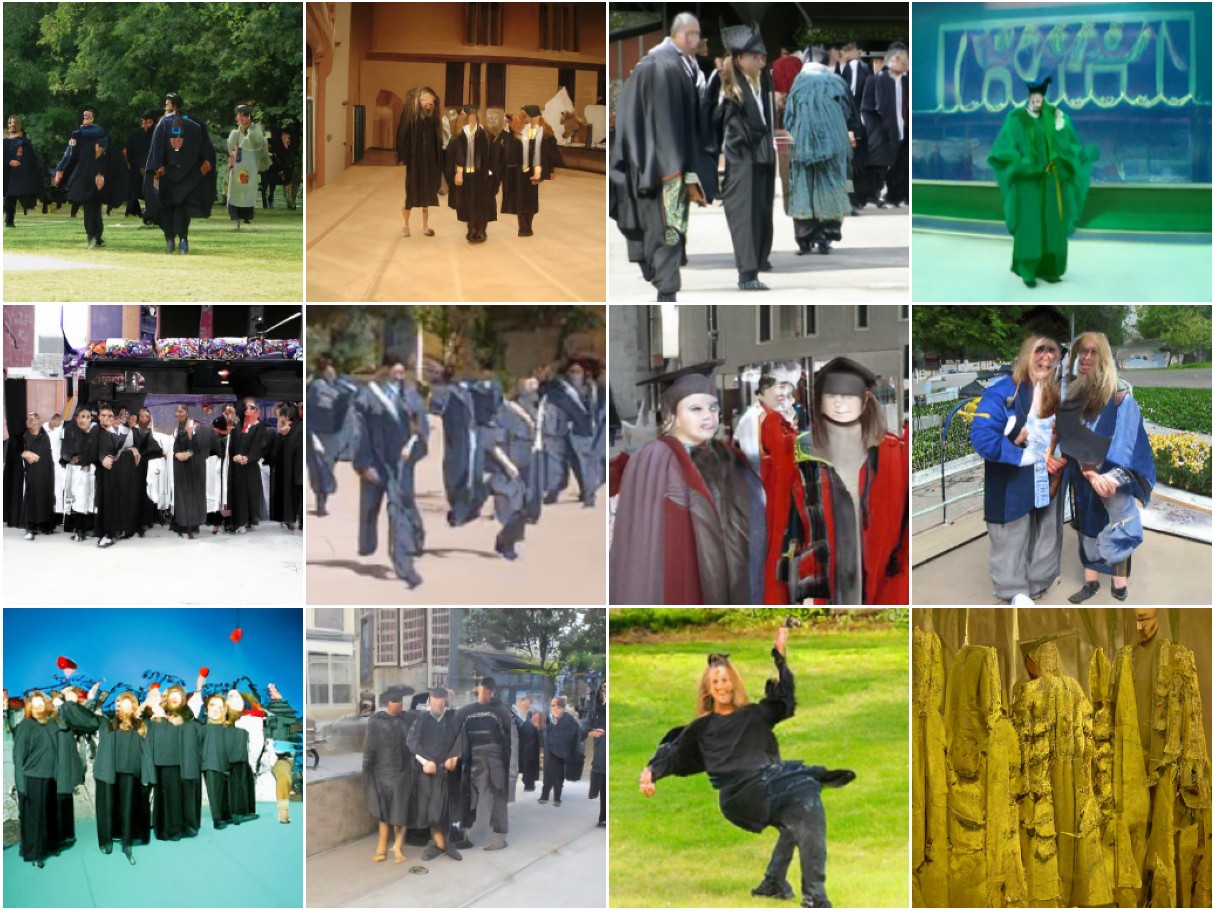}}%
        \vspace{10pt}
        \raisebox{-\height}{\includegraphics[width=\linewidth]{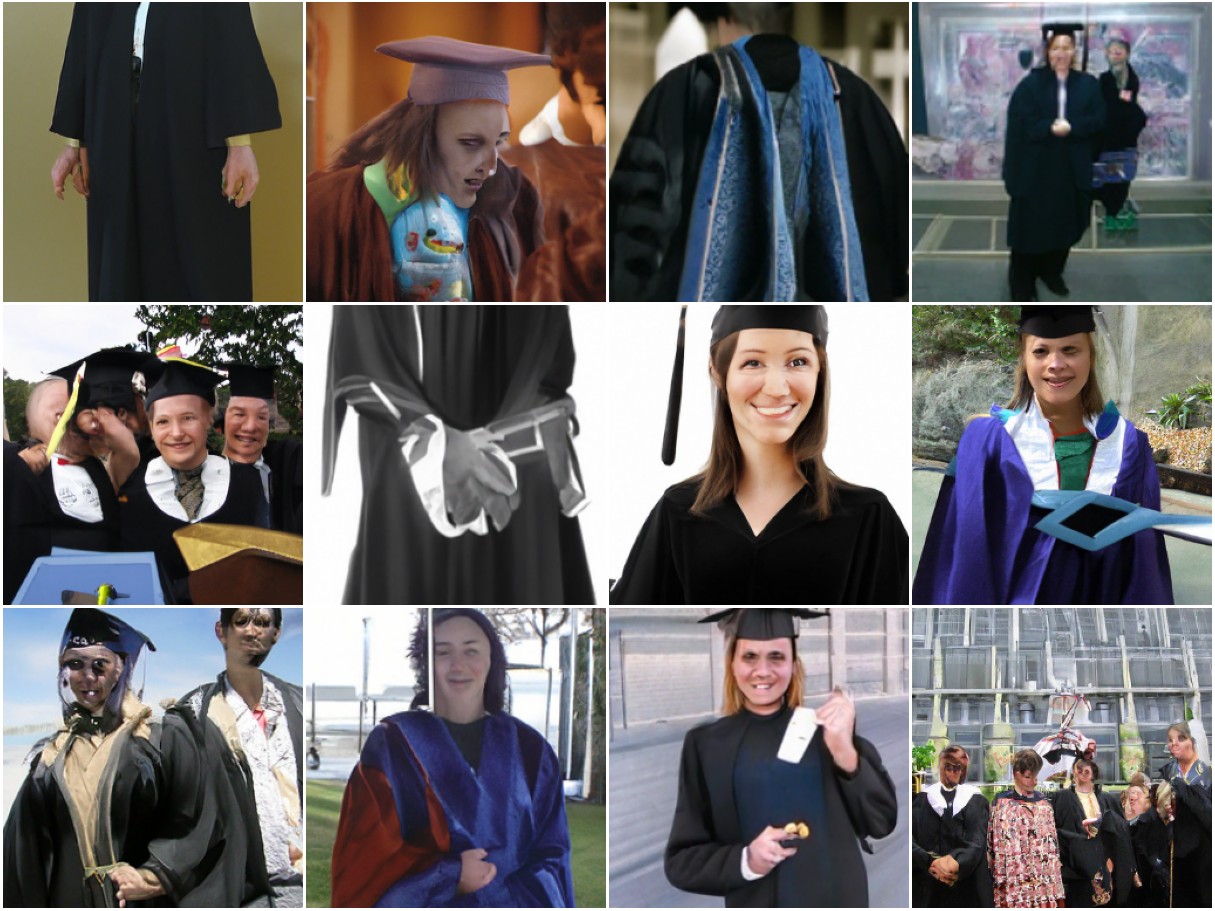}}
        \caption{Academic robe}
    \end{subfigure}
    \hfill
    \begin{subfigure}[t]{0.3\textwidth}
        \raisebox{-\height}{\includegraphics[width=\linewidth]{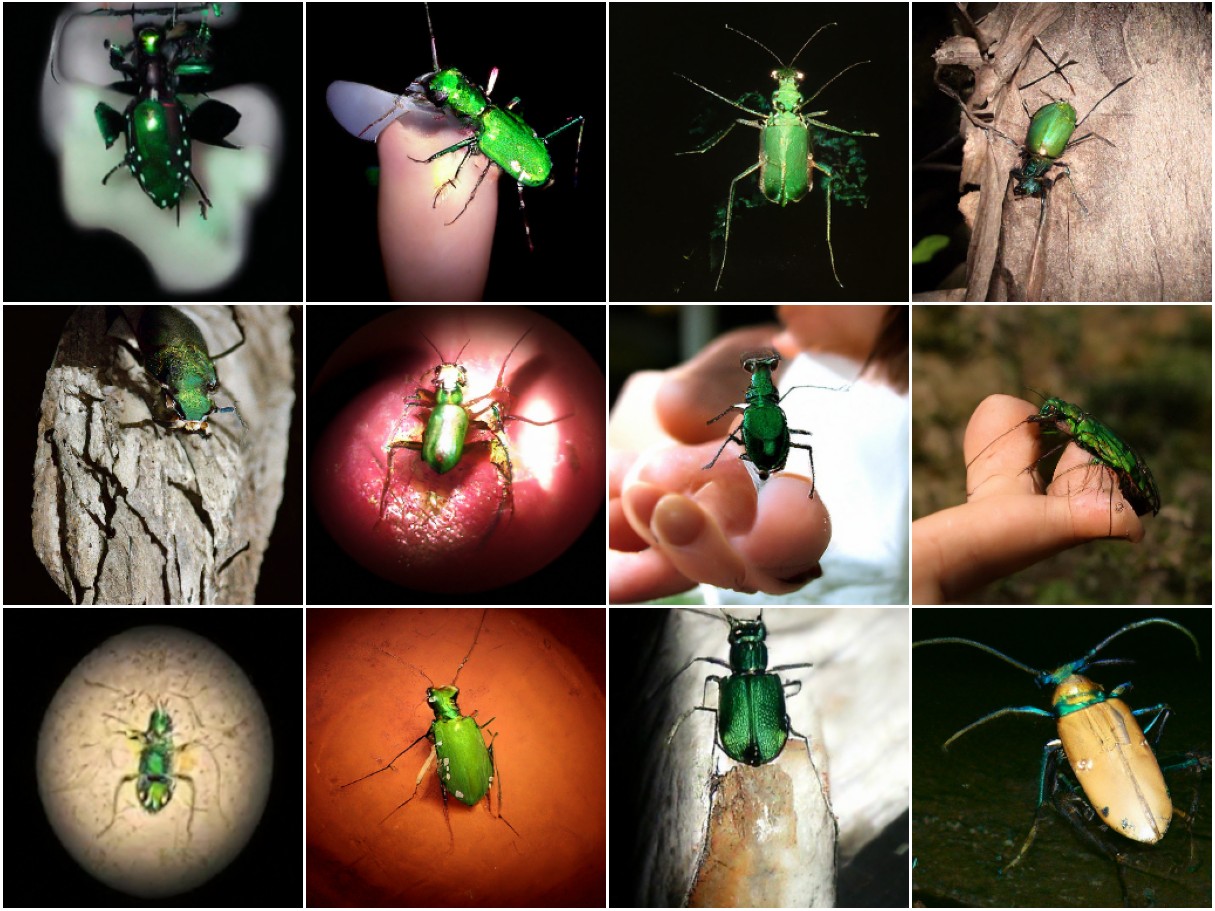}}%
        \vspace{10pt}
        \raisebox{-\height}{\includegraphics[width=\linewidth]{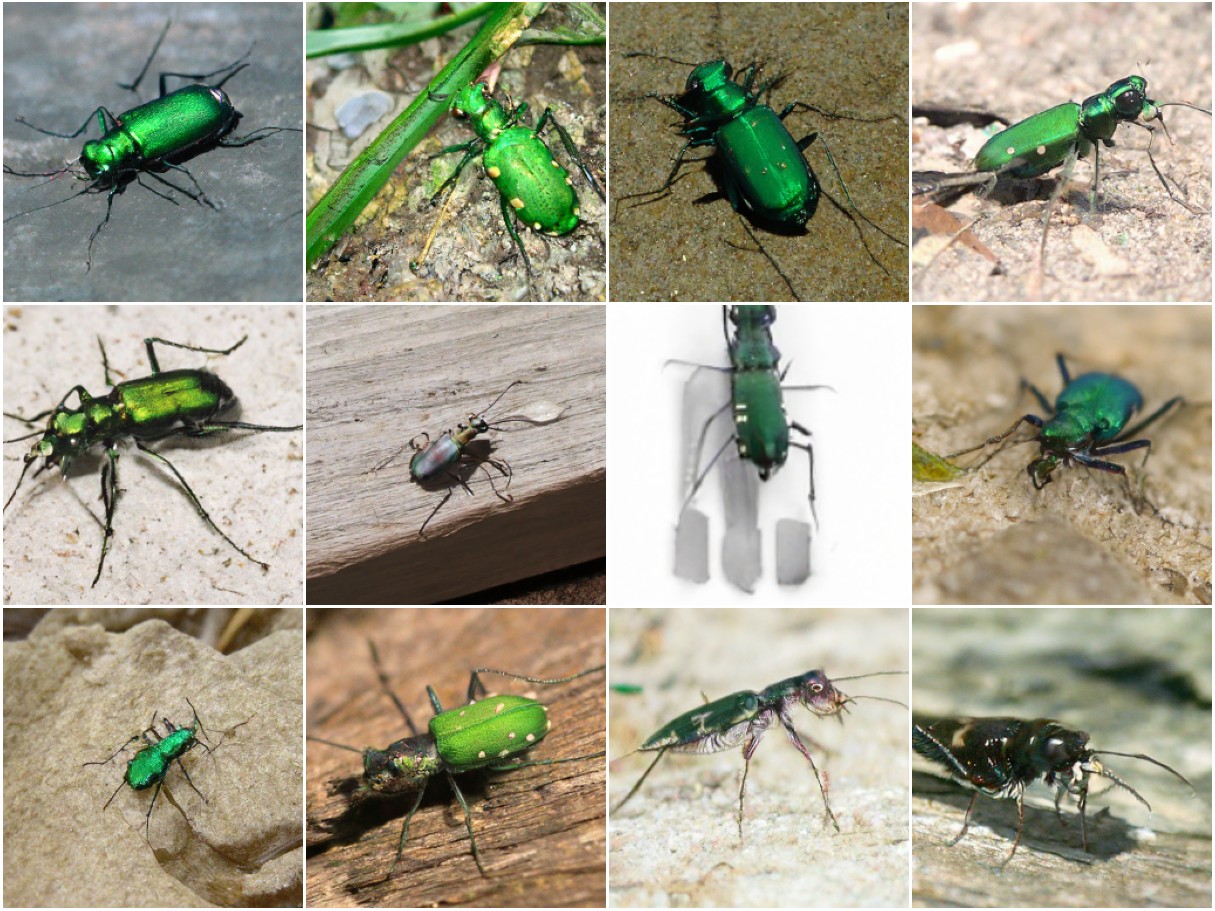}}
        \caption{Tiger beetle}
    \end{subfigure}
    \vspace{-5pt}
    \caption{\textbf{Comparing samples from proposed and baseline sampling process.} We compare synthetic images from our proposed sampling approach (top) with the baseline sampling process (bottom) on the ImageNet dataset. We use identical random seed for both stochastic sampling processes. Therefore, generation of each pair of images among the two approaches starts from the identical latent vectors and the only difference is the additional guidance terms in our approach.}
    \label{fig: add_demo_images_imagenet}
\end{figure*}

\begin{figure*}[!btp]
     \centering
    \begin{subfigure}[t]{0.48\textwidth}
        \raisebox{-\height}{\includegraphics[width=0.48\linewidth]{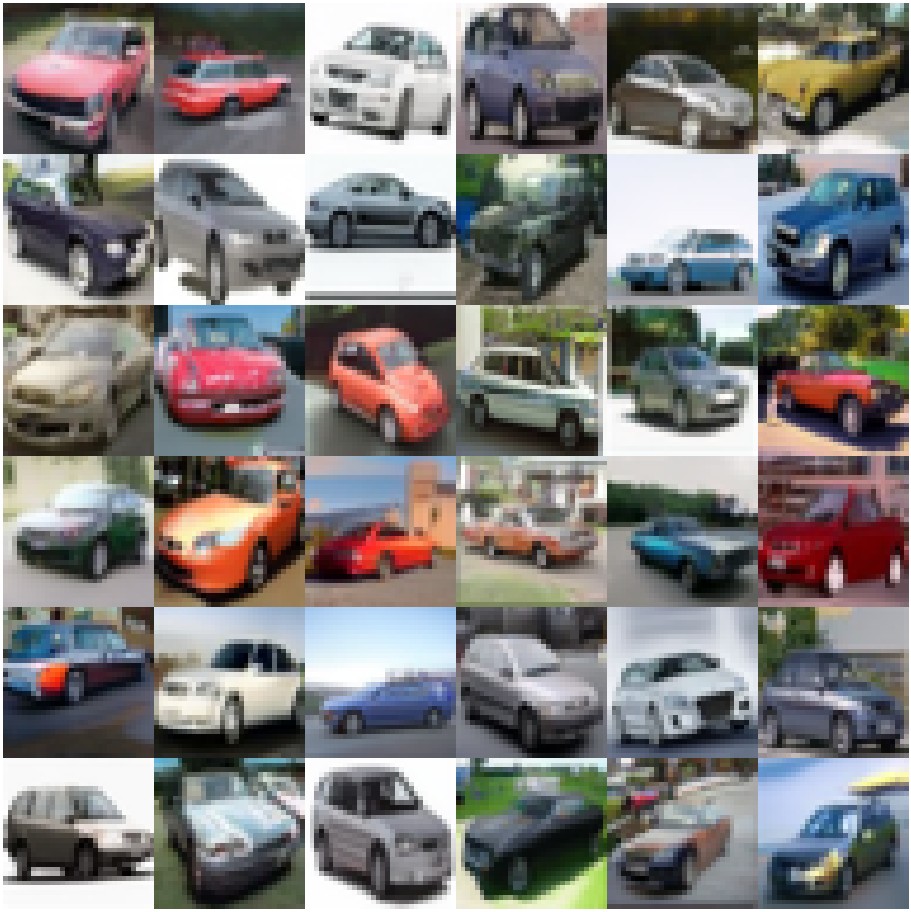}}%
        \hspace{5pt}
        \raisebox{-\height}{\includegraphics[width=0.48\linewidth]{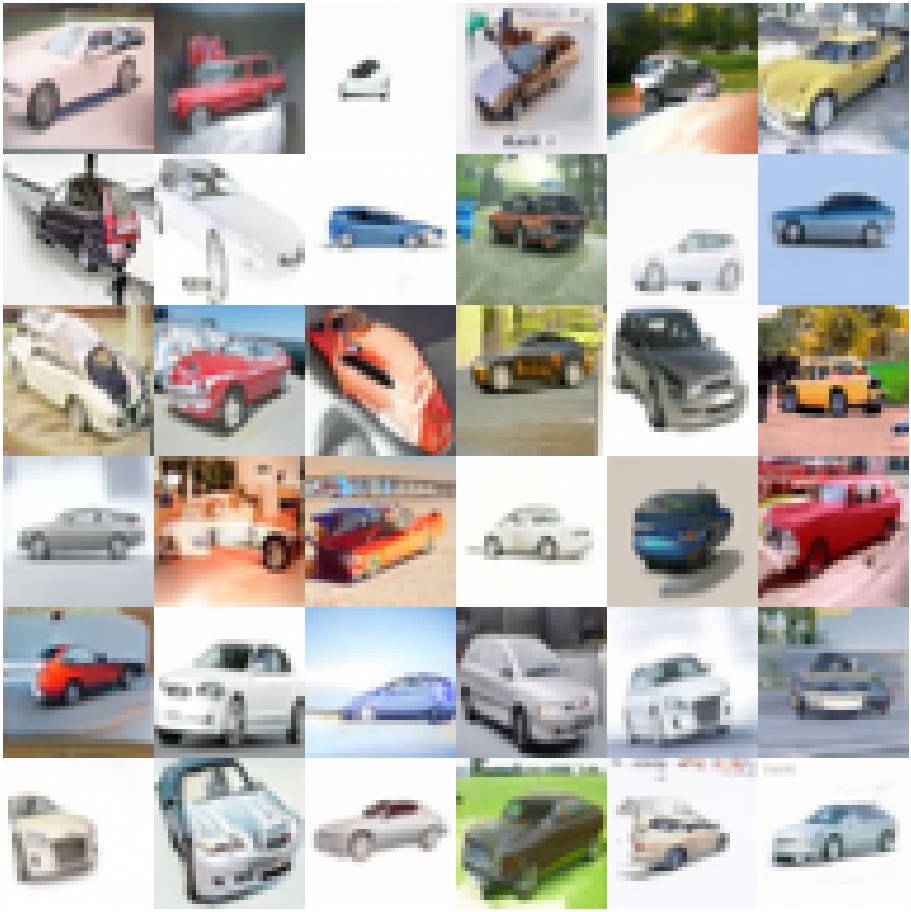}}
        \caption{Automobile}
    \end{subfigure}
    \hspace{10pt}
    \begin{subfigure}[t]{0.48\textwidth}
        \raisebox{-\height}{\includegraphics[width=0.48\linewidth]{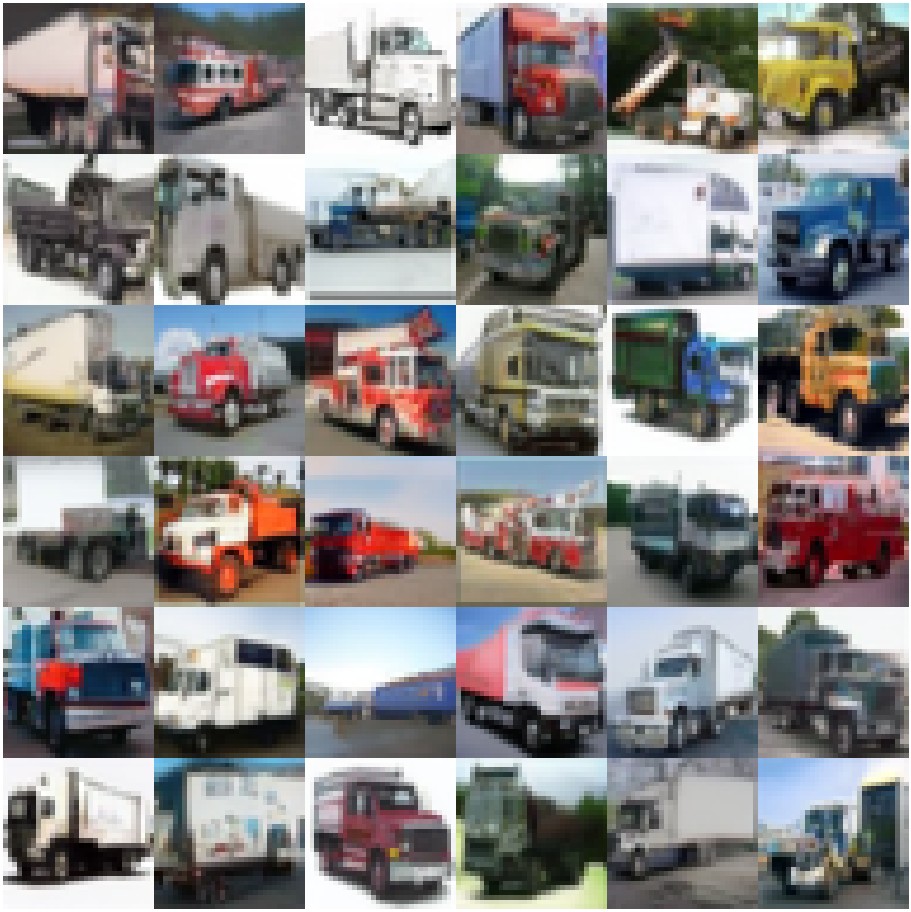}}%
        \hspace{5pt}
        \raisebox{-\height}{\includegraphics[width=0.48\linewidth]{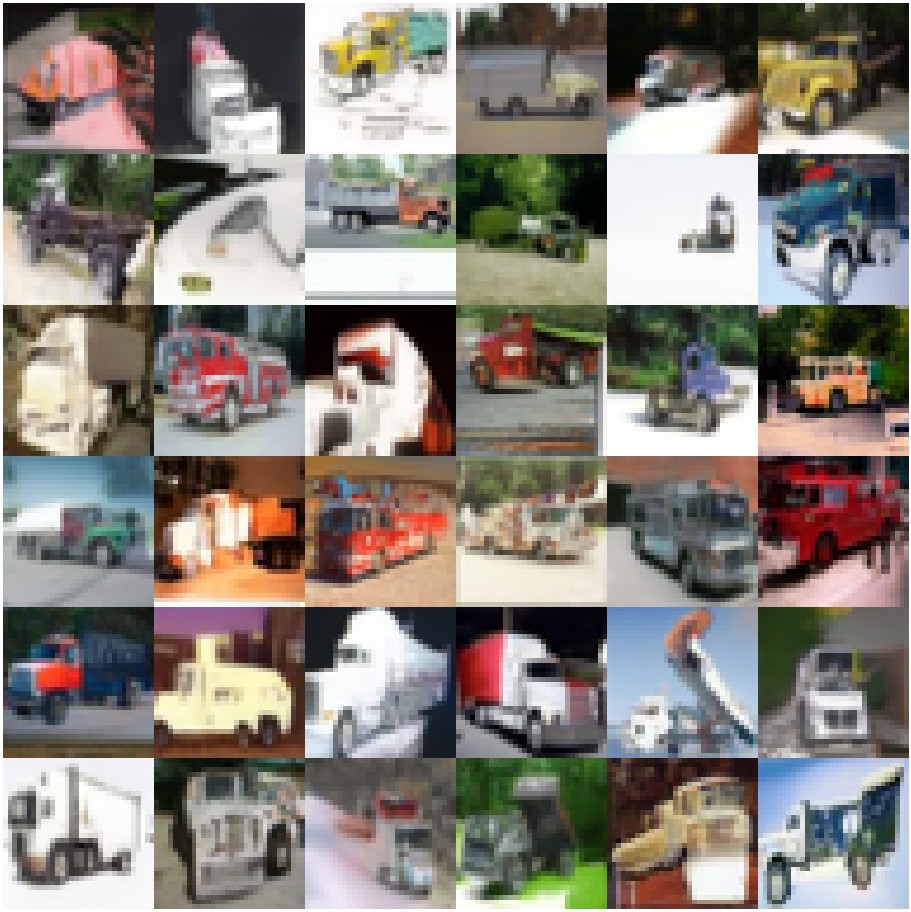}}
        \caption{Truck}
    \end{subfigure}
    \vspace{-5pt}
    \caption{\textbf{Comparison on CIFAR-10 dataset.} We compare synthetic images from the baseline sampling process (left) with our proposed sampling approach (right) on the CIFAR-10 dataset. We use the identical seed for random number generators for both processes.}
    \vspace{-10pt}
    \label{fig: add_demo_images_cifar10}
\end{figure*}

We consider two commonly used image datasets: CIFAR-10~\cite{krizhevsky2009learning} and ImageNet~\cite{deng2009imagenet}. When training the binary discriminator, $H'$, we first uniformly sample synthetic images equal to the size of the training dataset,~\ie, $50$K images for the CIFAR-10 dataset and $1.2$M images for the ImageNet dataset. We conduct a hyperparameter search for $\alpha$ and $\beta$ between $0.01$ and $1.0$. In most analyses, we sample $50$K synthetic images for ImageNet and $10$K synthetic images for the CIFAR-10 dataset, i.e., equal to the size of validation set for each dataset. We provide additional experimental details in Appendix~\ref{app: setup}. 

When sampling we optimize the likelihood estimate,~\ie, hardness score, calculated in the embedding space of the U-Net encoder model. To measure generalization to other representation spaces, we consider multiple other models to calculate hardness scores post sample generation. We present results with the ResNet-50 model in the main paper and the rest in the Appendix~\ref{app: kNNClassifier}.
\vikash{Add details on the cascading model and that we only for the smallest resolution.}

\vikash{Add the controlled experiment to demonstrate the success of our approach.}

\begin{figure*}
    \centering
    \begin{subfigure}[b]{0.3\linewidth}
         \includegraphics[width=\linewidth]{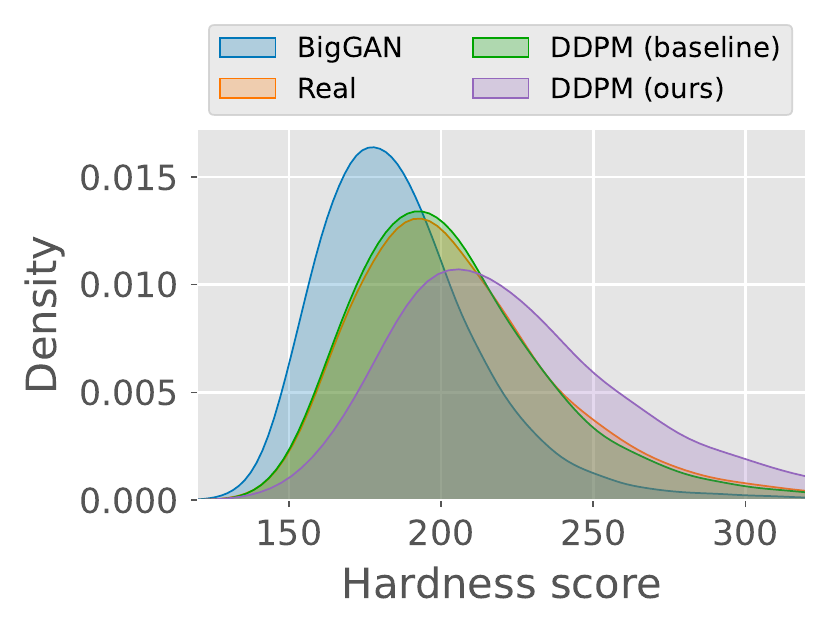}
          \caption{Hardness score}
    \end{subfigure}
    \hspace{10pt}
    \begin{subfigure}[b]{0.3\linewidth}
         \includegraphics[width=0.92\linewidth]{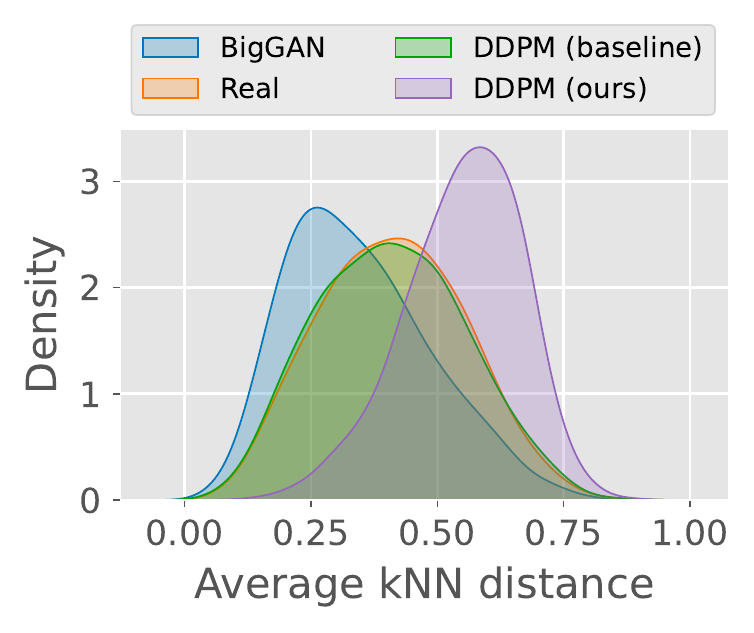}
         \caption{AvgkNN distance}
    \end{subfigure}
    \begin{subfigure}[b]{0.3\linewidth}
         \includegraphics[width=0.97\linewidth]{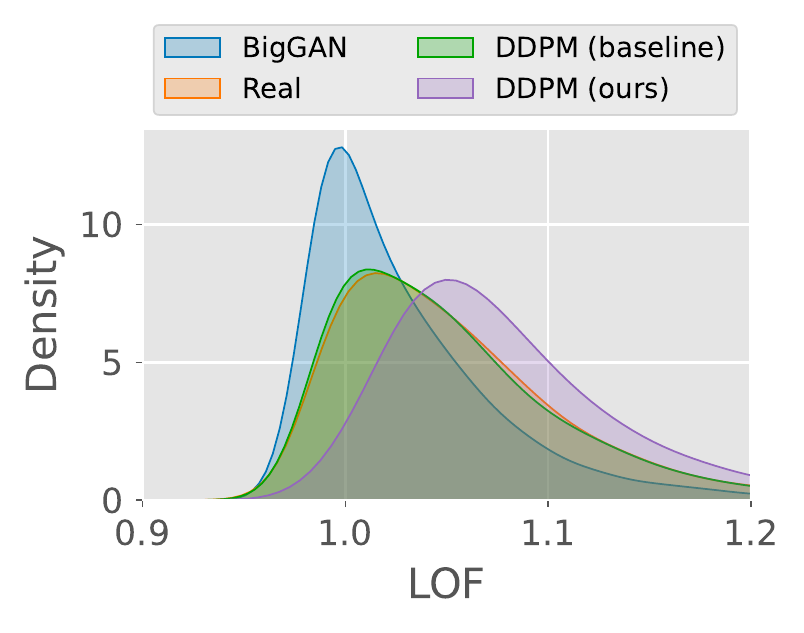}
         \caption{Local outlier factor}
    \end{subfigure}
    \vspace{-5pt}
    \caption{\textbf{Comparing neighborhood density.} We measure the density in the neighborhood of a given set of instances using three different metrics. All three metrics share a common trend: while baseline sampling generates synthetic samples that have similar density distribution as real data, our sampling process generates samples from low density neighborhoods with higher probability.}
    \label{fig: density_comp}
    \vspace{-10pt}
\end{figure*}

\subsection{Generating synthetic data using proposed $\alpha \text{-} \beta$ guided sampling process} \label{sec: alpha_beta_results}
\noindent \textbf{Validating the effect of hyperparameter $\alpha$ and $\beta$.} 
Our sampling process is designed such that we can sample images from the low-density regions by increasing $\alpha$ and improving the fidelity of these images using $\beta$. Our first goal is to validate the desired effect of both hyperparameters.

While using $\beta=0$, we first increase $\alpha$ value from $0$ to $1.0$ and measure the hardness score of sampled images at each value (Figure~\ref{fig: ablate_alpha}). Our results demonstrate that increasing $\alpha$ shifts the hardness score distribution to the right, i.e., higher probability of sampling images that have lower estimated likelihood.

Next we fix $\alpha=0.5$ and increase $\beta$ from zero to two. We use precision~\cite{sajjadi2018PrecisionRecall} to measure the fidelity of synthetic images. It broadly measures the fraction of images that are realistic or equivalently, the coverage of synthetic data by the support of training data distribution. Our results show that increasing $\beta$ does improve the realism of generated synthetic images (Figure~~\ref{fig: ablate_beta}).

Finally, we analyze the joint effect of parameters $\alpha$ and $\beta$. We perform a grid search over both $\alpha$ and $\beta$ and generate images for each pair of values. To avoid the impact of stochasticity, we use the same seed for all runs of the sampling process. We present the sampled images in Figure~\ref{fig: a_b_grid}. 

These visualizations validate our argument that solely increasing $\alpha$ to very high values degrades image fidelity. This is because a higher value of $\alpha$ encourages sampling of low-likelihood images. However, the model can satisfy this constraint by simply generating a poor-quality image. Increasing $\beta$ addresses this issue, in particular on high values of $\alpha$,  where it restores the key attributes of the image thus effectively moving it closer to the data manifold. We find that a $1:1$ ratio between $\alpha$ and $\beta$ strikes a modest trade-off between sample hardness and fidelity and use $\alpha=\beta=0.5$ for further experiments.



\noindent \textbf{Comparing our sampling process with the baseline sampling process.}
 We compare the synthetic images generated from the baseline and our sampling approach in Figure~\ref{fig: add_demo_images_imagenet},~\ref{fig: add_demo_images_cifar10}. We use identical experimental setup, including seeds for random number generators, for both sampling processes thus leaving guidance terms to be the only factor impacting final images. Images from our approach are visually distinguishable from the baseline approach since the diffusion model introduces significant changes in the foreground object semantics and background to satisfy the constraints on hardness and fidelity. We provide additional visualizations in Appendix~\ref{app: sampleComp}. 

\subsection{Quantitative comparison of neighborhood density}
To validate that our sampling process does generate data from low-density regions, we quantitatively compare the manifold density in the neighborhood of synthetic images with different baselines.

\noindent \textbf{Metrics to measure neighborhood density.} We use hardness score as the first validation metric since we maximize it in the sampling process. However, our sampling process might maximize hardness score without actually moving the sampling process to low-density regions. Thus we consider two additional metrics, namely Average nearest neighbor (AvgkNN) and local outlier factor (LOF)~\cite{breunig2000lof} to further validate the success of our approach. AvgkNN measures density using proximity to nearest neighbors. We choose five nearest neighbors, which is a common choice~\cite{devries2020instanceSelection}. In contrast, the local outlier factor improves on the nearest-neighbor distance metric to compare density around a given sample to density around its neighbors. Higher values of the local outlier factor indicate the sample lies in a much lower density region than its neighbors. We calculate all distances in the feature space of a ResNet50 network which is pre-trained on the ImageNet dataset. We ablate on the choice of feature extractor in Appendix~\ref{app: kNNClassifier} and show that our conclusions don't change with this choice. For this analysis, we sample $50$K synthetic images using recommended values of $\alpha$ and $\beta$ from Section~\ref{sec: alpha_beta_results}. We compare our approach with three baselines 1) BigGAN-deep 2) Real images from the ImageNet validation set and 3) synthetic images generated using baseline sampling from the DDPM model. We present our results in Figure~\ref{fig: density_comp}. 

\noindent \textbf{All three metrics validate the success of our approach.} 
Under all three metrics, our sampling process has a higher probability of generating synthetic images from low-density neighborhoods. It also validates the claims that the sample density in real data itself follows a long-tail distribution and an unmodified sampling process, i.e., baseline sampling process, from diffusion models closely approximates this distribution. In comparison, BigGAN samples are predominantly from low-density regions. Among the three metrics, the difference between our approach and baseline is most significant in AvgkNN distance. When ablating on the choice of the guidance loss function, we find that under sufficient hyper-parameter ablation, one can obtain equivalent results when optimizing likelihood in embedding space or softmax probabilities after the logit layer (Appendix~\ref{app: tempEffect}).

\vikash{Add some details on the number of images used for each dataset. Do this for other sections too as such details are often missing.}



\begin{table}[!b]
    \centering
    \vspace{-20pt}
    \caption{\textbf{Reduction in sampling cost.} Comparing the sample generation time of our method with uniform sampling. Each entry represents the time taken (in days) to generate $5$K $256\times256$ resolution synthetic images from the corresponding hardness score range on a single A100 GPU.}
    \label{tab: cost_comp}
    \vspace{-5pt}
    \begin{tabular}{cccc}  \toprule
     Score-range & $200-240$ & $240-280$ & $280-320$ \\ \midrule
     Baseline & $1.99$ & $5.74$ & $16.79$ \\ 
     Ours & $1.88$ \scriptsize{($\times \textbf{1.1}$)}  & $2.03$ \scriptsize{($\times \textbf{2.8}$)} & $2.78$ \scriptsize{($\times \textbf{6.0}$)} \\ \bottomrule
    \end{tabular}
\end{table}

\noindent \textbf{Equivalent reduction in computational cost.} Assume that we want to sample images from low-density neighborhoods,~\ie, the hardness score of each synthetic sample is greater than a threshold. A naive rejection sampling-based approach is to sample images uniformly at random and reject images that do not satisfy the criterion. However, due to the long-tail nature of sample density, the likelihood of a sample being from low-density regions is low, thus we would need to reject many samples to curate desired samples. Due to the iterative nature of the sampling process, generating synthetic data from diffusion models is computationally expensive, thus making rejection sampling a highly computationally costly approach (Table~\ref{tab: cost_comp}). Our approach does not depend on rejection sampling, thus it is up to $2-6\times$ faster than the former approach (Table~\ref{tab: cost_comp}). 
\section{Is our sampling process generating memorized samples from training data?}
Since a limited number of samples are available from low-density regions in our long-tailed datasets, the generative model might memorize these samples and fails to generate novel samples from these regions. Therefore we conduct a rigorous analysis to identify whether our sampling process is exploiting any memorization that might be happening in diffusion models.


\begin{figure}[!b]
    \centering
    \vspace{-10pt}
    \includegraphics[width=\linewidth]{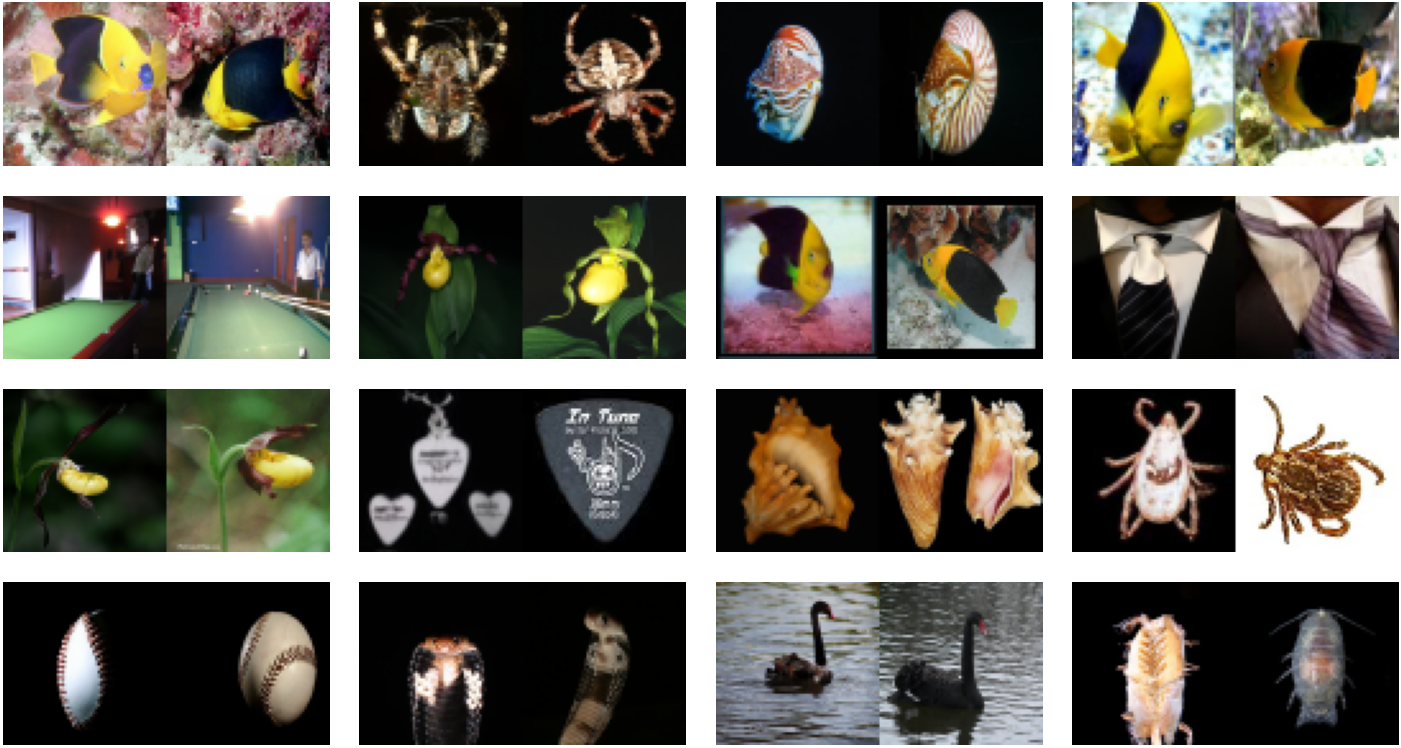}
    \vspace{-10pt}
    \caption{\textbf{Is low-density synthetic data being memorized?} Pairs of synthetic and real images with smallest euclidean distance in the feature space. In each pair, left and right image correspond to synthetic and real image, respectively. Our search space for these examples includes all pairs of 50K synthetic images and 1.2M real images. While the synthetic images share multiple attributes with the nearest real image, they are not identical to the real images.}
    \label{fig: nn_pairs}
    \vspace{-20pt}
\end{figure}

\begin{figure}[!ht]
    \centering
    \includegraphics[width=\linewidth]{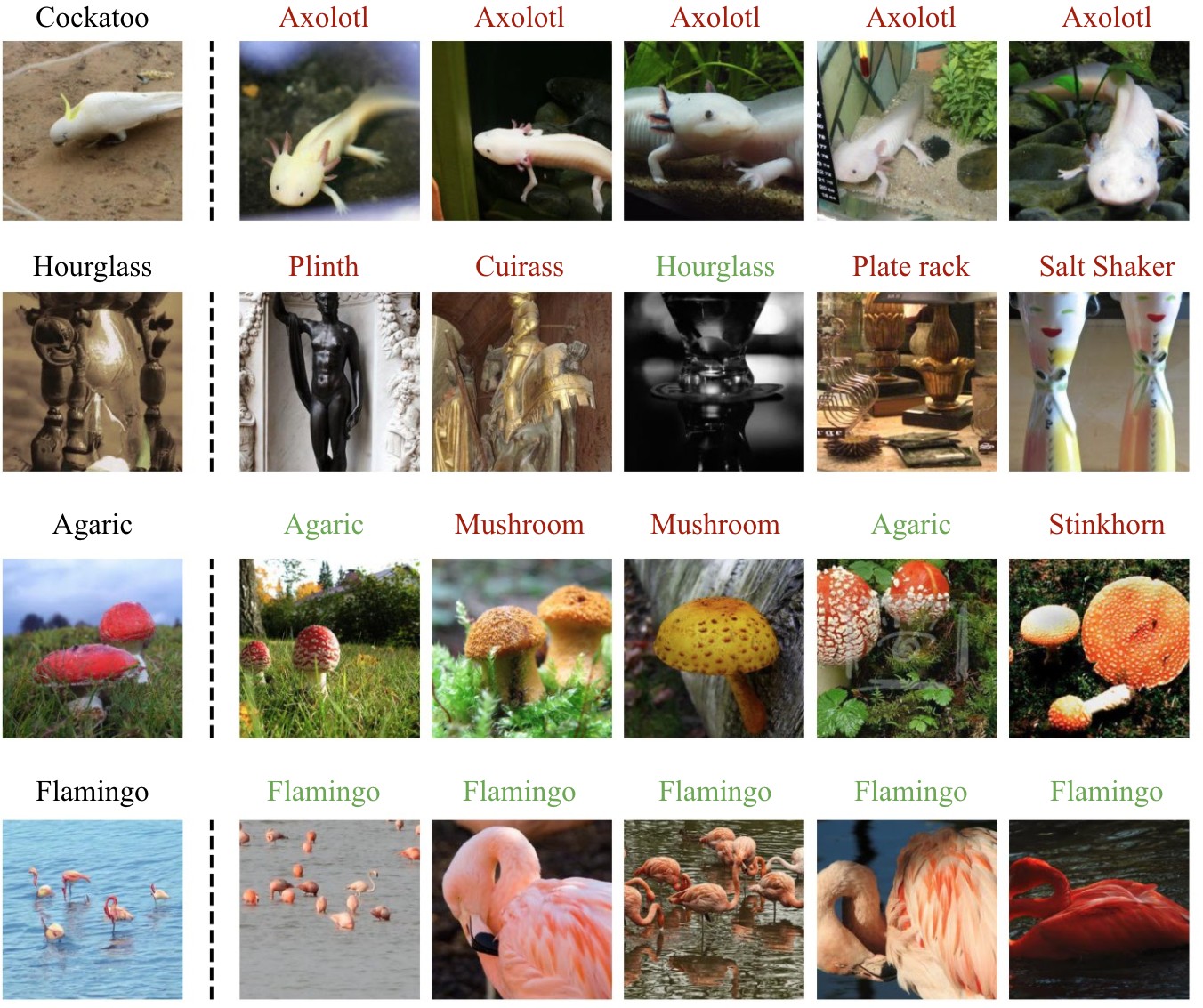}
    \vspace{-10pt}
    \caption{\textbf{Is low-density synthetic data novel?} 
    For each synthetic image, we analyze the class label of five nearest neighbors from real data. While each synthetic image has high fidelity and correctly represents the class, it often lies closer to samples of other classes in feature space. Even when the class label is not different, the synthetic image differs significantly from closest real samples. \vikash{Ablate the choice of embedding network to avoid the critique that it's because of this network. Include bus in sky example.}\label{fig: nn_ours}}
    \vspace{-30pt}
\end{figure}

\noindent \textbf{Analyzing nearest-neighbor distance.} We argue that if training data is being memorized, synthetic images will be substantially similar to training data. We measure this similarity by euclidean distance in the embedding space of well trained image classifiers. Thus, if a synthetic image is simply memorized from training data, its nearest-neighbor distance from real data will be very small. 

We sample $50$K images using our sampling approach and measure their nearest neighbor distance from $1.2$M real images in the training set of the ImageNet dataset. We compare these values with the nearest neighbor distance for real data in the validation set. If our approach has memorized training samples, its nearest neighbor distance should be much smaller than real samples. However, the average distance for our samples is $0.42$, much higher than $0.29$ for real samples. It supports our hypothesis that our sampling process is not simply generating memorized training samples. 

\noindent \textbf{Analyzing synthetic-real data pairs for signs of memorization.} Moving beyond comparing distribution statistics, now we analyze individual samples for signs of memorization. In particular, our goal is to manually analyze synthetic images and their closest neighbors for signs of memorization. Even more, we want to analyze pairs that would have the highest likelihood of being memorized, i.e., synthetic samples which are closest to real data. Across all $60$B pairs ($50\text{K}\times1.2$M)  of synthetic and real images, we manually analyze the top-500 pairs with the smallest pairwise distance. 

We observe that while images in these pairs share multiple attributes, such as object shape, texture, and identity, they are not being memorized. Instead, they are some semantic variation of the real images, highlighting that the diffusion model learned the data manifold instead of memorizing these samples. We present the top twelve pairs in figure~\ref{fig: nn_pairs} and the rest of them in Appendix~\ref{app: kNNPairs}.  

\noindent \textbf{Novel samples from low-density regions.} To validate that our sampling process is indeed generating novel images from low-density regions, we also consider the class label of its nearest neighbors from real data. In multiple cases, we find that the nearest neighbors have different class label than the synthetic sample. We provide few such examples in Figure~\ref{fig: nn_ours}. This phenomenon likely arises due to poorly learned representation by the embedding extractor in low-density regions, primarily due to the scarcity of training samples in these regions.





\section{Discussion}
We present an improved version of the sampling process in diffusion-based generative models that enables sampling from low-density neighborhoods of the data manifold. We achieve this by guiding the sampling process using two additional classifiers at each timestep. Our sampling process successfully generates novel samples from low-density regions. Our work also identifies another compelling advantage of diffusion models. Despite being trained on a small number of samples from low-density regions, diffusion models successfully interpolate in these regions, i.e., don't memorize the training data from these regions. 

\begin{figure}[!tb]
    \centering
    \begin{subfigure}[b]{\linewidth}
         \centering
         \includegraphics[width=\linewidth]{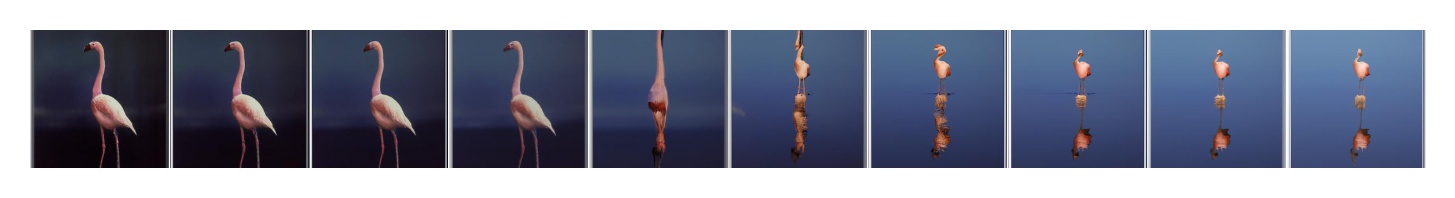}
         \vspace{-10pt}
    \end{subfigure}
    \begin{subfigure}[b]{\linewidth}
         \centering
         \vspace{-10pt}
         \includegraphics[width=\linewidth]{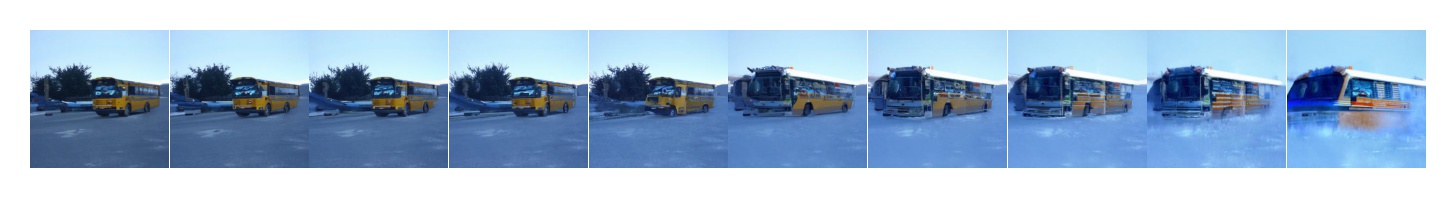}
         \vspace{-10pt}
    \end{subfigure}
    \begin{subfigure}[b]{\linewidth}
         \centering
         \vspace{-10pt}
         \includegraphics[width=\linewidth]{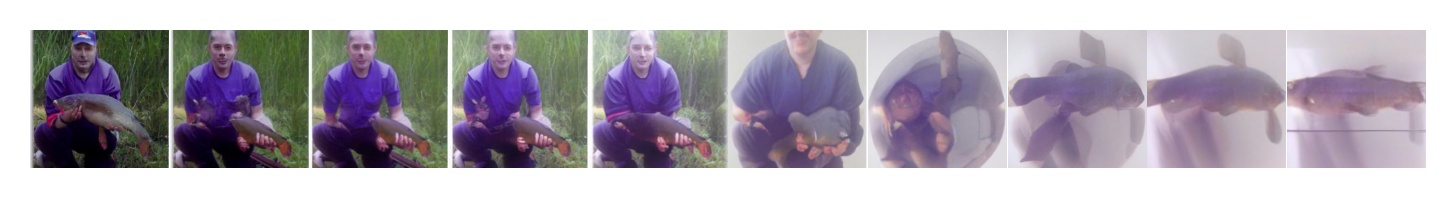}
         \vspace{-5pt}
    \end{subfigure}
    \vspace{-30pt}
    \caption{\textbf{Progressive sampling.} We incrementally increase $\alpha$ across different runs of the sampling process. It highlights how the guiding loss progressively moves the synthetic images to low density regions.}
    \vspace{-20pt}
    \label{fig: progress_sampling}
\end{figure}

We analyze the impact of our guiding loss by juxtaposing samples from baseline and our sampling process (Figure~\ref{fig: add_demo_images_imagenet},~\ref{fig: add_demo_images_cifar10}). These results demonstrate that the generative model exploits novel transformations in response to guiding loss objectives. We further analyze this effect, by progressively increasing $\alpha$ while keeping all other parameters fixed (Figure~\ref{fig: progress_sampling}). Higher values of $\alpha$ forces the model to generate low-likelihood samples. We find that the network sometimes exploits transformations such as photometric changes, zoom, viewpoint, and switching the background to reduce the likelihood of synthetic samples.

The sampling process in diffusion models iterates for hundreds of steps to generate a single sample. This challenge is often solved using a fast sampling process, which trades off sample quality for speed~\cite{song2020ddim, jolicoeur2021gottaGoFast}. To demonstrate that our approach can also integrate with fast sampling techniques, we integrate our modified sampling process with the fast sampling approach from Song et al.~\cite{song2020ddim}. We find no strikingly different trade-off between fidelity and sampling steps for the baseline and our approach (Appendix~\ref{app: ddim}). At a very low number of sampling steps, such as ten, both approaches struggle to generate high-quality images. However, with increasing the number of timesteps, the fidelity of both baseline and our approach quickly improves.

\section{Limitations and broader impact}


We guide the sampling process by navigating the data manifold through the feature space of image classifiers. While proximity in feature space of deep neural networks aligns with human perception~\cite{zhang2018Lpips}, deep neural networks are also well known to be biased towards certain attributes, such as texture~\cite{geirhos2018CnnBiasTexture} and background~\cite{sehwag2020BackgroundCheck, xiao2021NoiseorSignal}. Our sampling process can exploit these biases, such as by simply removing the background, to induce a large change in the likelihood in feature space. We also conduct an examination to investigate signs of memorization and whether our sampling process is exploiting them. While we didn't observe any memorization on the ImageNet dataset, diffusion models might memorize samples on even more complex and non-curated datasets than ImageNet. In event of such memorization, our sampling process might exploit it. 


Deep neural networks often struggle to generalize to novel and rarely observed samples from the distribution~\cite{hendrycks2021naturalAdv, koh2021WILDS}. We believe that our work can further assist in improving the distributional robustness of these networks. Our sampling process also reveals that diffusion models successfully generalize to low-density regions of data manifold which further strengthens the argument that these models hold the potential to provide tremendous benefits in representation learning~\cite{sehwag2021ProxyDist, gowal2021GeneratedData}.





\section*{Acknowledgements}
We would like to thank Michal Drozdzal for all the insightful discussions on the project. This work was done during a summer research internship at Meta AI. 

\bibliography{ref}
\bibliographystyle{ieee_fullname}


\appendix
\section{Experimental setup and common design choices}

\subsection{Additional details on experimental setup} \label{app: setup}
We conduct all our analyses with images of the default resolution, i.e., $224$ or $299$, on ImageNet models. Here we generate high-resolution images using the cascaded diffusion approach from Dhariwal et al.~\cite{dhariwal2021diffBeatGANs}. We first generate $64\times64$ size images using the first diffusion model and then upscale them to $256\times256$ resolution using the second diffusion model.

For feature extraction purposes, we use pretrained networks from the Timm~\cite{rw2019timm} library. We extract features from the last convolutional layer for all networks. We consider five neighbors for AvgkNN computation and twenty neighbors for the local outlier factor. We use the implementation from PyOD~\cite{zhao2019pyod} to calculate the local outlier factor. In our sampling process, we compute the hardness score at each time step. To calculate the hardness score, we first extract training data features at each timestep. Since the reverse process starts from white noise, we find that features from deep neural networks have extremely small variance at the start of reverse process. This makes the hardness score, thus gradients of the guidance loss, quite unstable at the start of the reverse process. We circumvent this issue by using an identity precision matrix.  We use PyTorch~\cite{pytorch2019} with an Nvidia A100 GPU cluster for our experiments. 

\subsection{Limitations of likelihood estimate from the diffusion model}  \label{app: diffLikelihood}
\begin{figure}[!htb]
    \centering
    \includegraphics[width=0.95\linewidth]{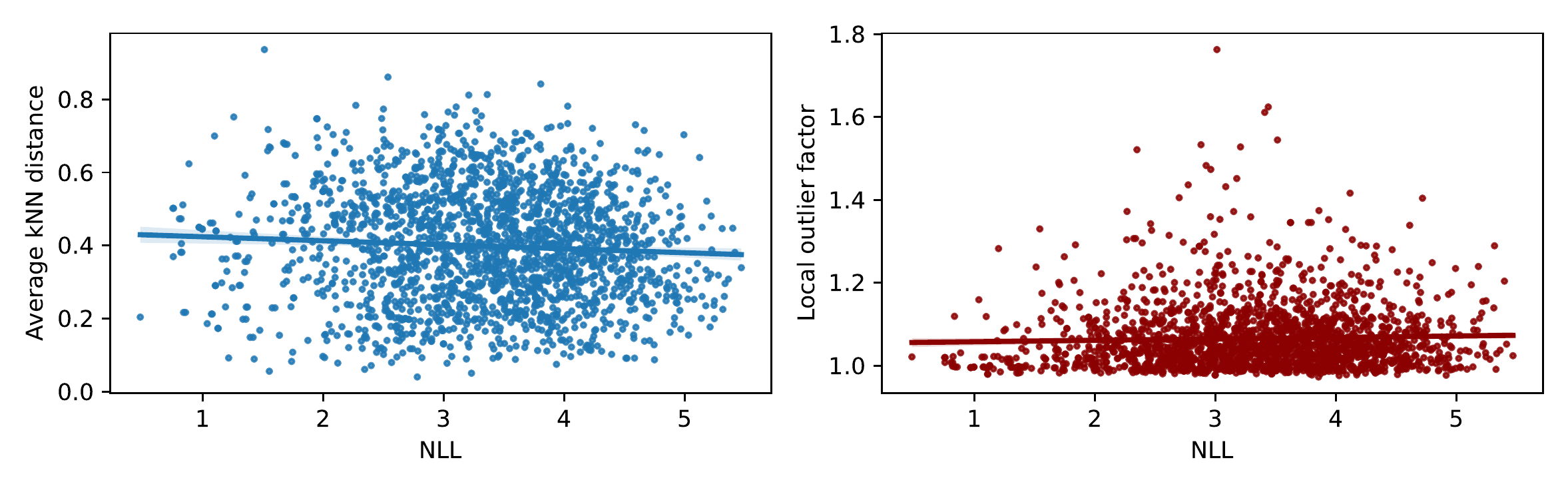}
    \caption{\textbf{Is NLL an effective measure of neighborhood density?} We compare the negative log-likelihood (NLL) estimates from the diffusion model with other commonly used metrics to measure data density. We find that NLL is poorly correlated with both of these metrics. Since NLL is computationally expensive to calculate for each image, we use $2$K random images from the validation set of the ImageNet dataset for this analysis.}
    \label{fig: nll}
\end{figure}

It is straightforward to obtain an estimate of the likelihood of a given sample using the diffusion model. When choosing a metric to identify low-density regions, it is natural to ask whether the likelihood estimates from diffusion models can serve as this metric. To answer this question we calculate the negative log-likelihood (NLL) of real images from the validation set of the ImageNet dataset. We compare NLL with two commonly used metrics to measure the density of neighborhoods (Figure~\ref{fig: nll}). We find that NLL shows poor correlation with both metrics, suggesting that it is not an effective predictor of neighborhood density. 

\begin{figure}[h]
    \centering
    \begin{subfigure}[b]{0.49\linewidth}
        \includegraphics[width=\linewidth]{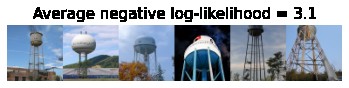}
        \includegraphics[width=\linewidth]{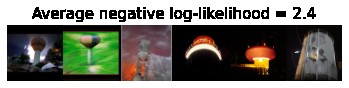}
    \end{subfigure}
    \hfil
    \begin{subfigure}[b]{0.49\linewidth}
        \includegraphics[width=\linewidth]{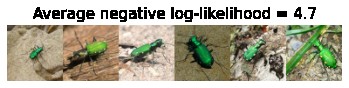}
        \includegraphics[width=\linewidth]{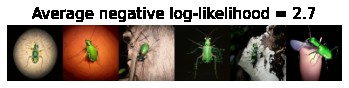}
    \end{subfigure}
    \caption{\textbf{Qualitative comparison.} \textit{Top} row (baseline sampling) vs \textit{Bottom} row (our sampling). Flow-based model surprisingly assigns much higher likelihood to our novel instances (lower value is higher).}
    \label{fig: exact_nll}
\end{figure}

\bluetext{
\noindent \textbf{Limitation of exact likelihood scores.} While diffusion models only provide an approximate likelihood score, one can obtain exact likelihood score from autoregressive or flow-based models~\cite{salimans2017pixelcnn++, grcic2021denselyNormFlow}. We find that the aforementioned limitation of likelihood scores also also extend to \textit{exact} likelihood values. We use DenseFlow~\cite{grcic2021denselyNormFlow}, which provides \textit{state-of-the-art} likelihood evaluation on ImageNet.

Surprisingly, the model assigns very high likelihood values to our low-density images (Table~\ref{tab: exact_nll}), even higher than highly photorealistic BigGAN images. We find that this observation is not limited to our synthetic samples, but a more \textit{fundamental} characteristic of likelihood scores. To highlight it, we consider low-density real images that are poorly represented in training dataset, such such as sketches~\cite{wang2019learningIMSketch}, renditions~\cite{hendrycks2021ImageNetR}, and near-distribution images (ImageNet-O~\cite{hendrycks2021naturalAdv}).

Similar to our low-density samples, DenseFlow assigns very high likelihood scores to all three novel variations of data (Table~\ref{tab: exact_nll}). Such variations (e.g., sketches) are rarely present in training data. Despite that, the model assigns a high likelihood to them. We also provide a qualitative comparison in Figure~\ref{fig: exact_nll}. Our observation is similar to the failure of exact likelihood scores on out-of-distribution data~\cite{nalisnick2018GenDontKnow}.

}

\begin{table}[!h]
    \centering
    \caption{\textbf{Quantitative evaluation.} State-of-the-art negative log-likelihood (NLL) evaluation using DenseFlow~\cite{grcic2021denselyNormFlow}. Lower value implies higher likelihood.}
    \Large
    \resizebox{\linewidth}{!}{
    \begin{tabular}{cccc|cccc} \toprule
       Dataset  &  Real & BigGAN & \begin{tabular}[c]{@{}c@{}}DDPM \\ (baseline)\end{tabular} & \begin{tabular}[c]{@{}c@{}}DDPM \\ (ours)\end{tabular} & Rendition & Sketch & ImageNet-O \\
       NLL  & $3.4$ & $3.1$ & $3.3$ & $2.8$ & $2.5$ & $1.2$ & $2.9$ \\ \bottomrule
    \end{tabular}}
    \label{tab: exact_nll}
\end{table}

\begin{figure*}[!htb]
    \centering
    \begin{subfigure}[b]{\linewidth}
        \centering
        \includegraphics[width=0.9\linewidth]{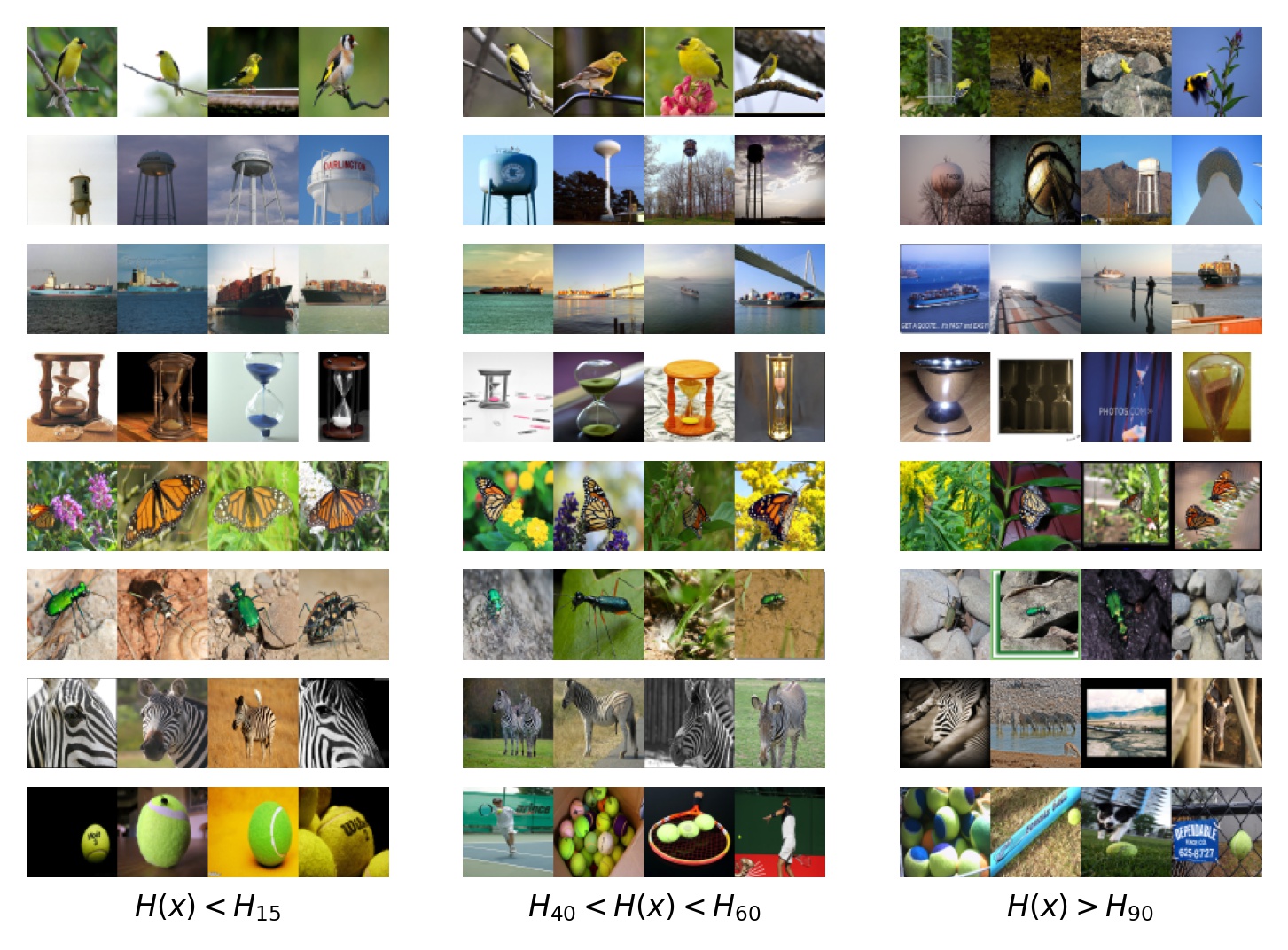}
        \caption{Visualizing images across the hardness scores axis for each class. $H_a$ refers to the $a_{th}$ percentile of hardness score. Classes are: goldfinch, water tower, container ship, hourglass, monarch butterfly, tiger beetle, zebra, and tennis ball.}
        \label{fig: why_hardness_score_a}
        \vspace{10pt}
    \end{subfigure}
    \begin{subfigure}[b]{\linewidth}
        \centering
        \includegraphics[width=0.8\linewidth]{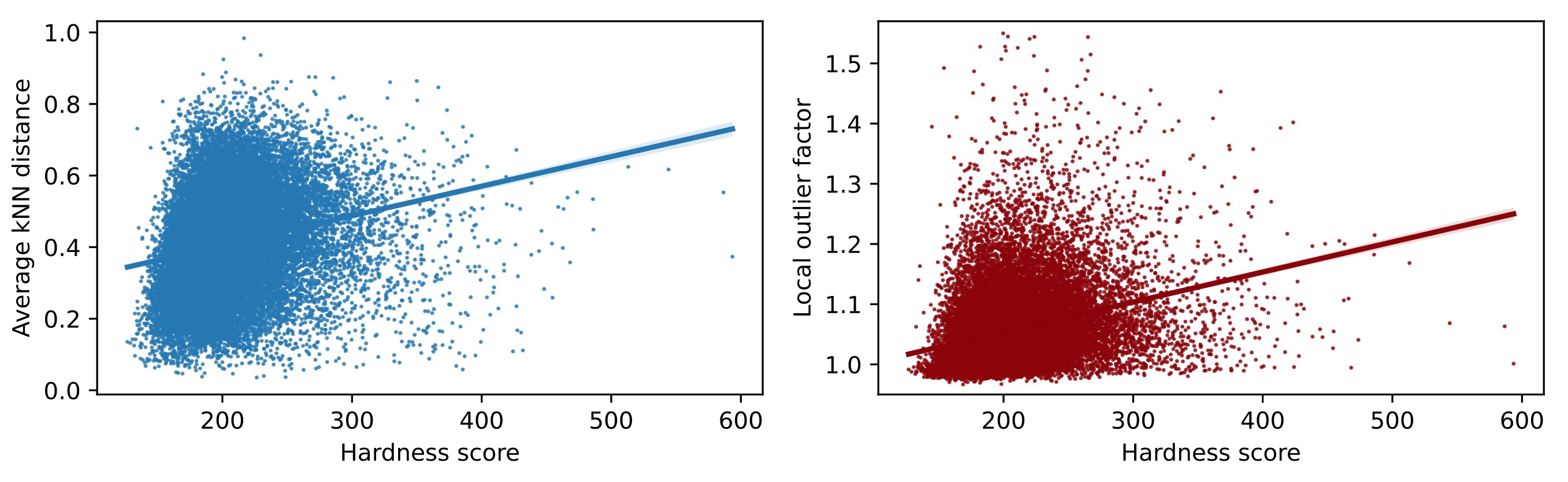}
        \caption{Correlation of hardness score with other metrics.}
        \label{fig: why_hardness_score_b}
    \end{subfigure}
    \caption{\textbf{Validating the effectiveness of hardness score.} To validate whether hardness score is a good proxy for neighborhood density, we first visualize images with increasing hardness scores and next show that it correlates with commonly used metrics to measure data density.}
\end{figure*}

\subsection{Higher hardness score implies lower neighborhood density}  \label{app: hardnessJustify}
In our sampling process, we maximize the hardness score of synthetic data. We argued that hardness score is a proxy to neighborhood density, thus maximizing it forces the model to generate low density samples. We provided the validation of its success in Figure~\ref{fig: density_comp}. Now we delve deeper into why hardness score acts as a proxy to neighborhood density. 

First we visualize real images across the spectrum of hardness score. Give a class index in the ImageNet validation set, we visualize its samples with lowest, moderate, and highest hardness scores (Figure~\ref{fig: why_hardness_score_a}). From these images, it is evident that the difficulty of individual instances increases with hardness scores. We also look into the correlation of hardness score with other known density metrics (Figure~\ref{fig: why_hardness_score_b}). We find that hardness score also have a positive correlation with other metrics.   

\subsection{Motivation to normalize gradients}  \label{app: normalGrad}
Slightly different from the classifier guidance approach in Dhariwal et al.~\cite{dhariwal2021diffBeatGANs}, we normalize classifier gradients before using them in the sampling process. We do so since it makes the scale of hyperparameters ($\alpha$ and $\beta$) independent of the magnitude of gradients of guiding losses ($L_{g_1}$ and $L_{g_2}$). In particular, we observed that the magnitude of gradients in the diffusion process is often quite small, thus needing a very high scaling parameter. In addition, the magnitude of gradients also fluctuates with timesteps of the sampling process, thus potentially requiring a different scaling parameter at different timesteps. We normalize gradients to have unit $\ell_{\infty}$ norm, which ensures a consistent magnitude of gradients across timesteps. Thus normalization isolates the choice of scaling hyperparameters from gradients magnitude, making this choice much simpler. 

\subsection{Effect of different guiding loss functions}  \label{app: tempEffect}

\begin{figure}
    \centering
   \includegraphics[width=0.9\linewidth]{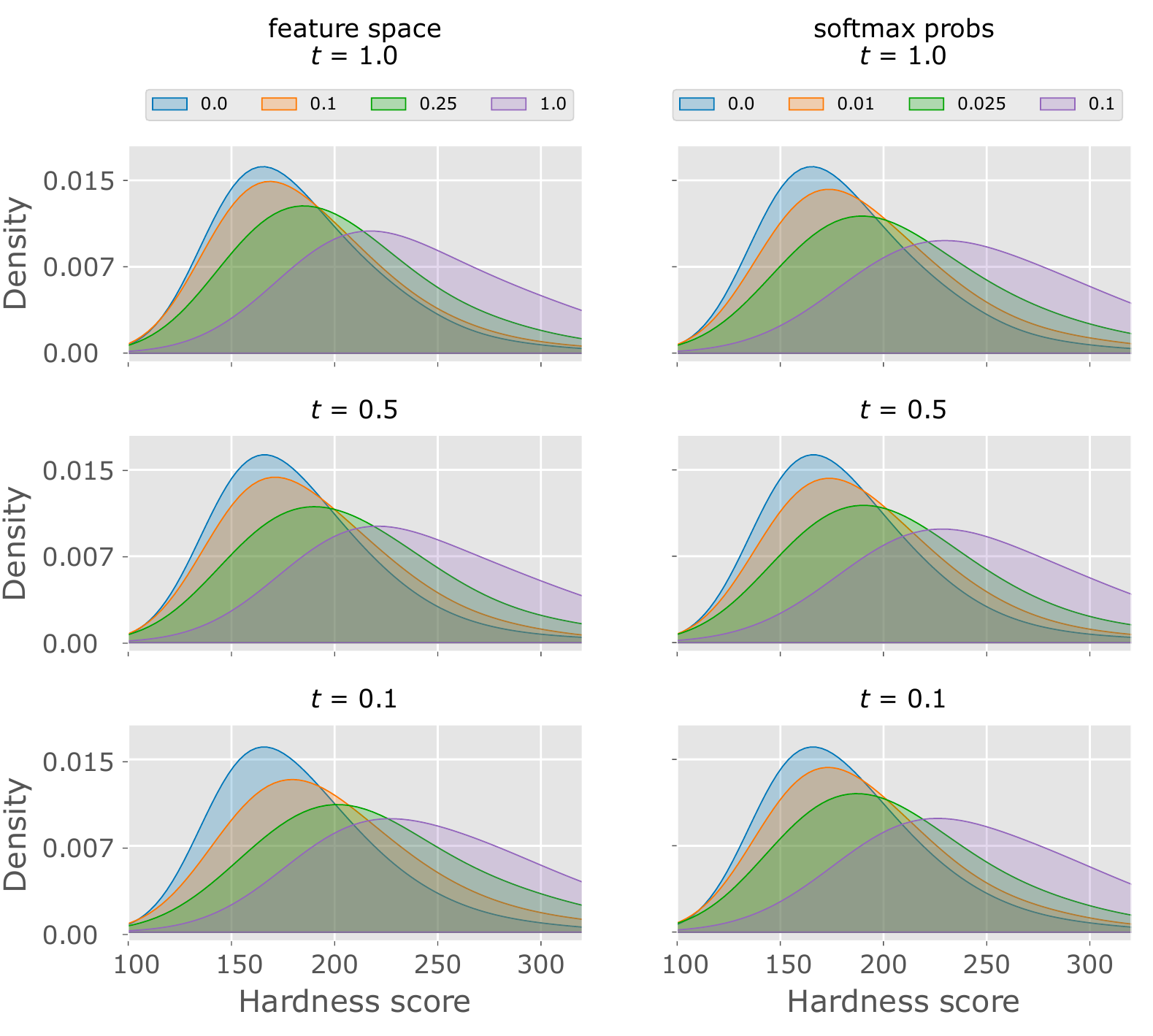}
  \caption{\textbf{Choice of loss function.} Loss function in feature space vs. in logits space.}
  \vspace{-20pt}
  \label{fig: app_loss_fxns}
\end{figure}
In our sampling process, our objective is to push synthetic images away from high-density neighborhoods. We achieve it by using a softmax-based loss function in the feature space of a pre-trained classifier. However, an equivalent loss function can be derived using softmax probabilities at the logit layer. Though both loss functions require a different scale of hyperparameters, they achieve similar results under properly calibrated scales (Figure~\ref{fig: app_loss_fxns}). We make use of feature space because multiple additional metrics to measure density, such as kNN distance and local outlier factor, can be also easily calculated in the feature space.

\subsection{Limitation of class embeddings smoothing}  \label{app: SmoothEmbed}
\begin{figure*}[!htb]
    \centering
    \includegraphics[width=0.98\linewidth]{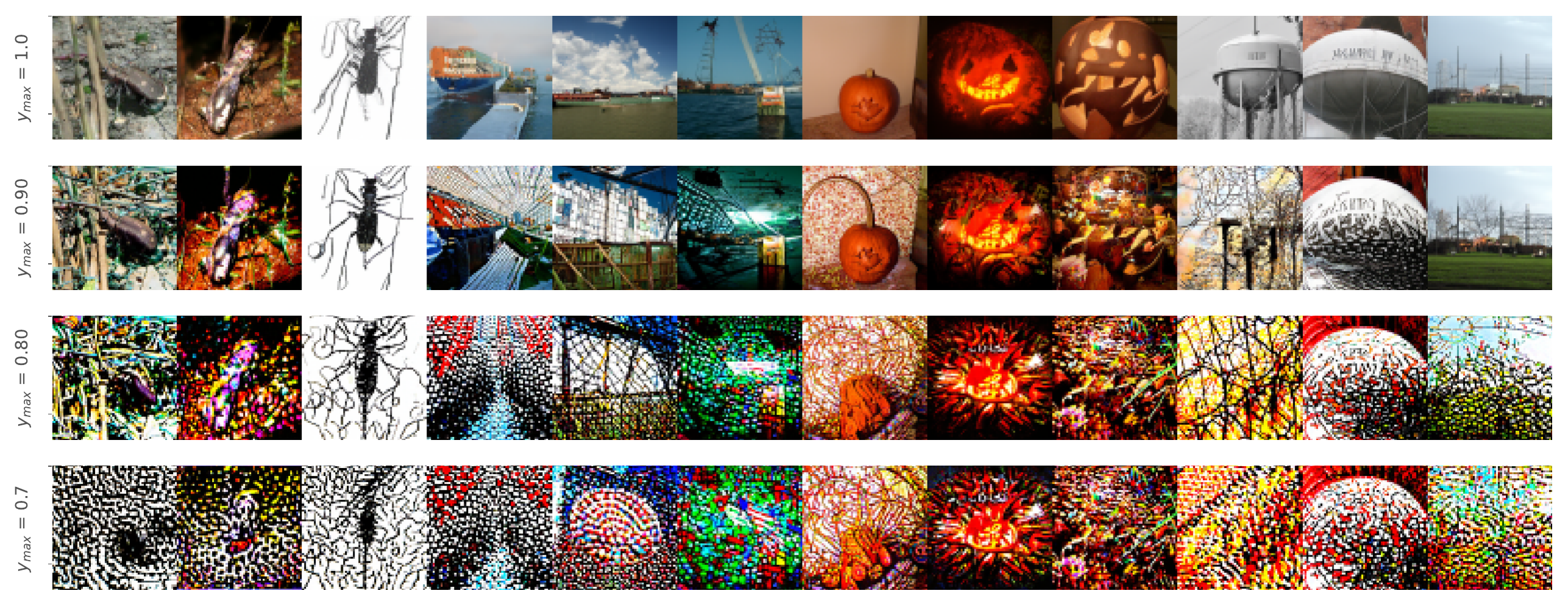}
    \caption{\textbf{Smoothing of class embeddings.} Demonstrating how smoothing of class embeddings leads to poor quality synthetic images with diffusion models.}
    \label{fig: app_smooth_embed}
\end{figure*}

Previously, Li et al.~\cite{li2020ClassEmbedSmoothGAN} showed that one can manipulate class-embeddings of a pre-trained BigGAN model to improve the diversity of generated images. When approaching the task of low-density sampling, it is natural to test whether it can be achieved by simply controlling class embeddings. To test the effect of class-embeddings, we smooth class embeddings for a diffusion model on the ImageNet dataset. The network is trained with one-hot encoded class embeddings. When sampling, we smooth the embeddings by reducing the correct class probability to $y_{max}$ and distribute the rest of the probability mass equally over all remaining classes. We find that the quality of synthetic images degrade very quickly with a reduction in $y_{max}$ (Figure~\ref{fig: app_smooth_embed}). Given this detrimental effect of smoothing in class embeddings, we chose to modify the sampling process itself, since the latter provides a much better control and quality of synthetic images.  

\subsection{Integration with fast sampling techniques}  \label{app: ddim}
\begin{figure}[!htb]
    \centering
    \begin{subfigure}[b]{0.3\linewidth}
        \includegraphics[width=\linewidth]{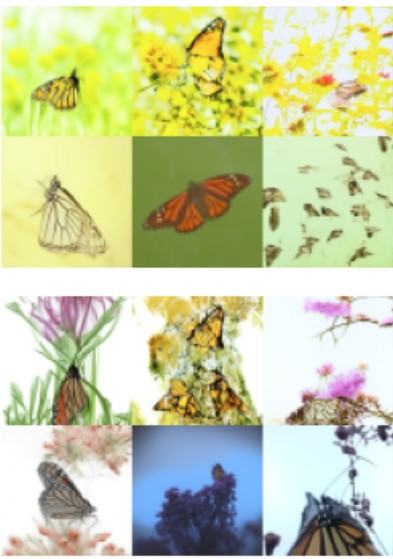}
        \caption{$T=10$}
    \end{subfigure}
    \begin{subfigure}[b]{0.3\linewidth}
        \includegraphics[width=\linewidth]{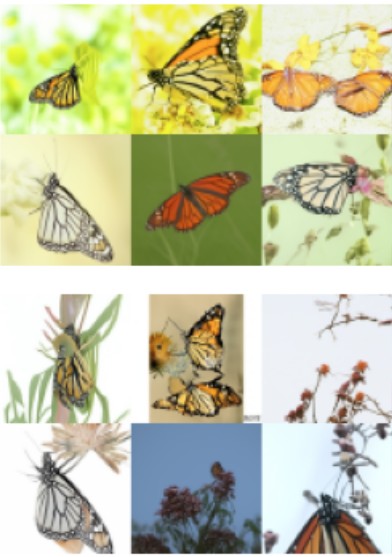}
        \caption{$T=20$}
    \end{subfigure}
    \begin{subfigure}[b]{0.3\linewidth}
        \includegraphics[width=\linewidth]{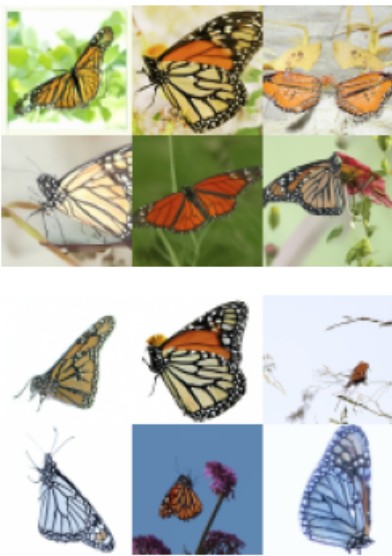}
        \caption{$T=50$}
    \end{subfigure}
    \caption{\textbf{Fast sampling.} We integrate our guiding objective in the fast DDIM sampling process~\cite{song2020ddim}. Top two rows correspond to the baseline DDIM sampling approach while bottom two correspond to our approach. We use the identical starting latent vectors for both approaches and across the three choices of the number of sampling steps.}
    \vspace{-10pt}
    \label{fig: ddim_comp}
\end{figure}

In the main paper, we discussed that with fast sampling approaches, our approach enjoys a similar trade-off as baseline sampling process. To support this claim, we provide a comparison of synthetic images sampled using DDIM~\cite{song2020ddim} sampling process from both baseline and our sampling process in figure~\ref{fig: ddim_comp}. We integrate the guiding loss in the DDIM sampling process in a similar manner as Dhariwal~\etal~\cite{dhariwal2021diffBeatGANs}.

\section{Experimental results}


\subsection{Neighborhood density with different feature extractors}  \label{app: kNNClassifier}
\begin{figure*}[!htb]
    \centering
    \begin{subfigure}[b]{\linewidth}
         \centering
         \includegraphics[width=0.28\linewidth]{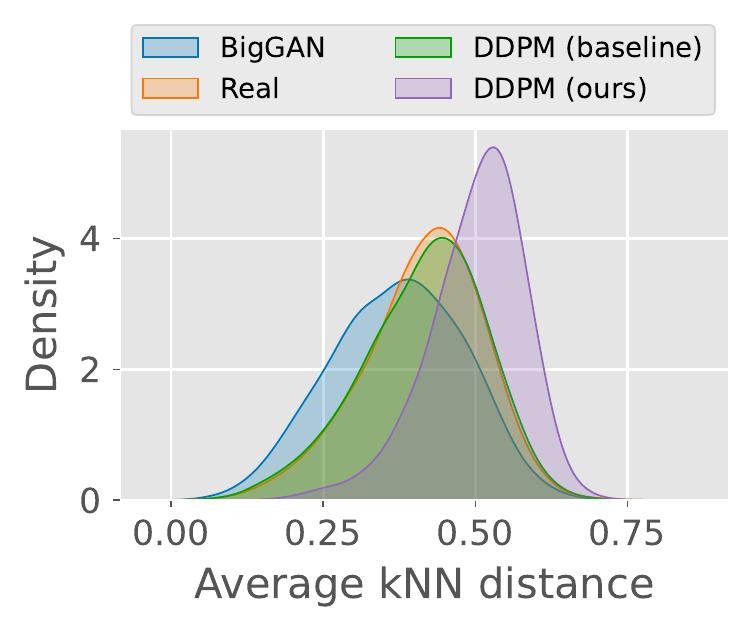}
         \includegraphics[width=0.3\linewidth]{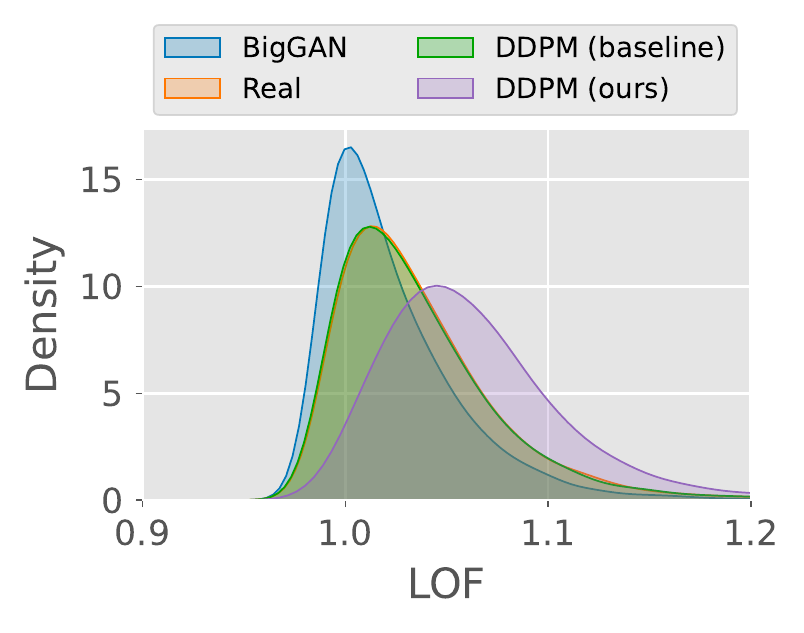}
         \caption{Inception-v3}
    \end{subfigure}
    
    \begin{subfigure}[b]{\linewidth}
          \centering
         \includegraphics[width=0.28\linewidth]{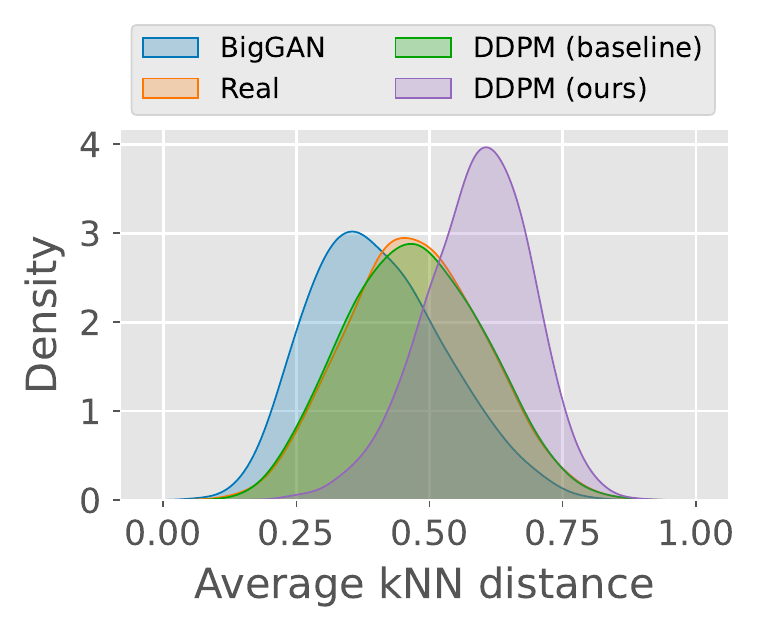}
         \includegraphics[width=0.3\linewidth]{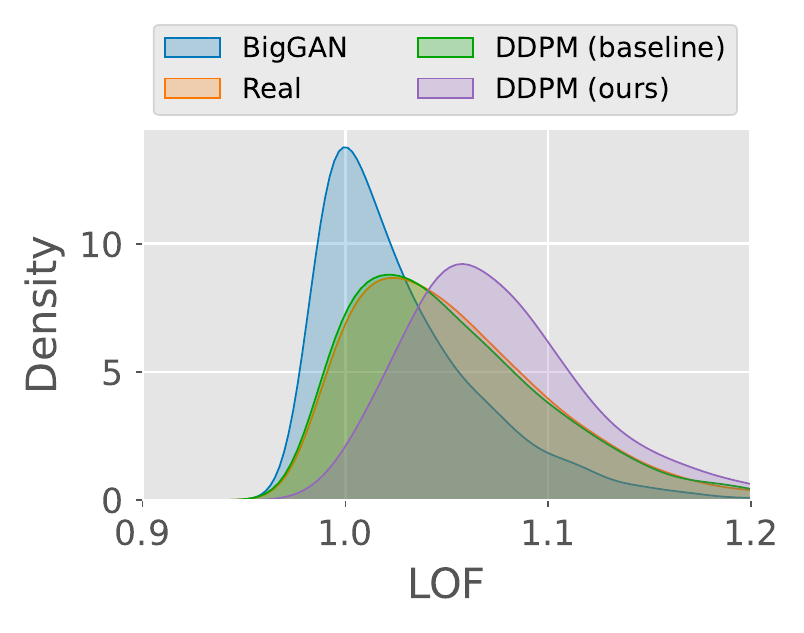}
         \caption{VGG19}
    \end{subfigure}
    \caption{\textbf{Comparing neighborhood density across different choices of feature extractors.} We use two additional feature extractors, namely Inception-v3 and VGG19.}
    \label{fig: density_comp_app}
\end{figure*}

We use a ResNet-50 classifier, which is pretrained on ImageNet dataset, as feature extractor. Though this is very common choice of deep neural network, we further investigate whether our claims are robust to the choice of the feature extractors. To test it, we consider two more deep neural networks, namely Inception-V3~\cite{szegedy2016inception_v3} and VGG~\cite{simonyan2014VGG}. We measure the neighborhood density in the feature space of both classifier and show that both classifier further validate the success of our approach (Figure~\ref{fig: density_comp_app}).   

\subsection{Additional nearest neighbor pairs for visualization}  \label{app: kNNPairs}
\begin{figure*}[!htb]
    \centering
    \includegraphics[width=0.9\linewidth]{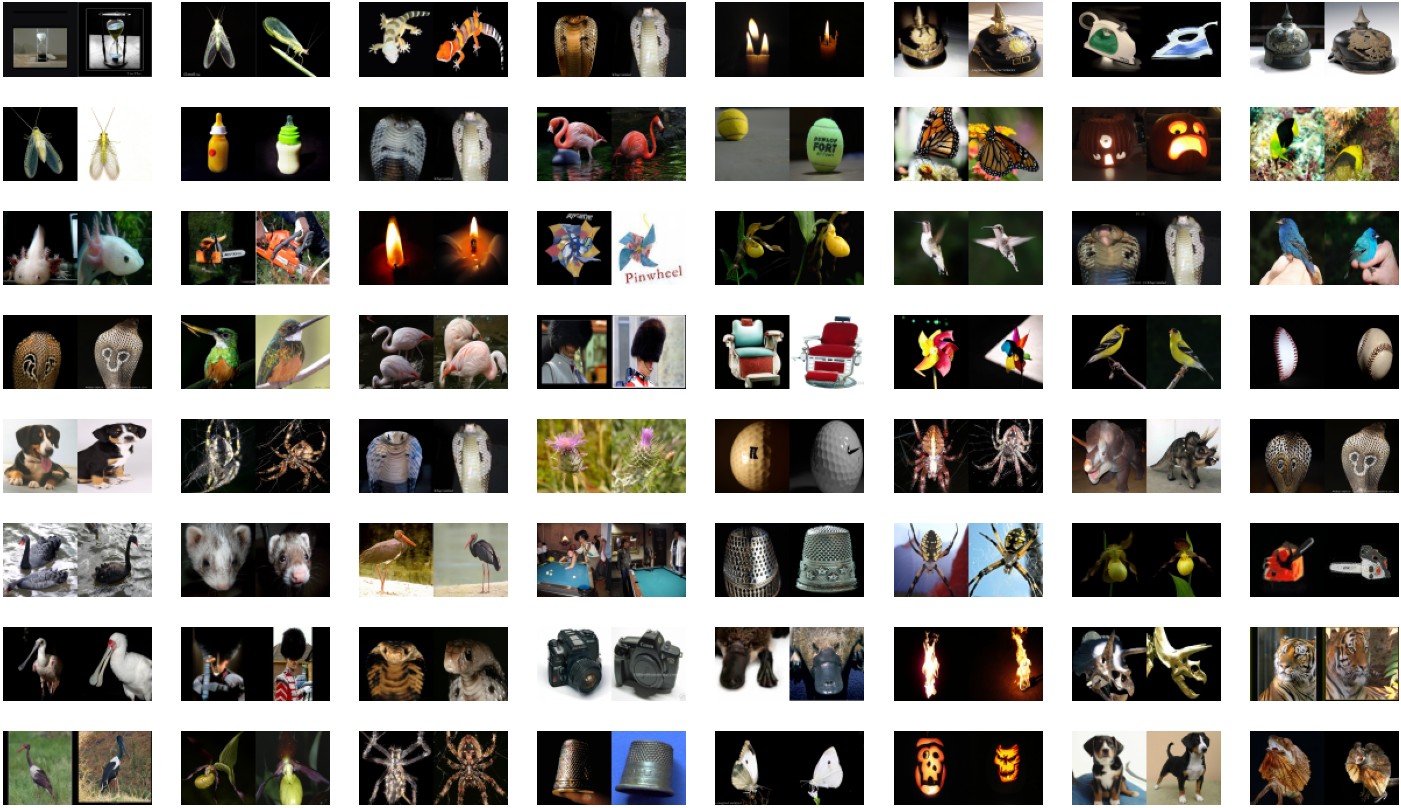}
    \caption{Nearest neighbour pairs of real and synthetic data with lowest pairwise distance. In each pair, the left and right image correspond to the synthetic and real image, respectively.}
    \label{fig: nn_app}
\end{figure*}

\begin{figure*}[!htb]
    \centering
    \includegraphics[width=0.9\linewidth]{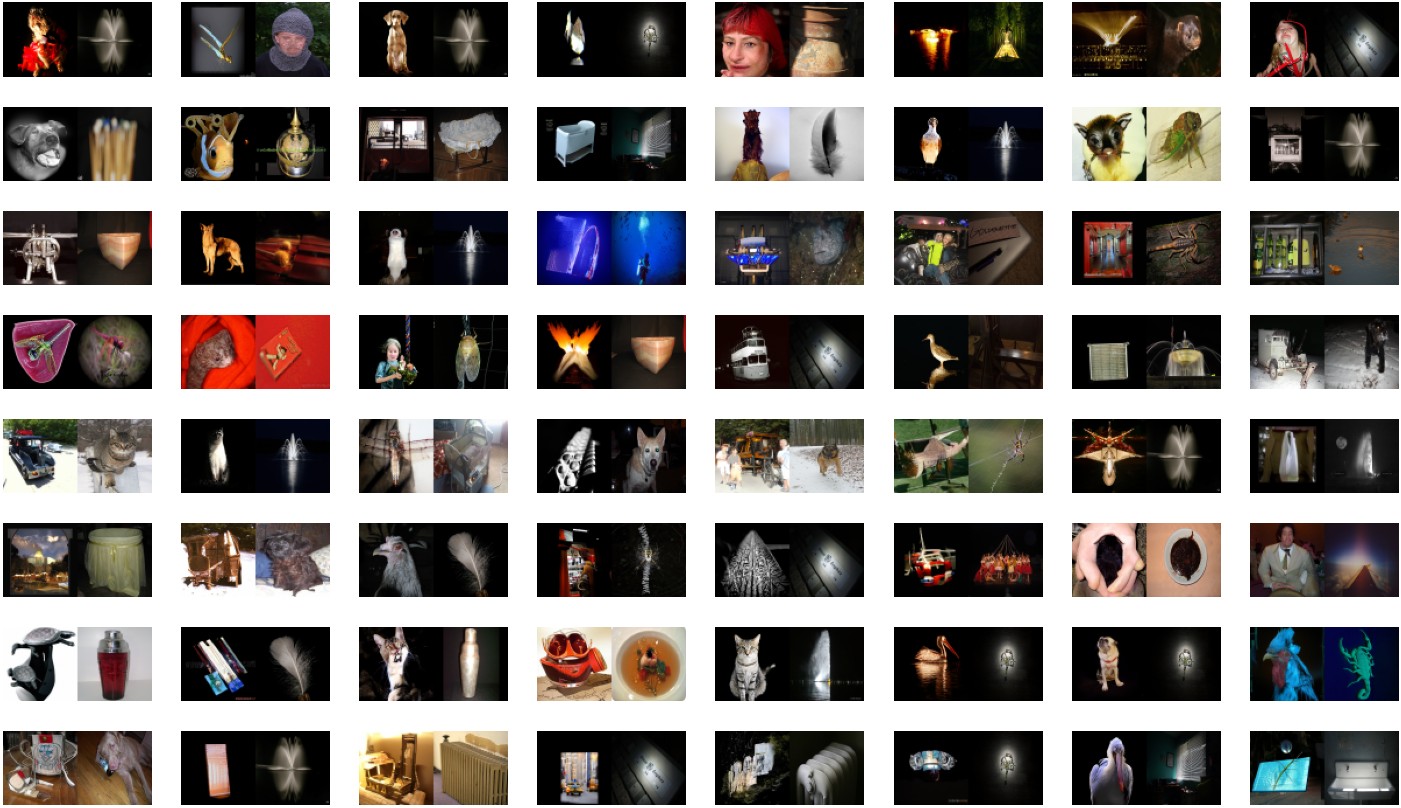}
    \caption{\bluetext{Nearest neighbour pairs of real and synthetic data with lowest pairwise distance in pixel space. In each pair, the left and right image correspond to the synthetic and real image, respectively.}}
    \label{fig: nn_app_pixel_space}
\end{figure*}

To analyze whether the diffusion model is simply memorizing training data, we visualize the nearest neighbors of each synthetic image from the real images. We synthesize the synthetic images using our sampling process. To complement the top-16 synthetic and real images with the smallest pairwise distance in Figure~\ref{fig: nn_pairs}, we present the next 64 pairs in Figure~\ref{fig: nn_app}. In each pair, the left and right images corresponds to the synthetic and real image, respectively. \bluetext{For completeness, we also analyze nearest neighbors in pixel space (Figure~\ref{fig: nn_app_pixel_space}). As expected, euclidean distance in pixel space doesn't correspond to semantic similarity between images and it is often highly biased toward background similarities between synthetic and real images.}

\subsection{Comparing our samples with baseline sampling process}
\label{app: sampleComp}
\begin{figure*}
    \centering
    \begin{subfigure}[b]{0.48\linewidth}
        \includegraphics[width=0.48\linewidth]{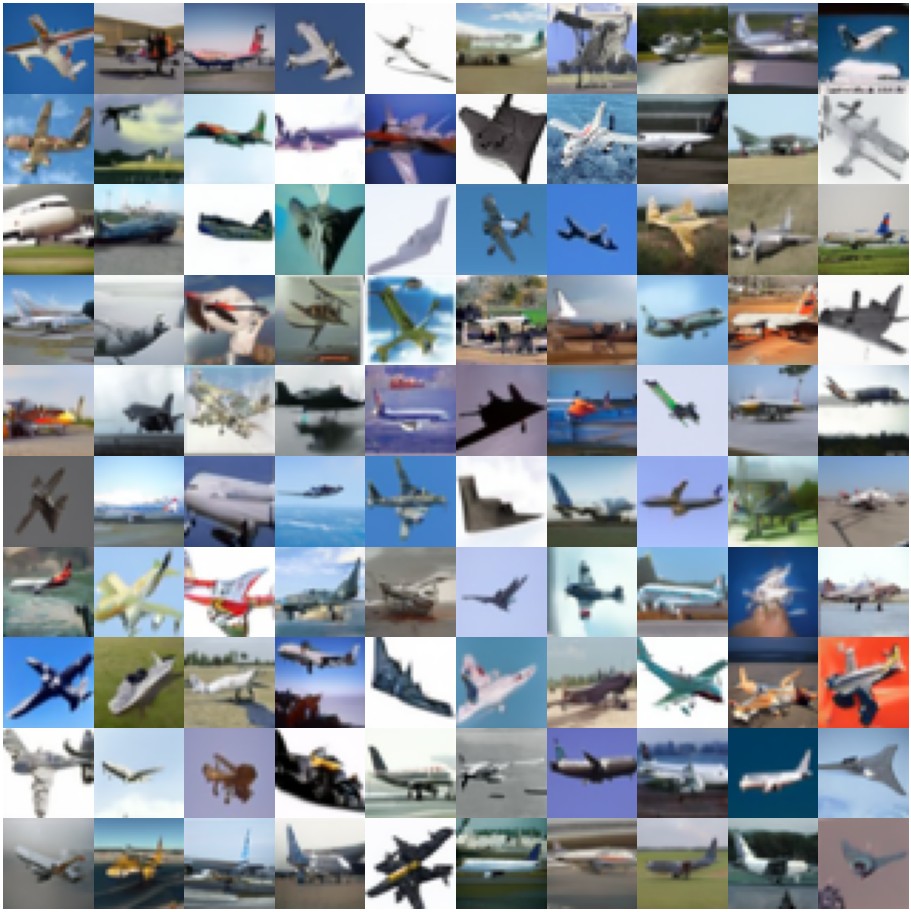}
        \includegraphics[width=0.48\linewidth]{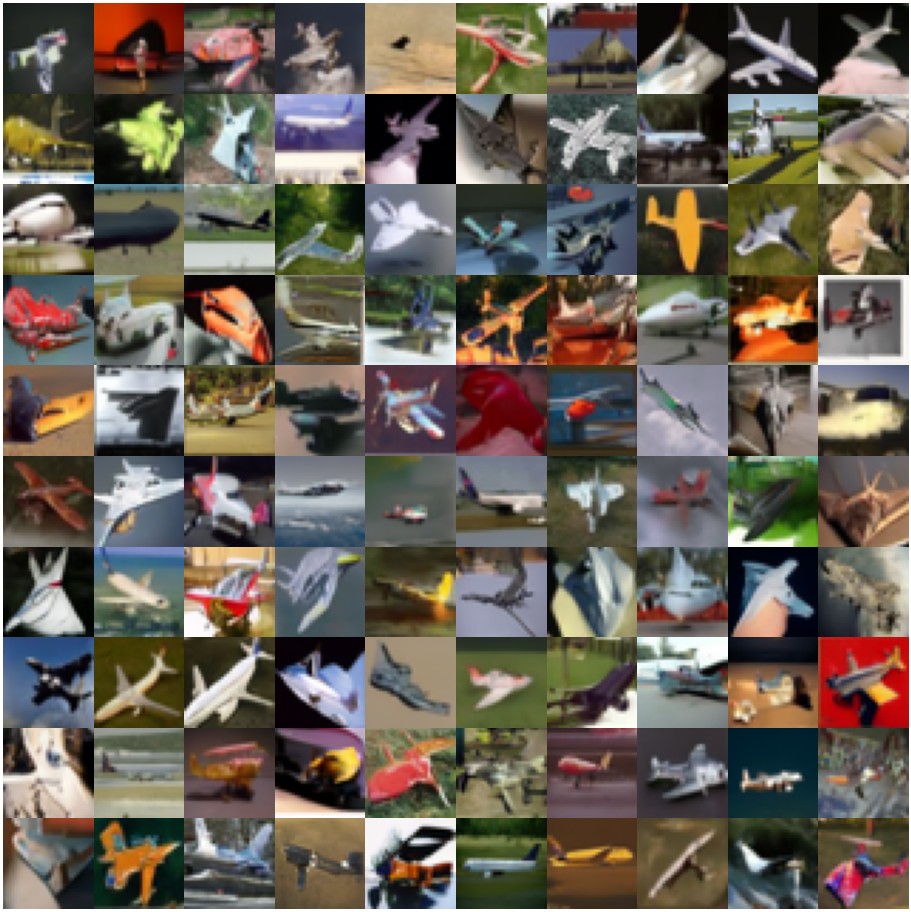}
    \end{subfigure}
    \begin{subfigure}[b]{0.48\linewidth}
        \includegraphics[width=0.48\linewidth]{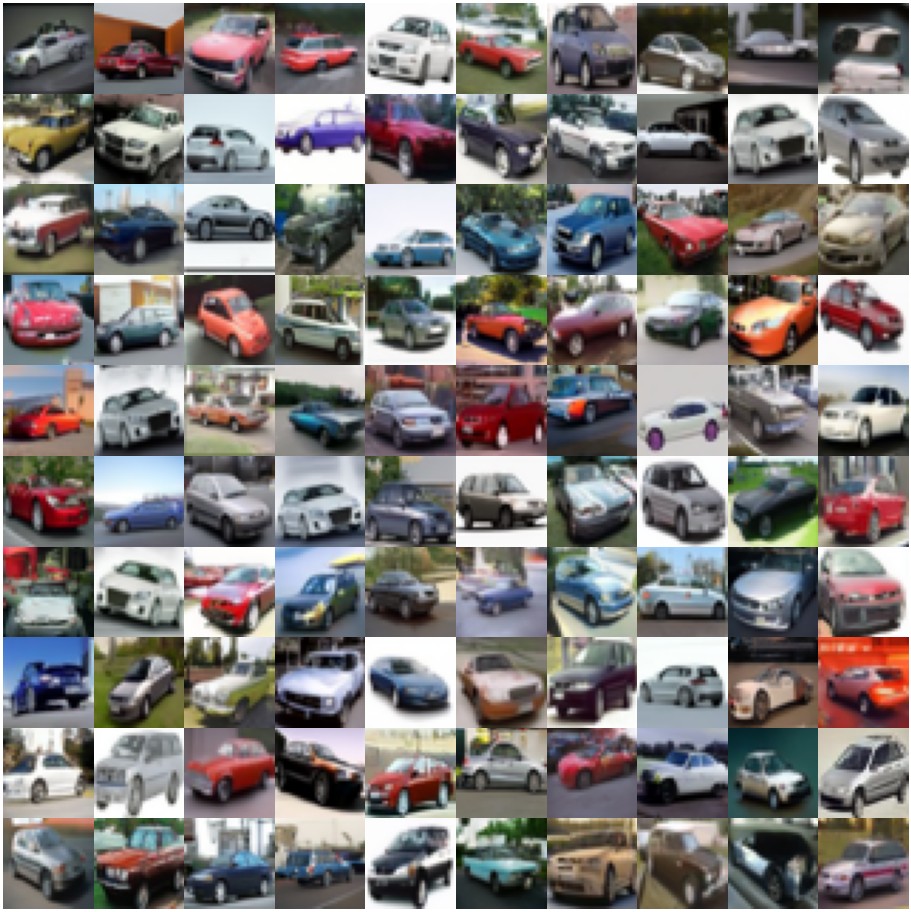}
        \includegraphics[width=0.48\linewidth]{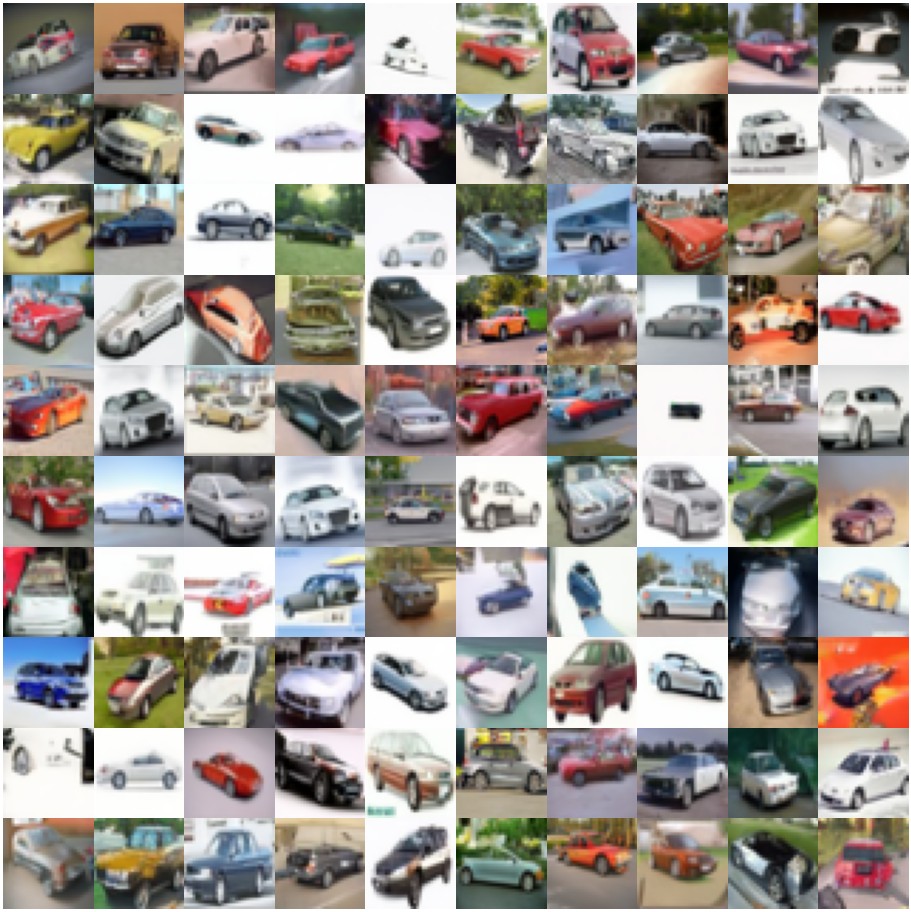}
    \end{subfigure}
    \begin{subfigure}[b]{0.48\linewidth}
        \includegraphics[width=0.48\linewidth]{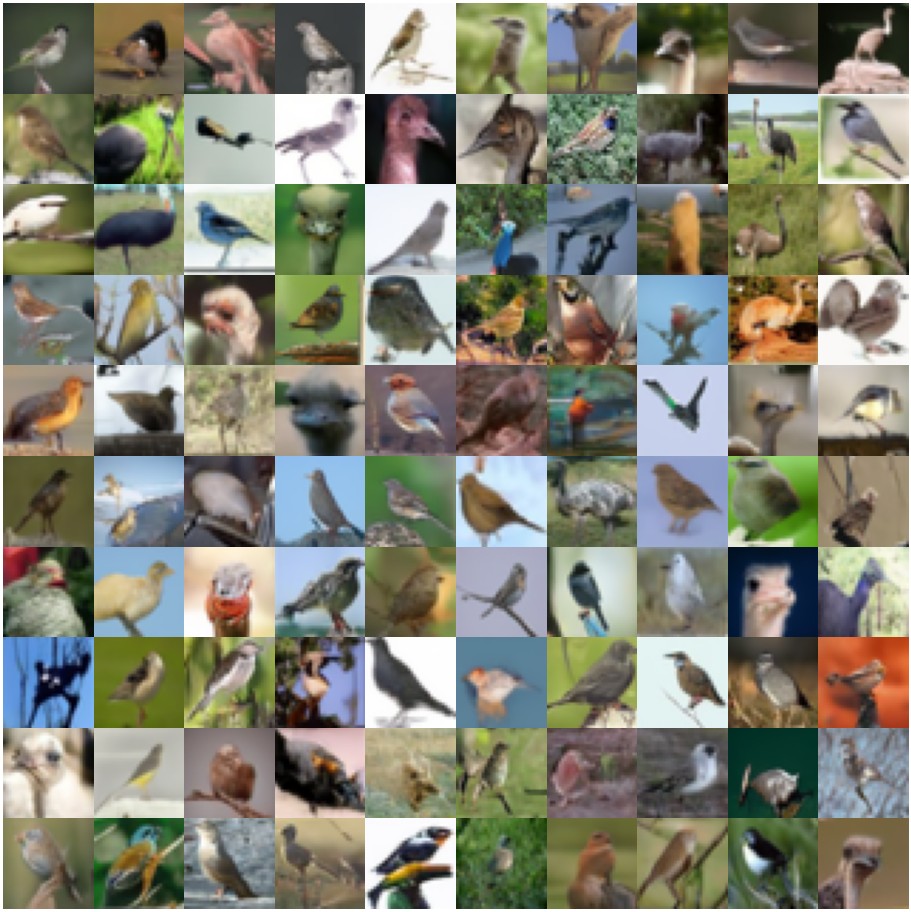}
        \includegraphics[width=0.48\linewidth]{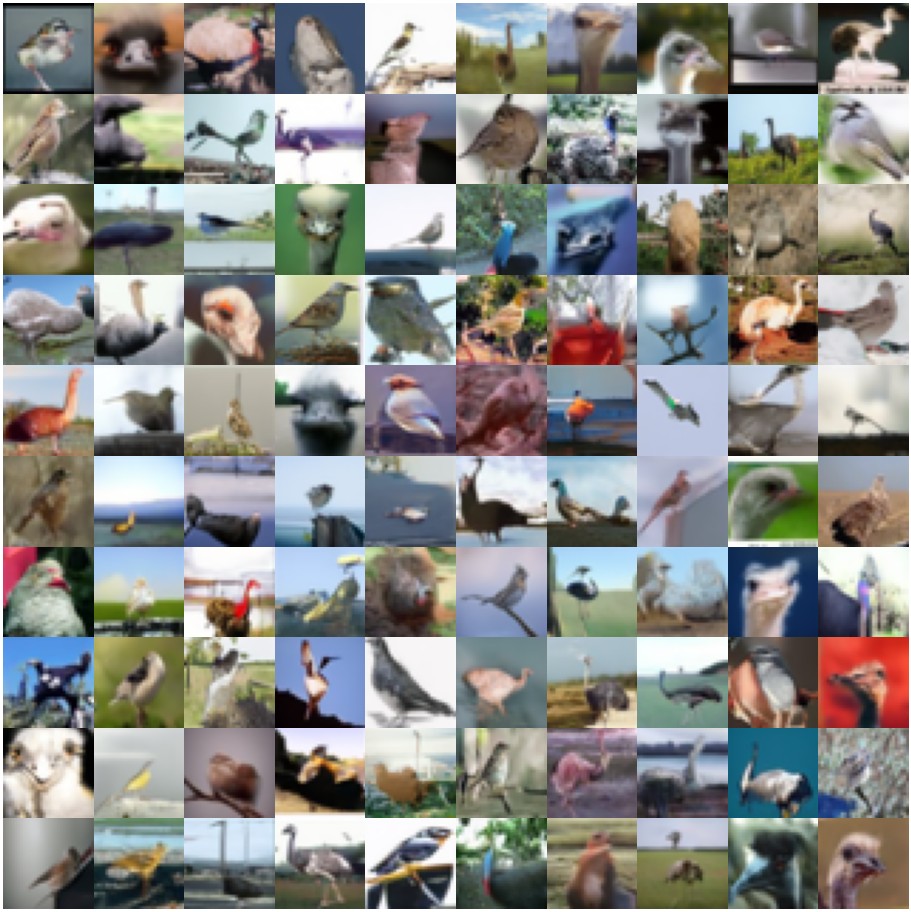}
    \end{subfigure}
    \begin{subfigure}[b]{0.48\linewidth}
        \includegraphics[width=0.48\linewidth]{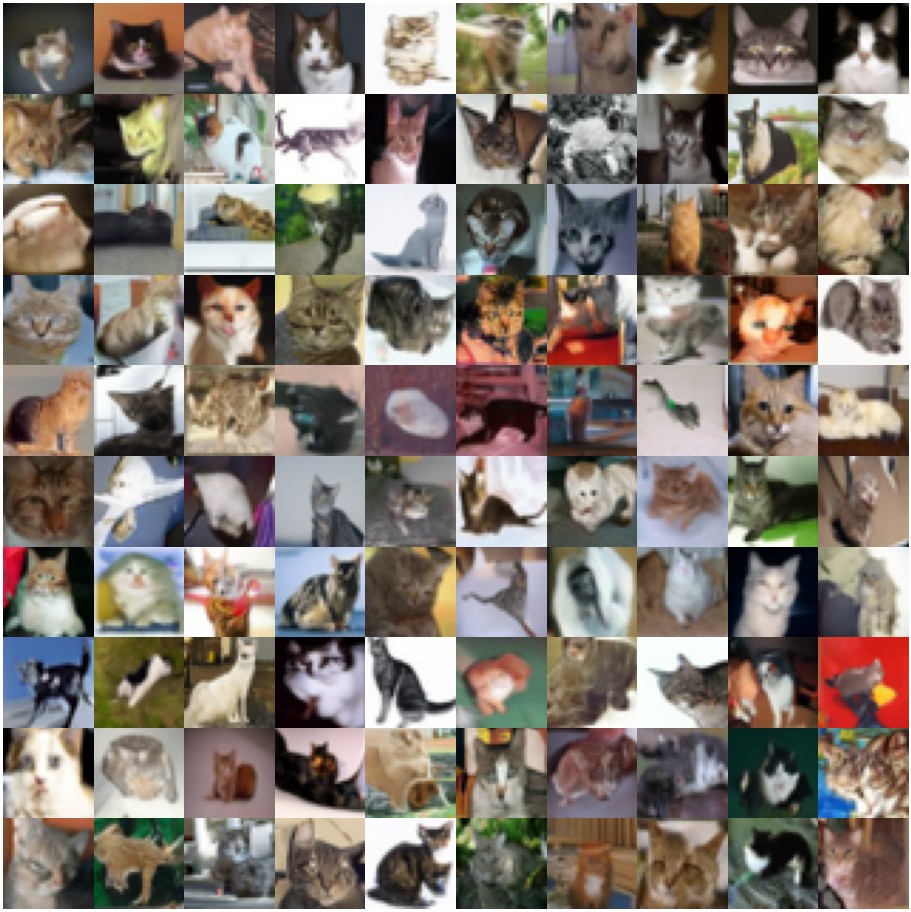}
        \includegraphics[width=0.48\linewidth]{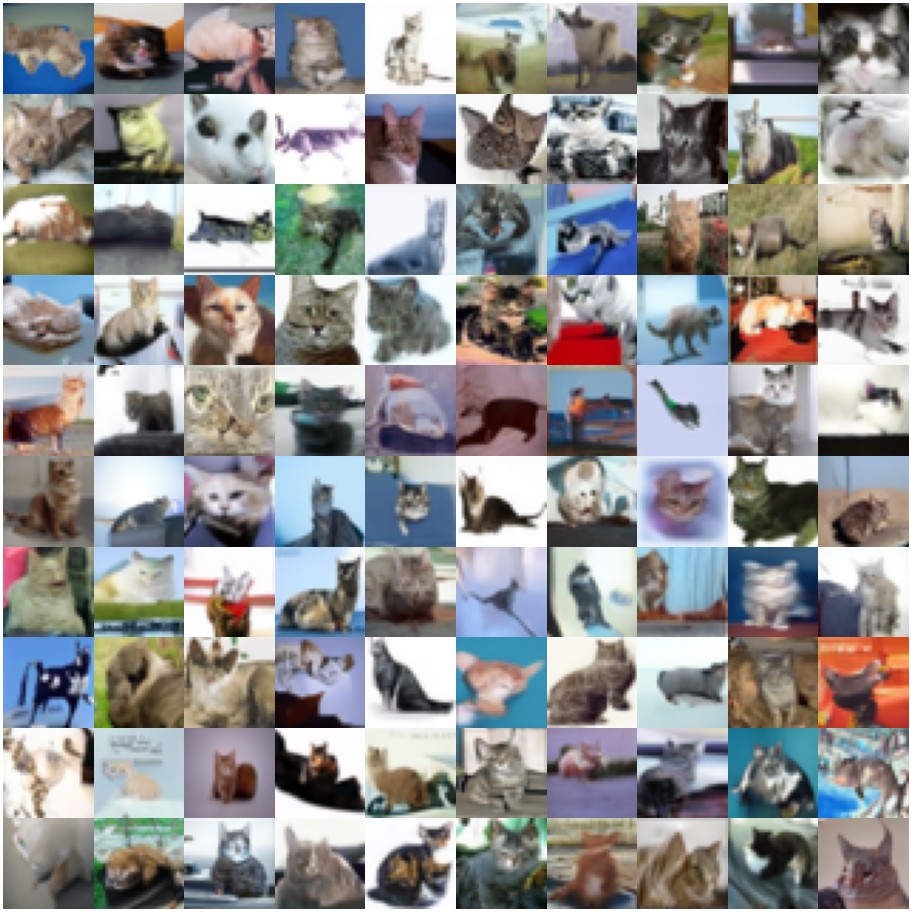}
    \end{subfigure}
    \begin{subfigure}[b]{0.48\linewidth}
        \includegraphics[width=0.48\linewidth]{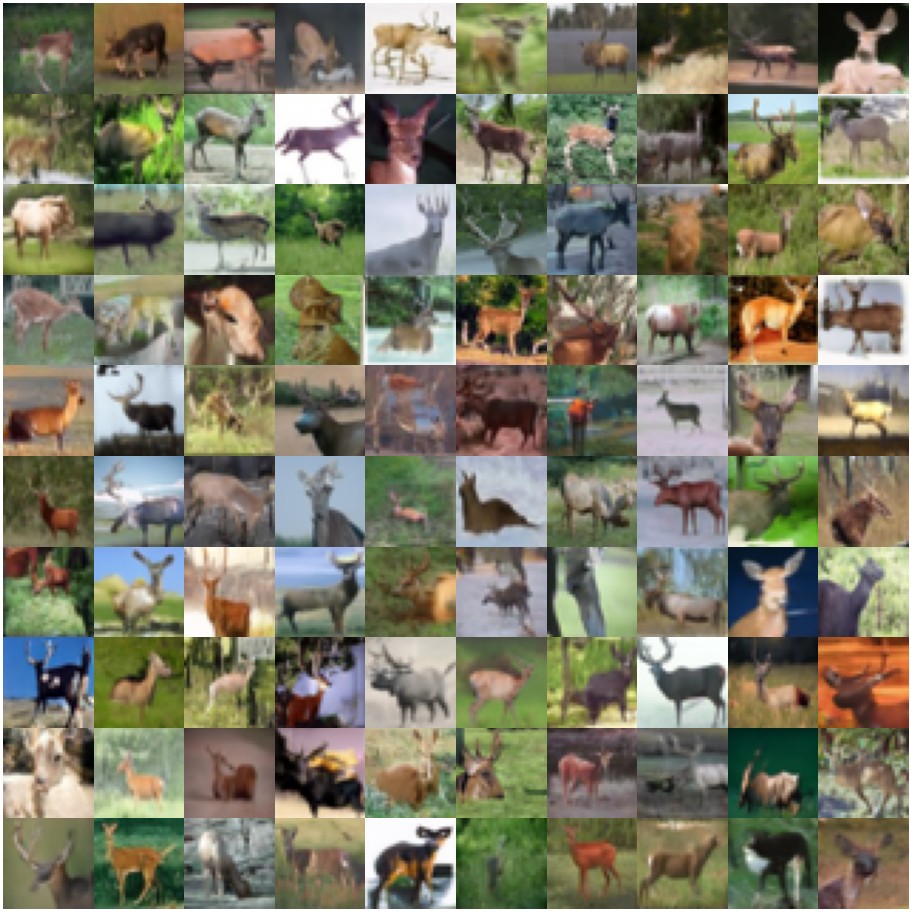}
        \includegraphics[width=0.48\linewidth]{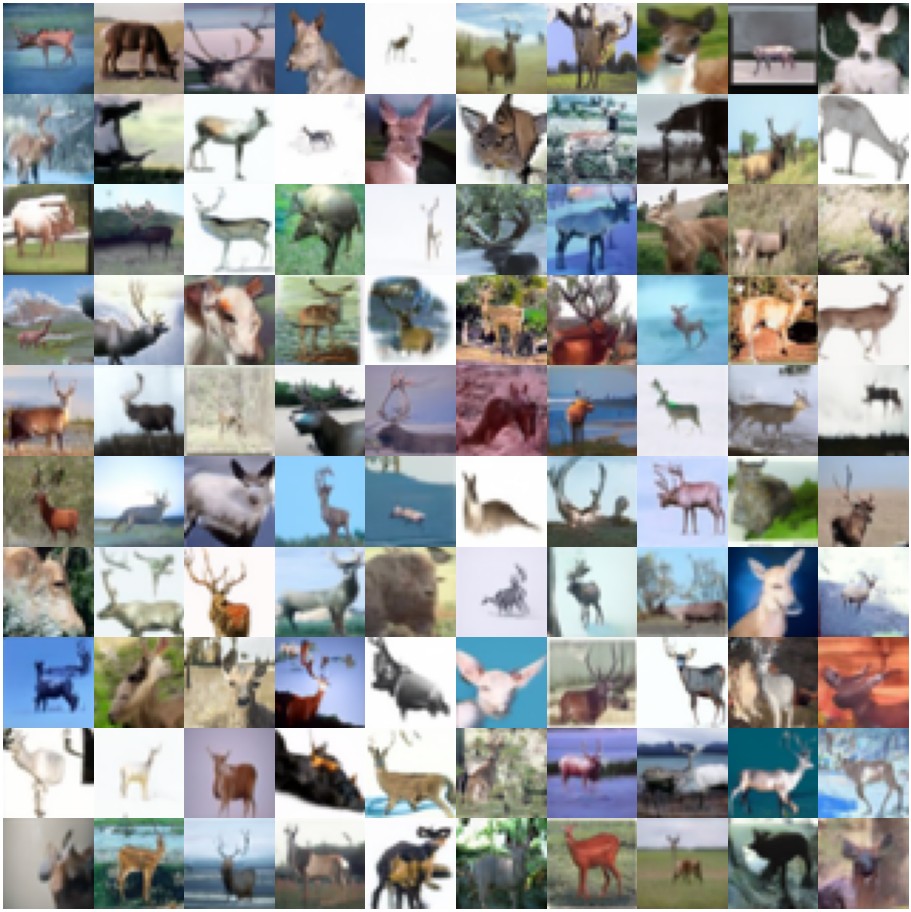}
    \end{subfigure}
    \begin{subfigure}[b]{0.48\linewidth}
        \includegraphics[width=0.48\linewidth]{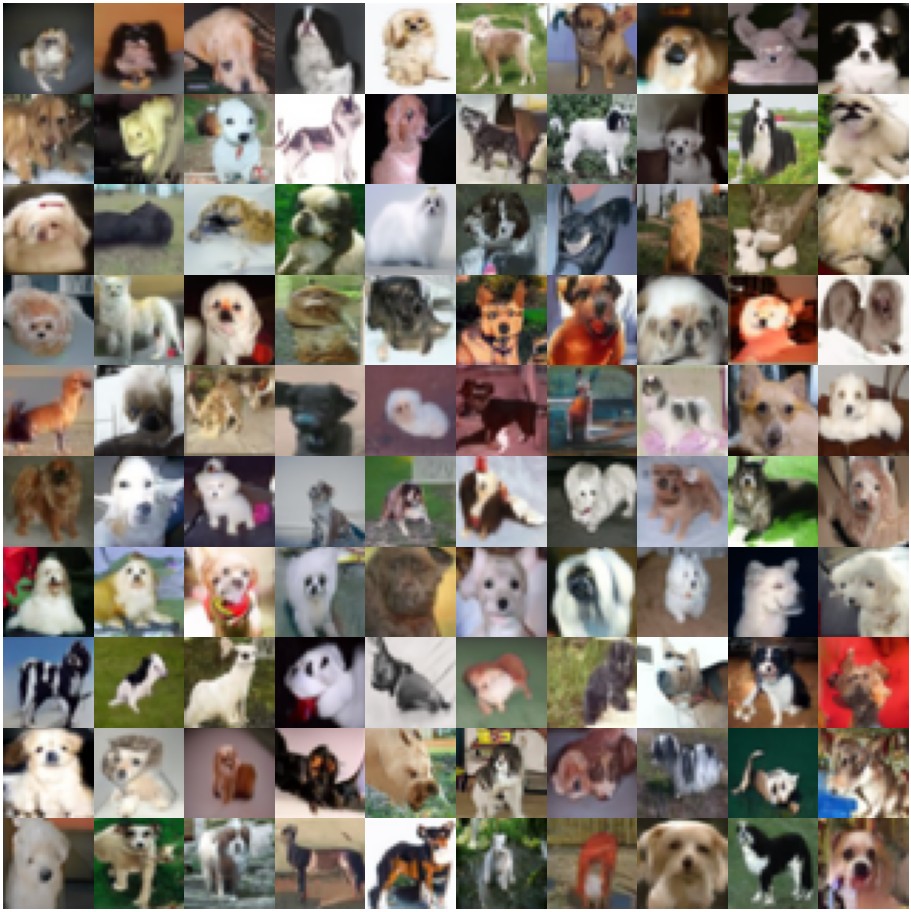}
        \includegraphics[width=0.48\linewidth]{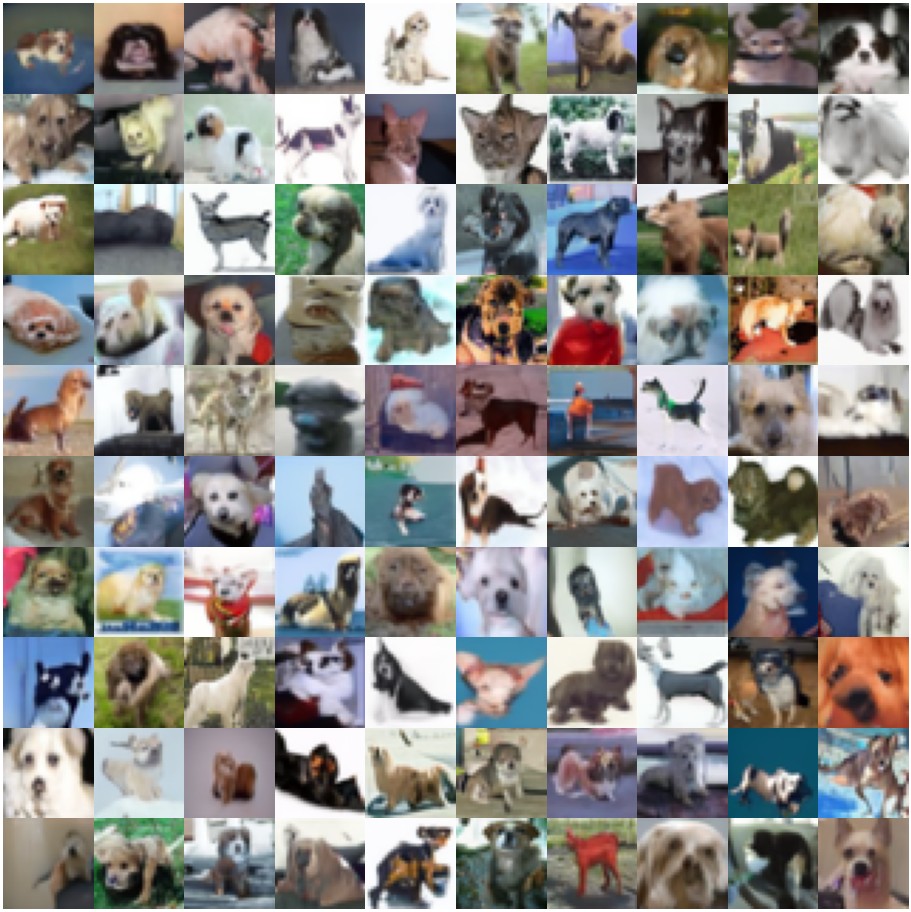}
    \end{subfigure}
    \begin{subfigure}[b]{0.48\linewidth}
        \includegraphics[width=0.48\linewidth]{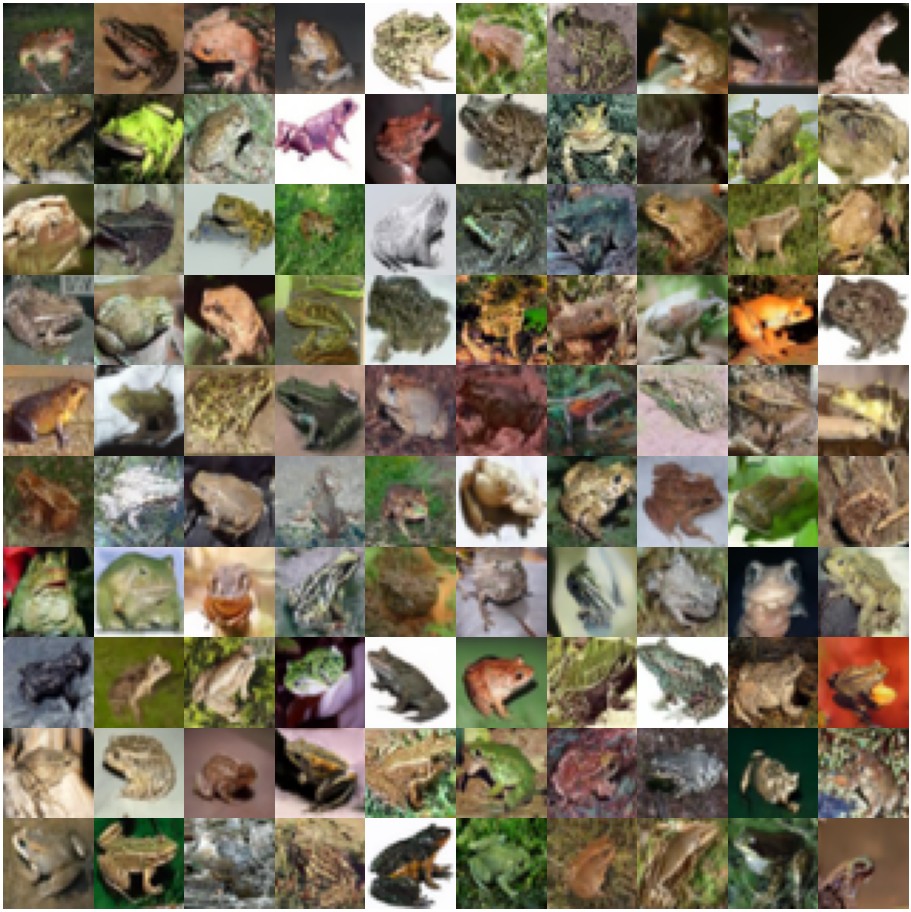}
        \includegraphics[width=0.48\linewidth]{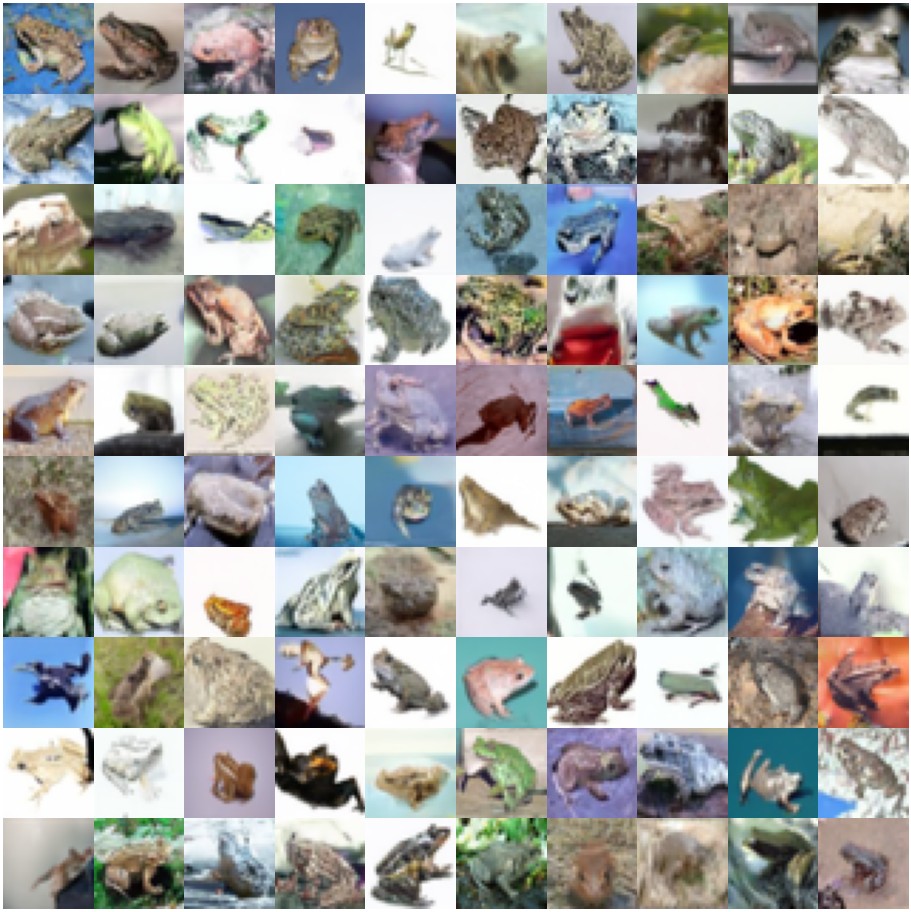}
    \end{subfigure}
    \begin{subfigure}[b]{0.48\linewidth}
        \includegraphics[width=0.48\linewidth]{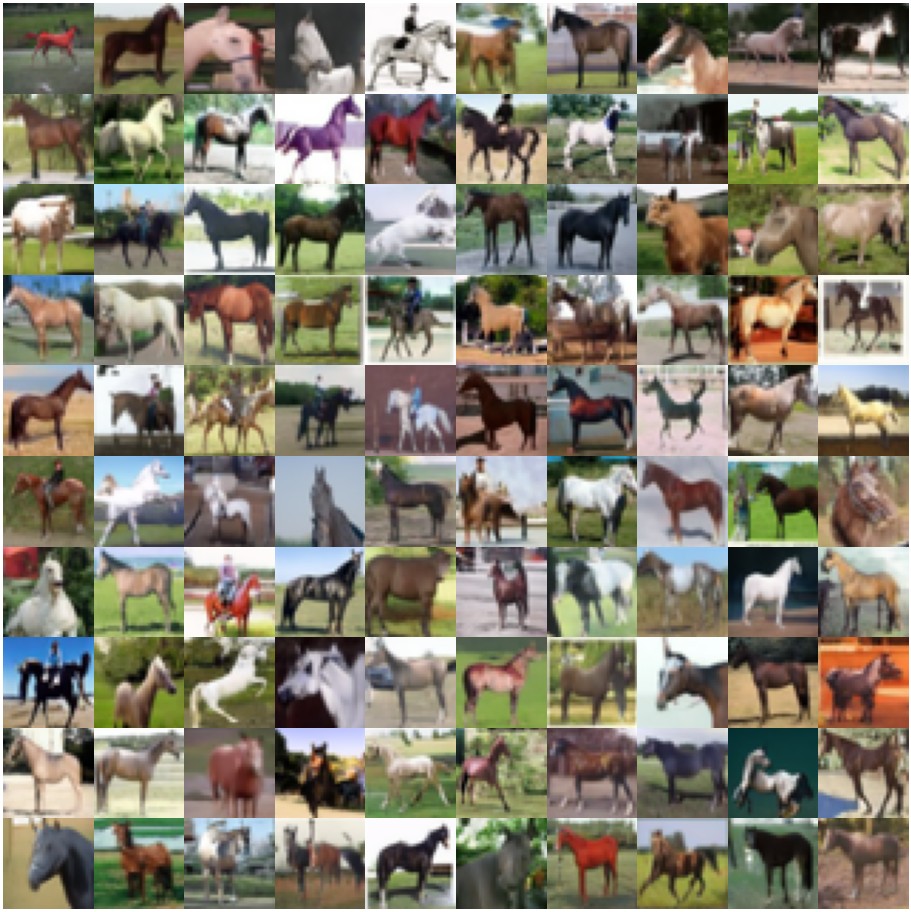}
        \includegraphics[width=0.48\linewidth]{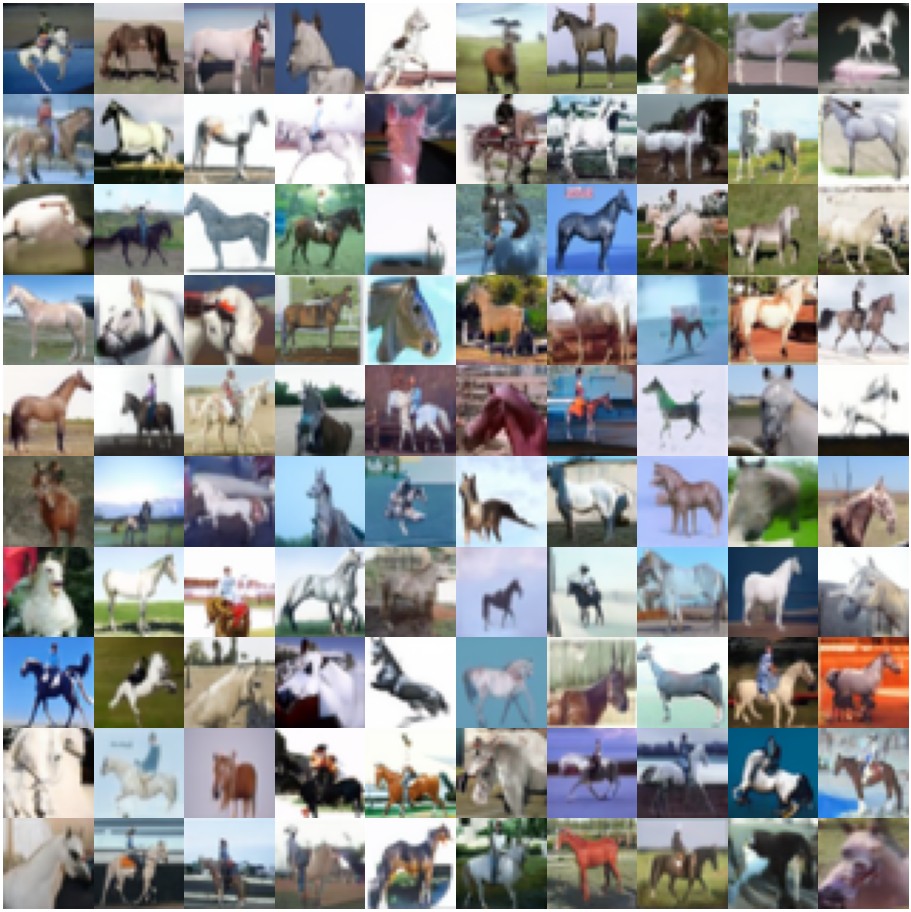}
    \end{subfigure}
    \begin{subfigure}[b]{0.48\linewidth}
        \includegraphics[width=0.48\linewidth]{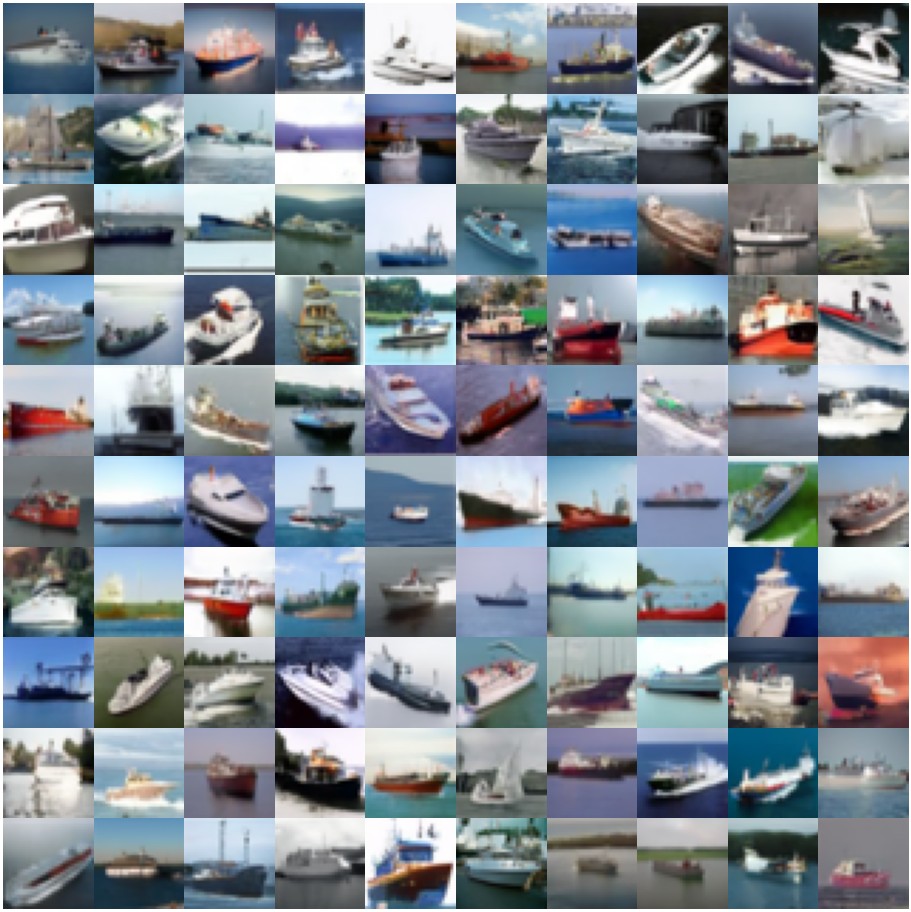}
        \includegraphics[width=0.48\linewidth]{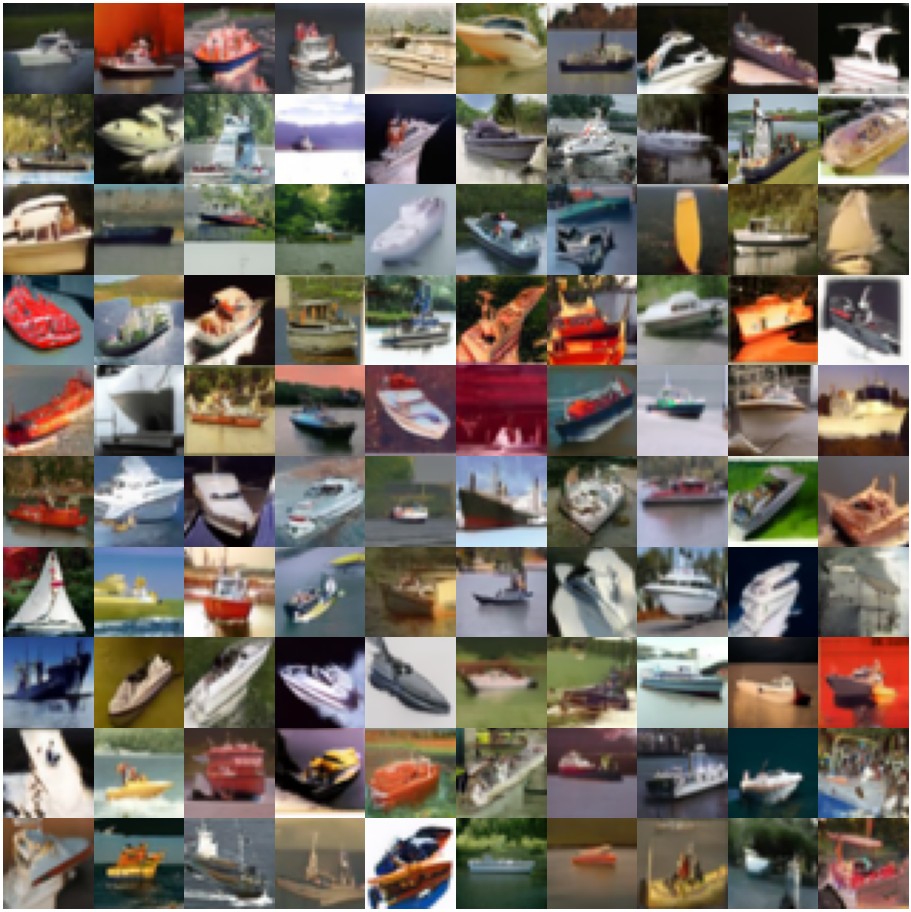}
    \end{subfigure}
    \begin{subfigure}[b]{0.48\linewidth}
        \includegraphics[width=0.48\linewidth]{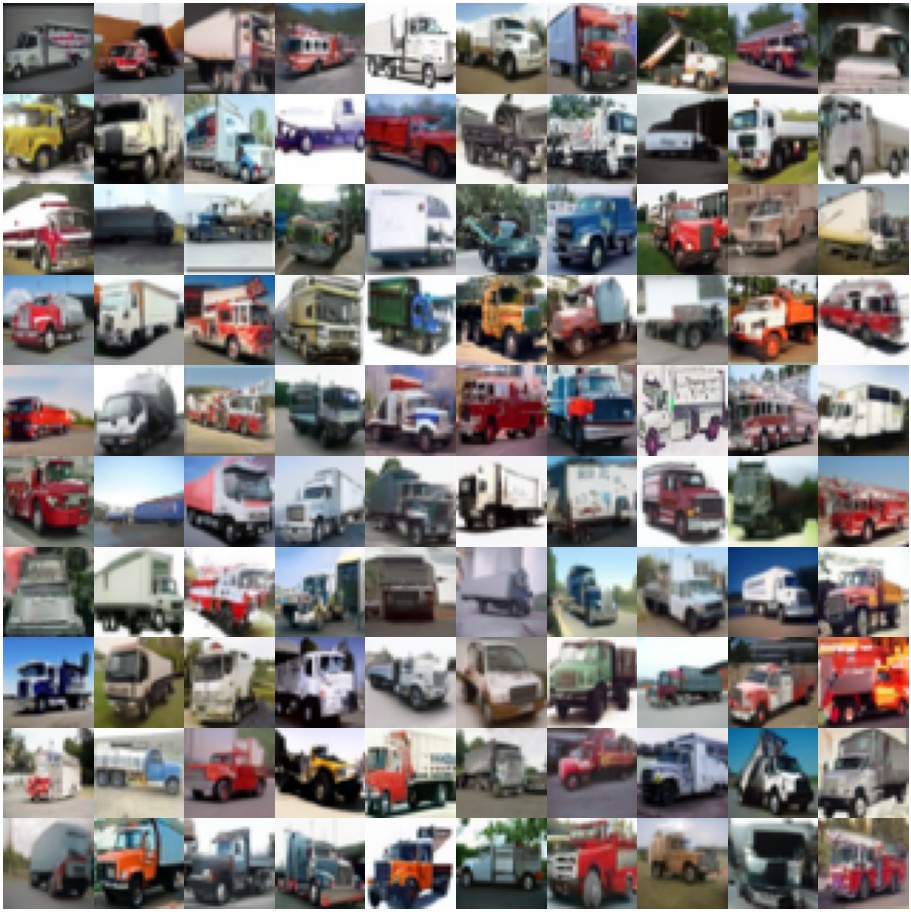}
        \includegraphics[width=0.48\linewidth]{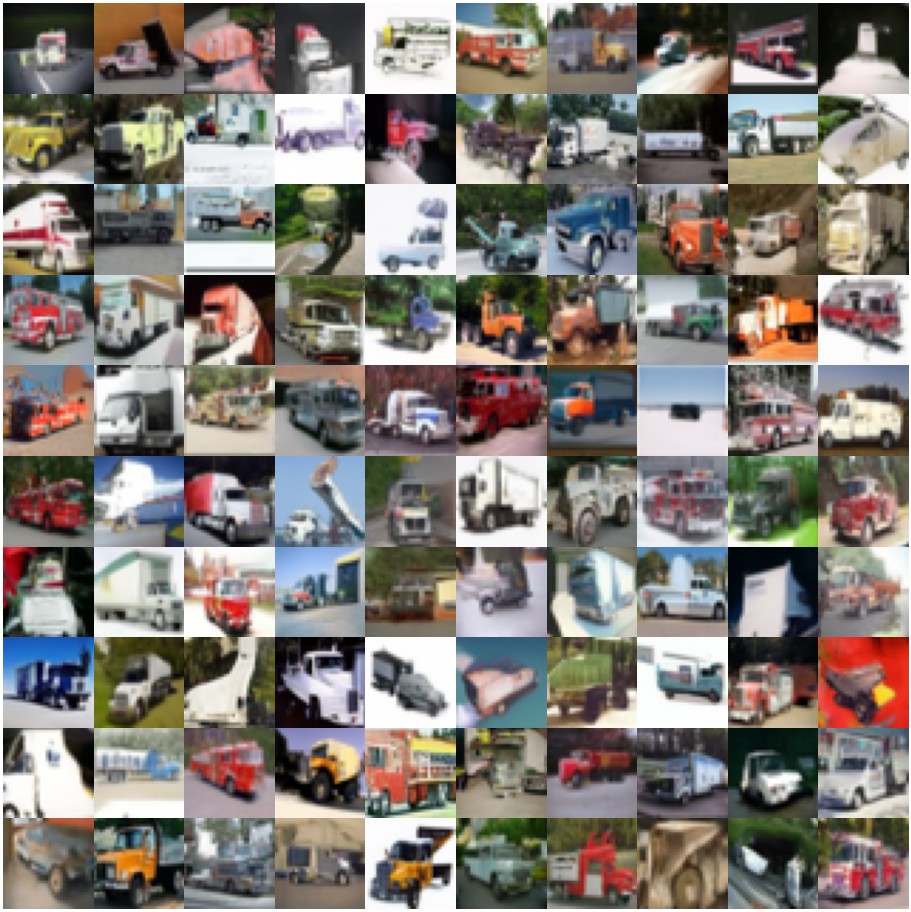}
    \end{subfigure}
    \caption{Synthetic images from the baseline sampling process (left) and our approach (right) for each class on the CIFAR-10 dataset. We use identical random seed for both approaches.}
    \label{fig: add_demo_images_cifar}
\end{figure*}

\begin{figure*}[!btp]
     \centering
     \begin{subfigure}[t]{0.32\textwidth}
        \raisebox{-\height}{\includegraphics[width=\linewidth]{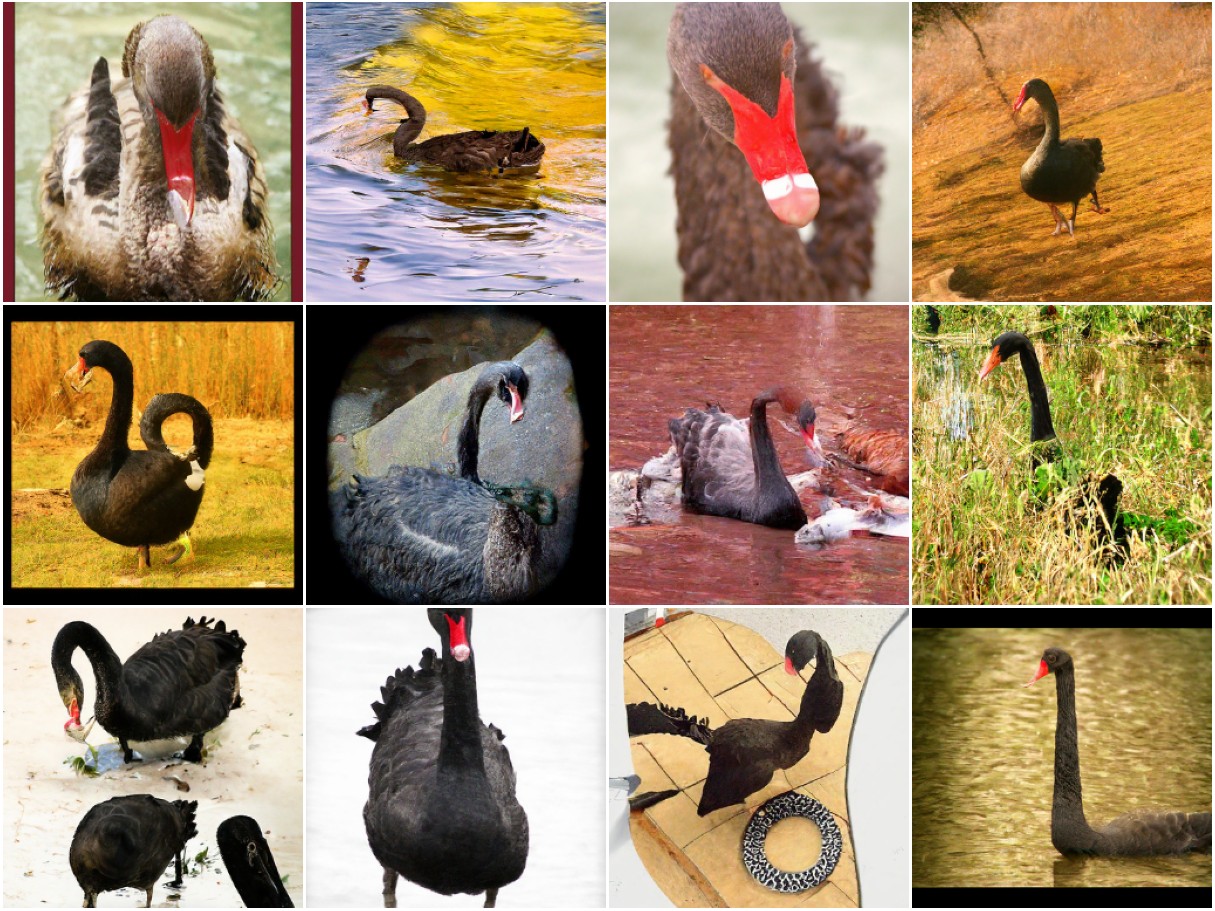}}%
        \vspace{.6ex}
        \raisebox{-\height}{\includegraphics[width=\linewidth]{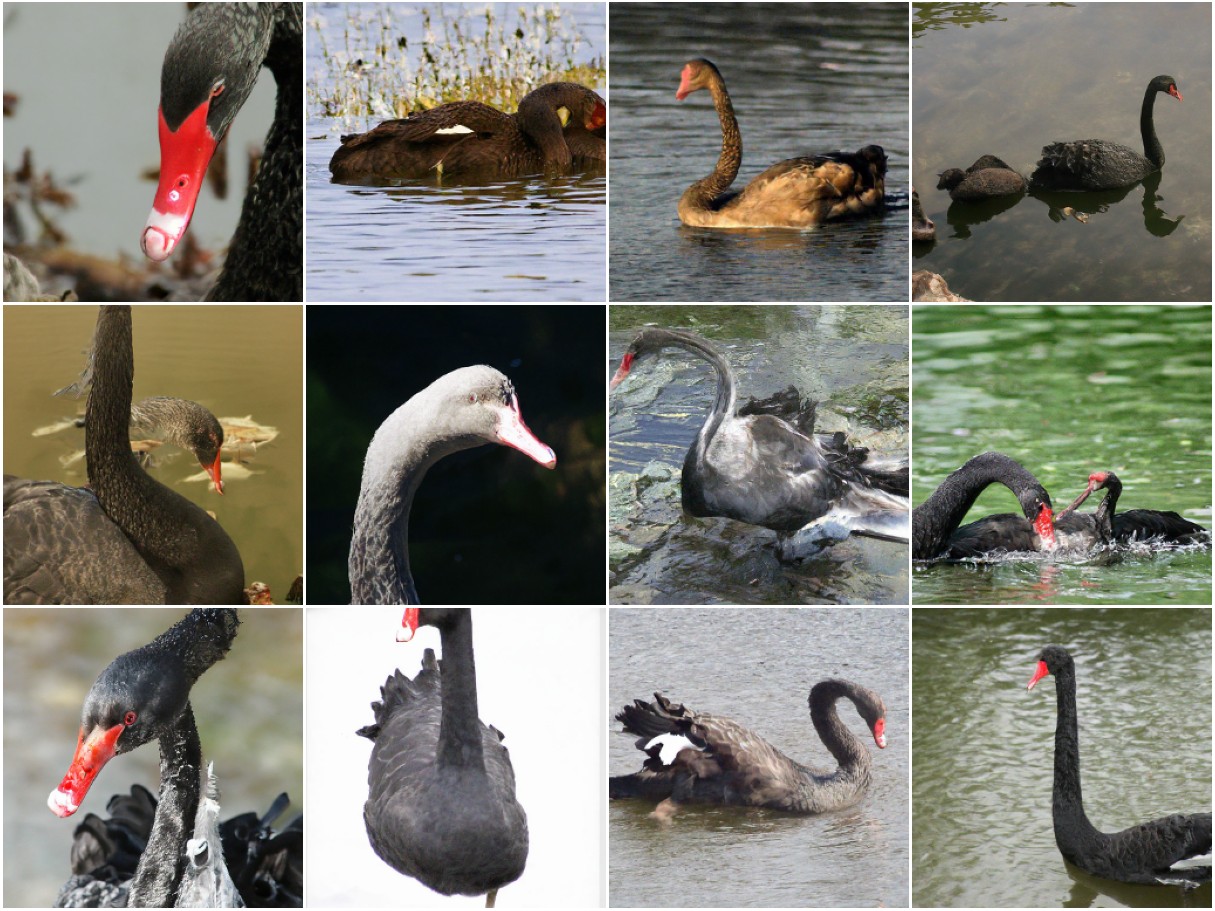}}
        \caption{ImageNet (class 0 and 7)}
    \end{subfigure}
    \hfill
    \begin{subfigure}[t]{0.32\textwidth}
        \raisebox{-\height}{\includegraphics[width=\linewidth]{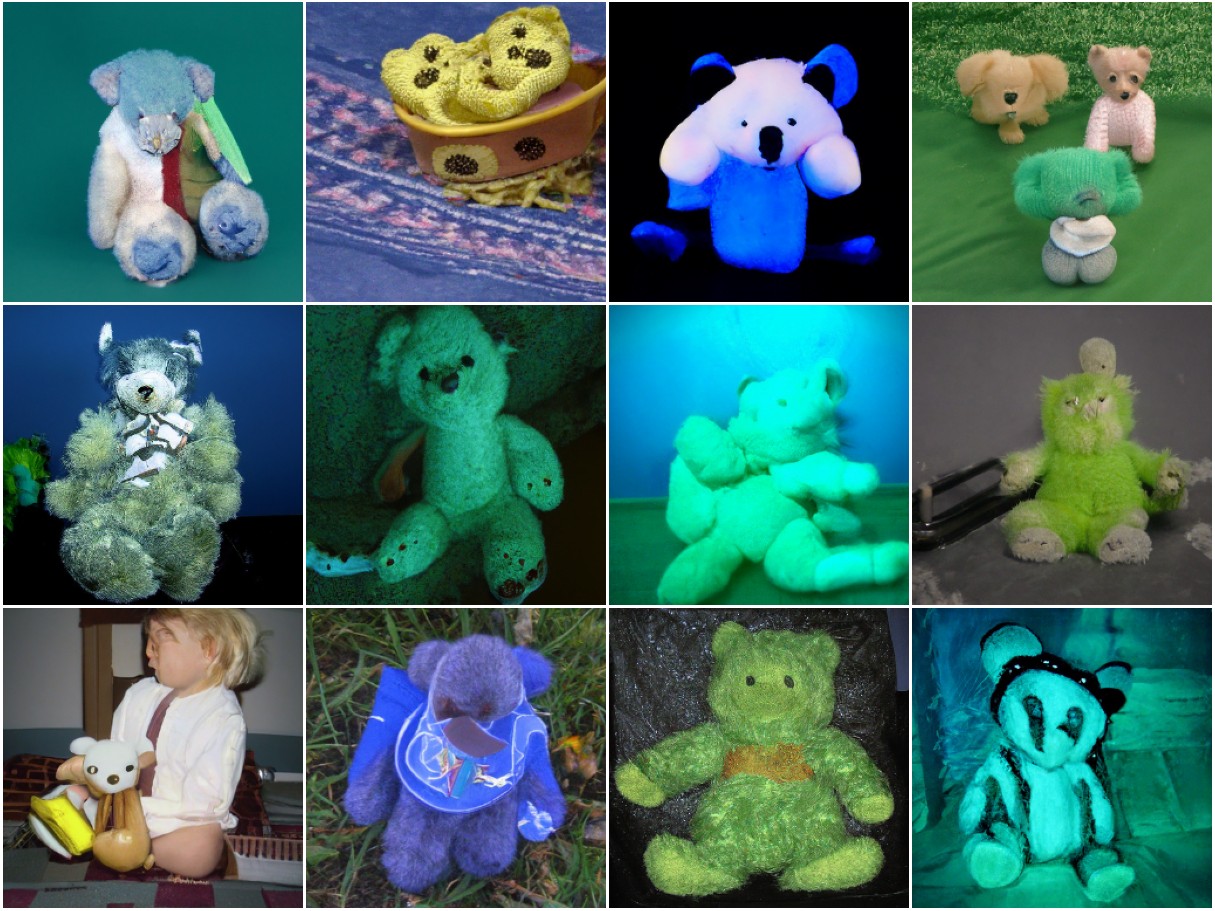}}%
        \vspace{.6ex}
        \raisebox{-\height}{\includegraphics[width=\linewidth]{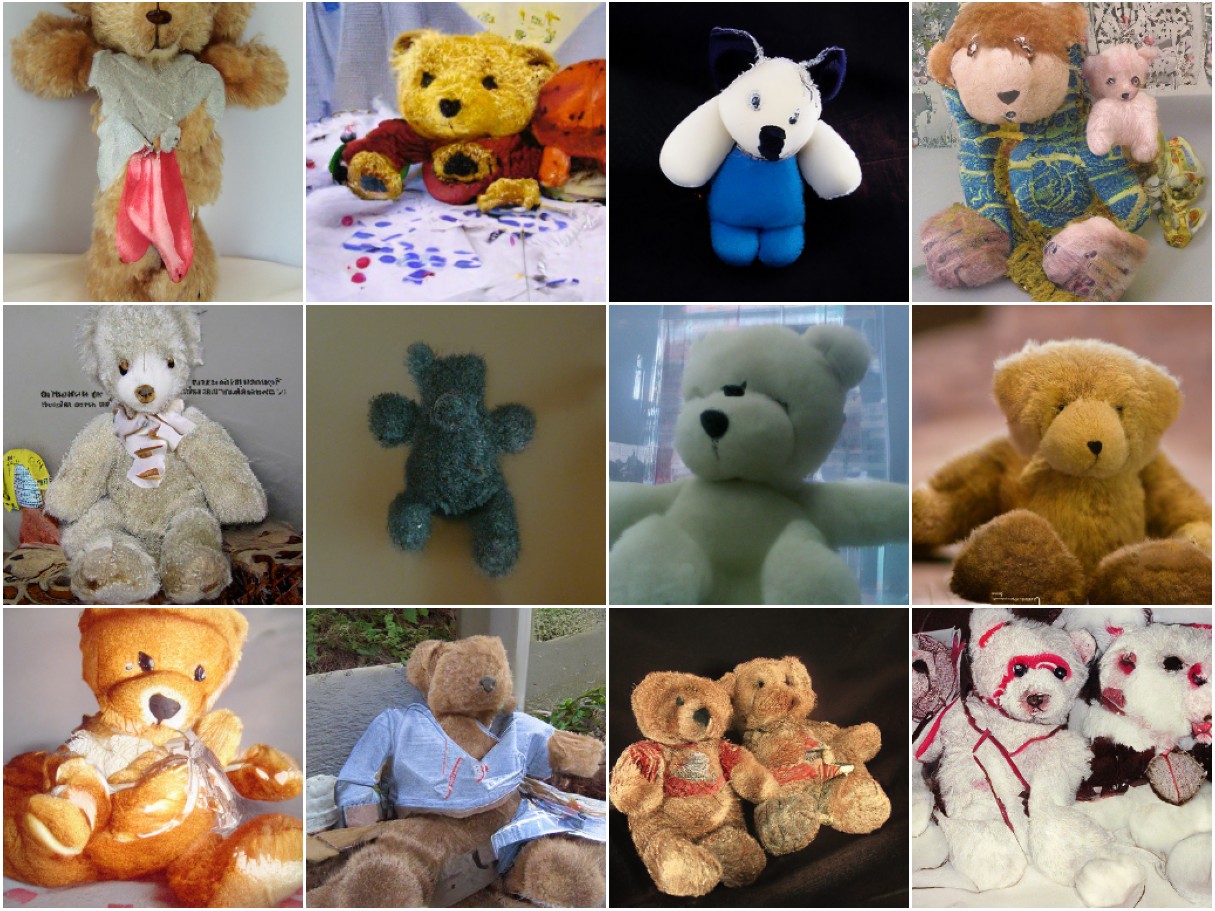}}
        \caption{ImageNet (class 73) and CIFAR-10}
    \end{subfigure}
    \hfill
    \begin{subfigure}[t]{0.32\textwidth}
        \raisebox{-\height}{\includegraphics[width=\linewidth]{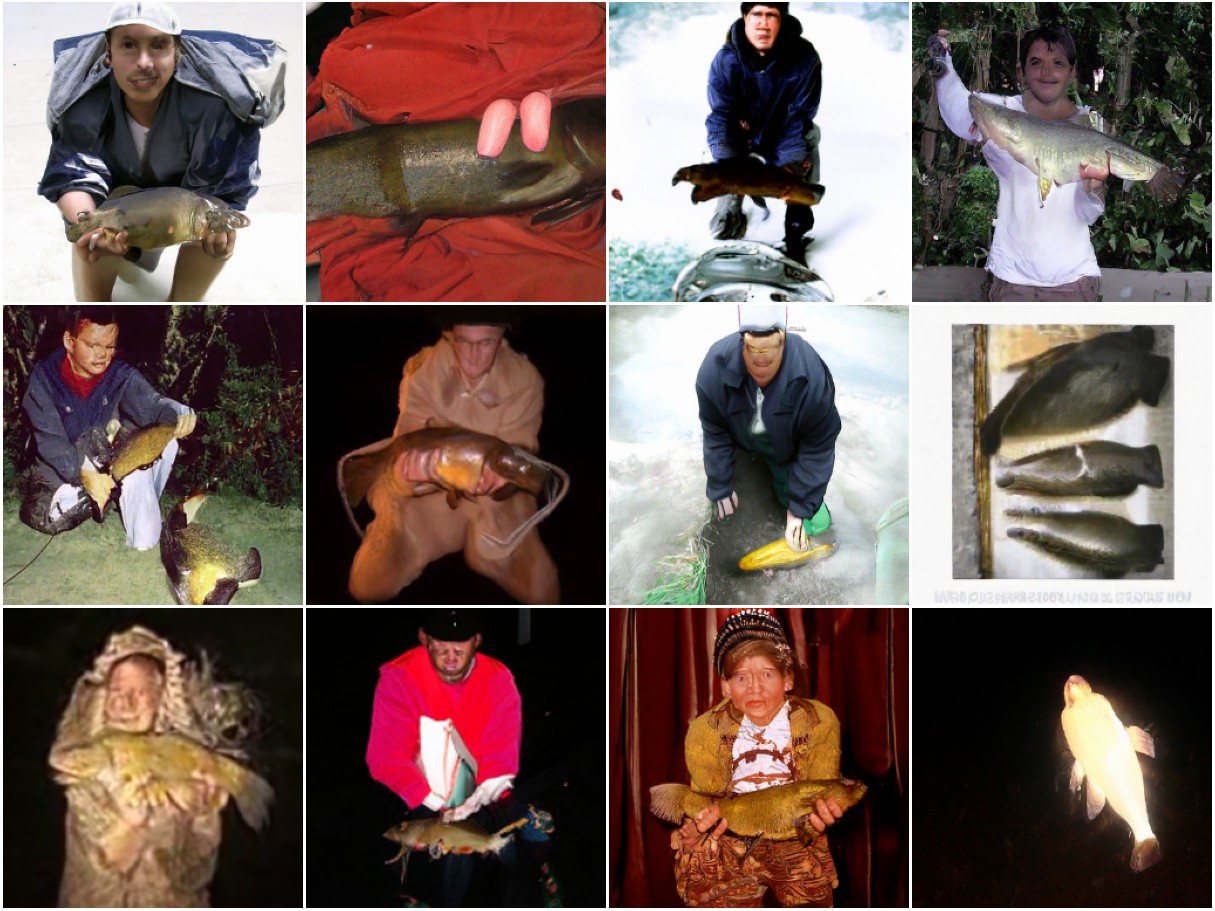}}%
        \vspace{.6ex}
        \raisebox{-\height}{\includegraphics[width=\linewidth]{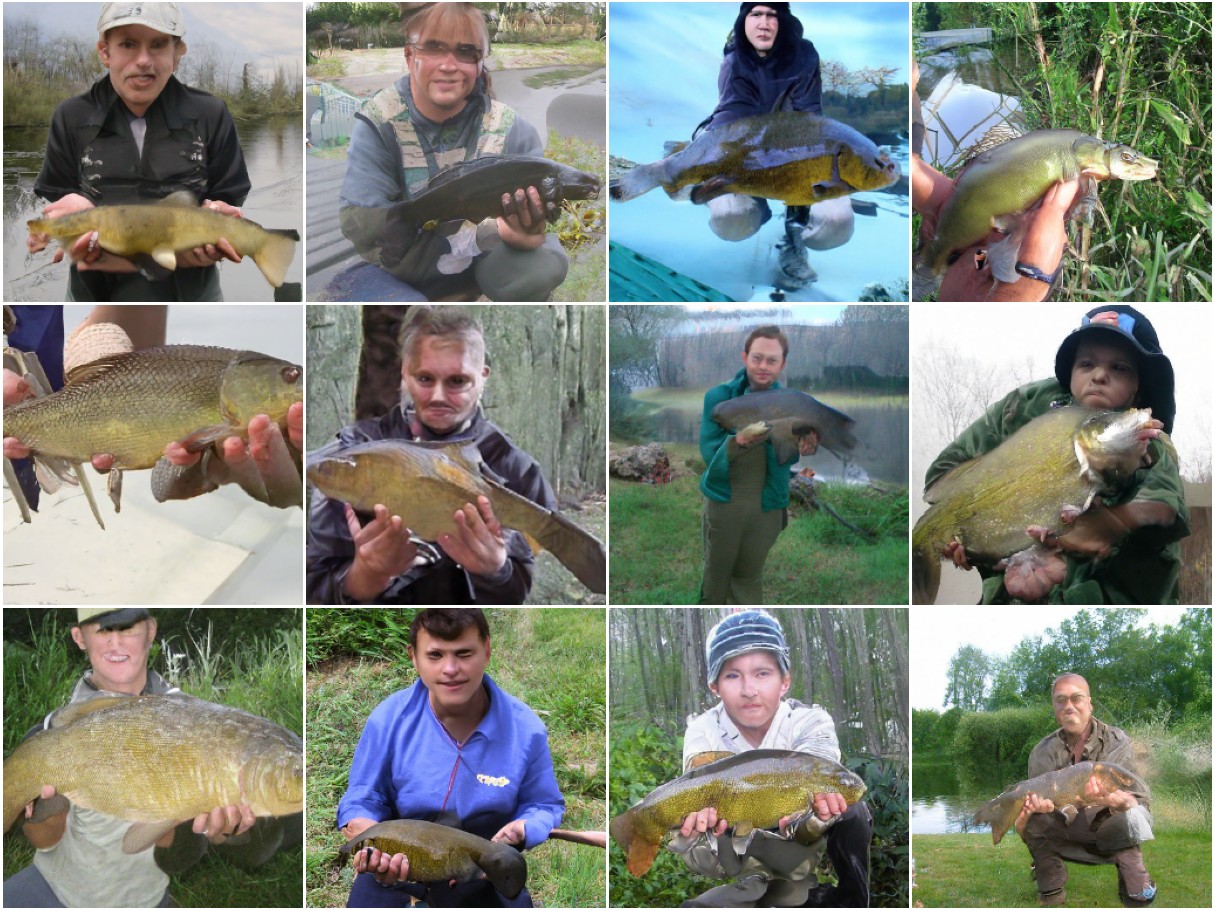}}
        \caption{ImageNet (class 73) and CIFAR-10}
    \end{subfigure}
    \caption{Synthetic images from the baseline sampling process (bottom) and our approach (top) for few classes on the ImageNet dataset. We use identical random seed for both approaches.}
    \label{fig: add_demo_images_im1k}
\end{figure*}

We present additional images to compare the baseline and our sampling process in figure~\ref{fig: add_demo_images_cifar} and ~\ref{fig: add_demo_images_im1k}.  

\end{document}